\documentclass{article}

\PassOptionsToPackage{numbers, compress}{natbib}

 \usepackage[preprint]{neurips_2026}

\usepackage[utf8]{inputenc} 
\usepackage[T1]{fontenc}    
\usepackage{hyperref}       
\usepackage{url}            
\usepackage{booktabs}       
\usepackage{amsfonts}       
\usepackage{nicefrac}       
\usepackage{microtype}      
\usepackage{xcolor}         
\usepackage{algorithm}
\usepackage{algorithmic}
\usepackage{amsmath} 
\usepackage{amssymb} 
\usepackage{graphicx} 
\usepackage{makecell}
\usepackage{wrapfig}
\usepackage{enumitem}
\usepackage{amsthm}  

\usepackage{listings}   
\usepackage{multirow} 
\usepackage{tabularx, array, colortbl, xcolor}
\definecolor{lightgray}{gray}{0.9}
\usepackage{caption} 
\usepackage{subcaption}    
\usepackage{rotating}
\usepackage{makecell}
\usepackage{graphicx}
\usepackage{float}
\usepackage{placeins}
\usepackage{wrapfig}
\usepackage{xurl}
\usepackage{hyperref}

\newcommand{\blackcircnum}[1]{%
  \tikz[baseline=(c.base)]{
    \node[draw,circle,inner sep=0.5pt,thick,fill=black,text=white](c){\footnotesize #1};
  }%
}

\usepackage[most]{tcolorbox}

\newcounter{prompt}

\newtcolorbox[use counter=prompt]{stageprompt}[1]{
  breakable,
  colback=gray!5,
  colframe=black!50,
  title={#1}, 
  fonttitle=\bfseries,
  fontupper=\small,
  boxrule=0.5pt,
  arc=2pt,
  left=6pt,
  right=6pt,
  top=6pt,
  bottom=6pt,
  boxsep=3pt,
  before skip=6pt,
  after skip=6pt
}

\tcbuselibrary{skins}
\usepackage{chngcntr}

\newcounter{finding}[section]
\renewcommand{\thefinding}{\thesection.\arabic{finding}}
\newcommand{\finding}[1]{%
    \refstepcounter{finding}%
    \label{#1}%
}

\definecolor{TakeawayBlue}{HTML}{F3F7FB}
\definecolor{TakeawayBorder}{HTML}{4A6FA5}

\newtcolorbox{takeawaybox}{
    colback=TakeawayBlue,
    colframe=TakeawayBorder,
    boxrule=0.6pt,
    arc=2pt,
    left=5pt,
    right=5pt,
    top=4pt,
    bottom=4pt,
    before skip=6pt,
    after skip=6pt,
    fonttitle=\bfseries,
    title=Takeaway,
}

\definecolor{gs1}{HTML}{FEF5EE}  
\definecolor{gs2}{HTML}{FCE5D2}  
\definecolor{gs3}{HTML}{F8CDAE}  
\definecolor{gs4}{HTML}{F2B286}  
 
\definecolor{cs1}{HTML}{F1F8FD}  
\definecolor{cs2}{HTML}{D9ECF8}
\definecolor{cs3}{HTML}{B8DAF0}
\definecolor{cs4}{HTML}{8FC3E5}  
 
\definecolor{rk1}{HTML}{F8C9C0}  
\definecolor{rk2}{HTML}{FBE8A6}  
\definecolor{rk3}{HTML}{C9E6C5}  
 
\newcommand{\first}[1]{\cellcolor{rk1}$\mathbf{#1}$}
\newcommand{\second}[1]{\cellcolor{rk2}\underline{#1}}
\newcommand{\third}[1]{\cellcolor{rk3}#1}
 
\definecolor{panelBg}{HTML}{DDE6F0}
\definecolor{gapColumn}{HTML}{EDEDED}

\usepackage{pifont}
\usepackage{tikz}
\definecolor{mygreen}{RGB}{46,125,50}
\definecolor{myred}{RGB}{198,40,40}
\definecolor{myorange}{RGB}{245,124,0}
\definecolor{mygray}{RGB}{120,120,120}
\definecolor{CASpurple}{HTML}{7A3E9D}

\usepackage{tikz}
\usepackage{xcolor}
\usepackage{helvet}

\newcolumntype{Y}{>{\centering\arraybackslash}X}

\newcolumntype{L}[1]{>{\raggedright\arraybackslash}p{#1}}

\newcommand{\dataset}{{\color{black}\textsc{GraphGym}}}

\title{Unified Multi-Dimensional Benchmark for Complex Graph Reasoning in Large Language Models}
\author{%
  Fali Wang$^{1}$,
  Ali Al-Lawati$^{1}$,
  Iliyas Bektas$^{1}$,
  Jinxuan Fang$^{1}$,
  Alek Melenski$^{1}$,
  Tianxiang Zhao$^{1}$,\\
  \textbf{Yao Ma}$^{2}$,
  \textbf{Suhang Wang}$^{1}$
  \\
  $^1$The Pennsylvania State University, University Park, PA, USA \\
  $^2$Rensselaer Polytechnic Institute, Troy, NY, US \\
  \texttt{\{fqw5095,szw494\}@psu.edu} \\
}

\begin{document}

\maketitle

\begin{abstract}
Graph reasoning provides a promising testbed for evaluating the reasoning ability of large language models (LLMs), as graph instances can be programmatically generated, structurally controlled, and naturally scaled to long-input settings. 
However, existing graph reasoning benchmarks have limited coverage of data complexity, rely heavily on manual construction, and lack unified evaluation across text-based and code-based reasoning modes. 
To address these limitations, we propose {\dataset}, a five-stage \textit{semi-automatic} framework for constructing complex graph reasoning benchmarks. 
It expands benchmark coverage along five dimensions: \textit{Graph Size}, \textit{Task Complexity}, \textit{Task Description}, \textit{Graph Loading}, and \textit{Task Source}. 
The framework uses an LLM-based data generator to automatically produce task descriptions, graph data, reference solutions, graph-loading scripts, question forms, and evaluation scripts, while retaining human validation at key quality-control stages. 
Based on it, we construct a benchmark with $202$ tasks and evaluate LLMs under text-based, code-based, and augmented reasoning settings. 
Experiments show that the complexity dimensions reveal model limitations that are less visible in existing benchmarks; existing fine-tuned models struggle to generalize to {\dataset}, whereas 
retrieval-augmented methods show scenario-dependent adaptability, improving textual reasoning but not consistently improving coding reasoning.
These findings suggest that ours serves as a challenging and diagnostic benchmark for graph reasoning and provides empirical guidance for future enhancement methods.
Code and dataset will be published soon.

\end{abstract}

\section{Introduction}
Large language models (LLMs)
have shown strong reasoning ability on benchmarks for mathematical reasoning \cite{hendrycks2021measuring, huang2025mathperturb}, code generation \cite{yu2025humaneval, zheng2025livecodebench}, and commonsense and logical reasoning \cite{lin2025zebralogic, white2025livebench}. However, these mainstream evaluation paradigms present two major weaknesses. First, prior works~\cite{deng2024investigating, matton2024leakage} show that several mathematical and coding benchmarks have leaked into LLM training or development pipelines, raising concerns whether strong benchmark performance reflects memorization rather than genuine reasoning~\cite{xu2024benchmarking, aiyappa2023can}.
Moreover, refreshing these benchmarks with new constructed problems is costly, and once released, they may again be ingested into future training corpora. 
Second, existing reasoning benchmarks are generally too short to test long-context reasoning: the average problem lengths of GSM8K \cite{cobbe2021training} and MATH \cite{hendrycks2021measuring} are only 60 and 67 tokens, respectively, while GPQA \cite{rein2024gpqa}, including both question stem and answer options, average about 169 tokens only.

These weaknesses have motivated the development of \textit{graph-based} reasoning benchmarks~\cite{wang2024languagemodelssolvegraph, guo2023gpt4graph, fatemi2023talklikegraphencoding, zhang2024llm4dyg, luo2024graphinstruct, tang2025grapharena, dai2025how, xu2026graphomni, zhang2026exposing, zhang2024can, wu2025grapheval36k, li2024can, hu2025rethinking, wang2026graphskill}, which enable controllable data generation while mitigating data leakage, and naturally scaling to longer inputs.
In particular, compared with traditional math or code problems, graph reasoning benchmarks offer three advantages: (i) they require reasoning over explicit structures and algorithmic relations, such as connectivity, traversal, shortest paths, and flow, making them suitable for evaluating structured multi-step reasoning; (ii) they support programmatic generation of random graph instances and labels, reducing reliance on fixed public questions and helping alleviate contamination, e.g., 
GraphInstruct \cite{luo2024graphinstruct} utilizes classical generators such as Erd\H{o}s--R\'enyi to sample random graphs \cite{erdos1959random},
while GrAlgoBench \cite{zhang2026exposing} performs random walks to samples 8- to 160-node subgraphs from real-world graphs; (iii) the scale of the graph naturally controls prompt length, enabling systematic evaluation under longer contexts. In sparse graphs, node sizes of about 10, 100, 1k, and 10k correspond to prompt lengths of roughly 200, 1k, 9k, and 117k tokens, respectively. More detailed prompt length statistics are in Figure~\ref{fig:prompt-length}, Table~\ref{tab:length-snapshot}, and Appendix~\ref{sec:prompt-length}.

\textbf{Limitations of existing graph benchmarks.}
Existing graph algorithmic reasoning benchmarks can be broadly grouped into two categories. The first is based on classical graph problems, such as connectivity in NLGraph \cite{wang2024languagemodelssolvegraph} and GraphQA \cite{fatemi2023talklikegraphencoding}, which typically predefine task sets, generate graph instances and reference solutions for labeling, and provide template-based problem descriptions together with graph data as model input; these benchmarks mainly target text-based graph reasoning. The second is based on graph problems collected from online assessment (OA) platforms. For example, GraphEval36K \cite{wu2025grapheval36k} selects graph-related problems from LeetCode \cite{leetcode_graph} and pairs them with crafted reference solutions and evaluation scripts, making it suitable for code-based graph reasoning.

\begin{wrapfigure}[12]{r}{0.52\linewidth}
    \centering
    \vskip -0.2em
    \vspace{-1\baselineskip}
    \includegraphics[width=\linewidth]{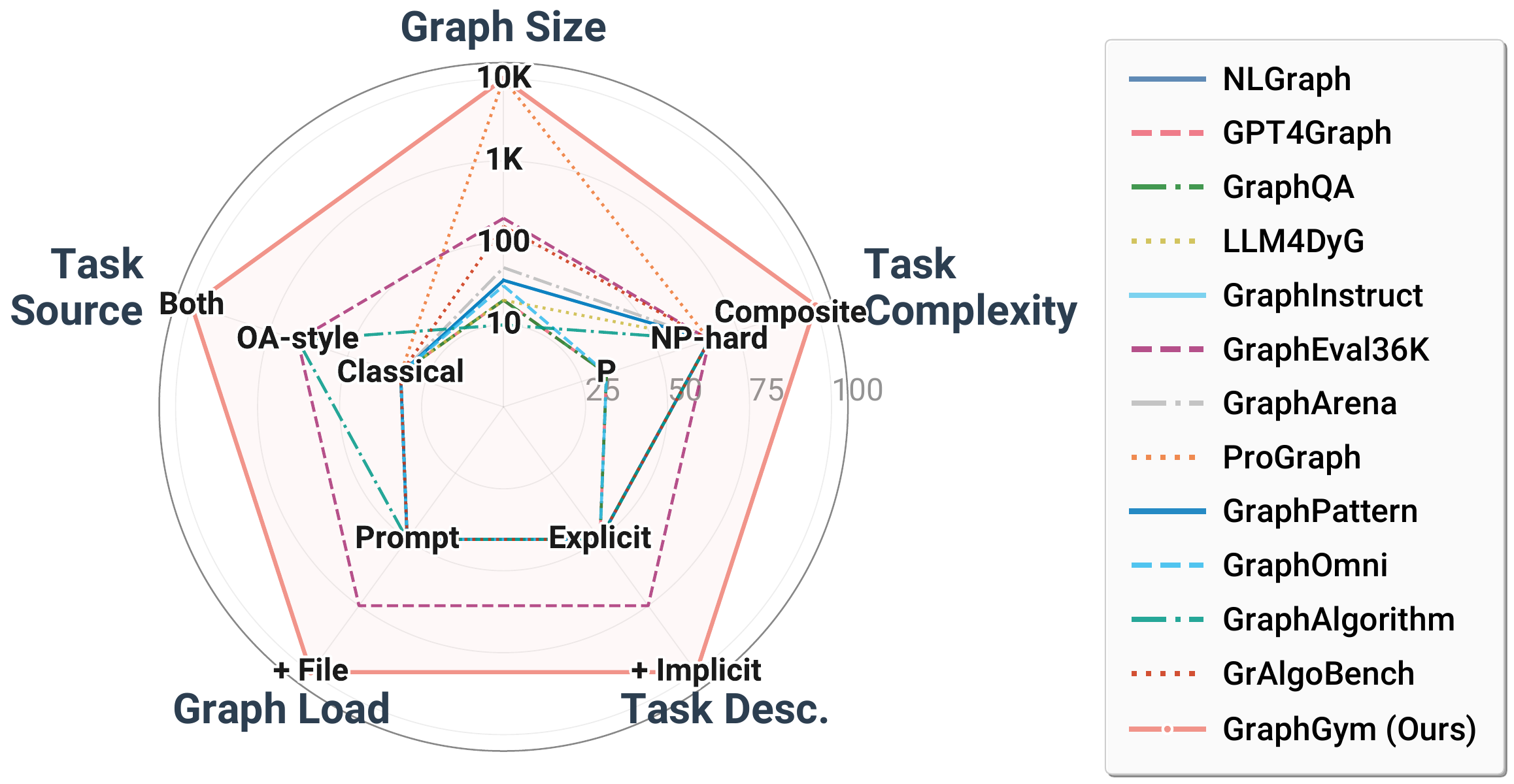}
    \vskip -0.5em
    \caption{Coverage of representative graph reasoning benchmarks across five data-complexity dimensions 
    }
    \label{fig:radar_benchmarks}
    \vspace{-1.4\baselineskip}
\end{wrapfigure}
Despite the growing number of such datasets, we identify three common limitations. \textbf{First, existing benchmarks provide insufficient coverage of data complexity.} To characterize this issue systematically, we analyze benchmarks along five dimensions: \textit{Graph Size}, \textit{Task Complexity}, \textit{Task Description}, \textit{Graph Loading}, and \textit{Task Source} (see Appendix~\ref{sec:benchmark-scoring} for detailed definitions). Figure~\ref{fig:radar_benchmarks} summarizes the coverage of 13 representative benchmarks (including ours) along these dimensions. 
Overall, we observe that existing benchmarks: (i) rarely include 1K- or 10K-node instances for meaningful long-context stress testing, (ii) focus primarily on single-task settings rather than genuinely challenging compositional tasks, (iii) typically adopt explicit graph terms rather than requiring models to infer graph problems from implicit natural-language descriptions, and (iv) usually embed graph data directly in the prompt, with local-file-based graph loading rarely considered. Moreover, (v) most benchmarks are centered on classical graph problems. While this is reasonable for evaluating text-based reasoning, it substantially reduces difficulty in the code reasoning setting, since many classical graph tasks already have mature implementations in Python graph libraries such as NetworkX \cite{hagberg2008exploring}. Prior work shows that, under code-based reasoning, models can often solve such tasks almost perfectly by invoking existing graph libraries \cite{zhang2024gcoderimprovinglargelanguage, zhang2026exposing}. Benchmarks relying solely on classical graph problems are therefore insufficient for fully characterizing LLM performance in more open-ended and challenging graph reasoning scenarios.

\textbf{Second, existing benchmarks largely rely on manually designed data construction pipelines} and therefore do not support automatic expansion. Whether built from classical graph problems or OA-style tasks, they typically require manual task design, graph data generation programming, reference-solution implementation, and evaluation scripting, making scalable benchmark extension human-consuming. In contrast, an automated construction framework with minimal human involvement can reduce this cost and support continual benchmark updates. It can also mitigate data contamination by continuously generating new task instances, thereby reducing memorization of fixed public problems.

\textbf{Third, existing benchmarks are usually designed for only one reasoning mode, either text-based or code-based reasoning}, and lack a unified benchmark for jointly evaluating and systematically comparing both. However, both are critical for understanding the capability boundary of LLMs: text-based reasoning tests internal reasoning, while code-based reasoning additionally evaluates problem understanding, computational modeling, and the effective use of code and external tools. As the two modes differ fundamentally in solving mechanism, capability requirements, and sources of error, evaluating only one provides an incomplete picture of LLM's graph reasoning capability.

To address the above limitations, we propose {\dataset}, a five-stage \textit{semi-automatic} framework for constructing graph reasoning benchmarks. It expands data complexity along five dimensions: \textit{Graph Size}, with instances up to $10{,}000$ nodes; \textit{Task Complexity}, through composite tasks built from multiple seed tasks; \textit{Task Description}, with both explicit graph-theoretic descriptions and implicit real-world scenarios; \textit{Graph Loading}, with both inline and file-based loading supported by generated scripts; and \textit{Task Source}, covering both classical graph tasks and OA-style tasks. To reduce manual construction cost, an LLM-based data generator automatically produces task definitions, graph instances, graph-loading scripts, reference solutions, questions, and evaluation scripts, with human validation reserved for key stages. 
To unify text- and code-based reasoning modes, we construct respective question formats and provide corresponding automated evaluators, enabling comparison across reasoning modes.
We construct {\dataset} with $202$ tasks. Each task equips two descriptions, three reasoning-loading scenarios, and four graph sizes, yielding up to $24$ instances.
Leveraging {\dataset}, we study two categories of research questions: \textbf{RQ1}: How do data-complexity dimensions affect LLM graph reasoning? \textbf{RQ2}: Can augmentation methods, such as instruction tuning and retrieval augmentation, improve performance on complex graph reasoning tasks? Our experiments show that the proposed complexity dimensions effectively expose model limitations that are less visible in existing benchmarks: classical graph tasks may be close to saturated under code-based reasoning because many can be solved with mature Python graph libraries; text-based reasoning degrades sharply on large graphs, with Qwen2.5-72B dropping from $31.7$ EM at size $10$ to $0.5$ at size $10{,}000$; and increasing combo size, i.e., the number of seed tasks in a composite task, from $1$ to $4$ reduces performance by about $20$ points on average. These findings demonstrate that {\dataset} provides a challenging and diagnostic testbed for assessing LLM graph reasoning. We further find that existing fine-tuned models struggle to generalize to {\dataset}, whereas RAG shows better adaptability. This suggests that {\dataset} not only supports evaluation, but also provides empirical guidance for designing future graph reasoning enhancement methods. 

Our \textbf{main contributions} are threefold. \textbf{(i)} We propose a five-stage LLM-based, semi-automatic framework for complex graph reasoning benchmark construction, which can generate task descriptions, graph data, reference solutions, questions, graph-loading and evaluation scripts with limited human validation. \textbf{(ii)} We construct {\dataset}, a dataset with $202$ tasks covering five data-complexity dimensions.
\textbf{(iii)} We systematically evaluate LLMs under text-based, code-based, and enhanced reasoning settings, revealing key findings that affect graph reasoning performance.

\section{Related Work}
\noindent\textbf{Benchmarks for LLM-based Graph Reasoning.}
Graph reasoning provides an effective testbed for evaluating LLMs because task difficulty is controllable, data can be generated programmatically at scale, and instances are less likely to overlap with public pretraining corpora. Existing benchmarks can be broadly grouped into two categories. The first uses classical graph tasks, such as connectivity and shortest paths, including NLGraph~\cite{wang2024languagemodelssolvegraph}, GPT4Graph~\cite{guo2023gpt4graph}, GraphQA~\cite{fatemi2023talklikegraphencoding}, GraphArena~\cite{tang2025grapharena}, ProGraph~\cite{li2024can}, GraphOmni~\cite{xu2026graphomni}, and GrAlgoBench~\cite{zhang2026exposing}. The second evaluates code reasoning with OA-style graph problems, such as GraphEval36K~\cite{wu2025grapheval36k} and GraphAlgorithm~\cite{hu2025rethinking}. Specialized benchmarks also study dynamic graphs~\cite{zhang2024llm4dyg} and graph pattern understanding~\cite{dai2025how}. However, existing benchmarks remain limited by predefined task sets, substantial manual effort for labeling and evaluation, and coarse treatment of task complexity. We address these limitations with a semi-automatic graph benchmark construction and evaluation framework. Details are provided in Appendix~\ref{sec:appendix_benchmarks4graph}.

\noindent\textbf{Methods for LLM-based Graph Reasoning.}
Based on execution mode, existing methods for improving LLM graph reasoning can be broadly divided into text-based and code-based methods. 
\textit{Text-based graph reasoning} treats graph problems as natural-language reasoning tasks over textualized graphs. Prior work improves this setting through prompting methods, such as zero-shot, few-shot, chain-of-thought, and self-consistency \cite{fatemi2023talklikegraphencoding}, Build-a-Graph and Algorithmic Prompting \cite{wang2024languagemodelssolvegraph}, and reason-then-code strategies~\cite{hu2025rethinking}, as well as fine-tuning methods such as GraphWiz \cite{chen2024graphwizinstructionfollowinglanguagemodel}, GraphInstruct \cite{luo2024graphinstruct}, and GraphSilo~\cite{peng2025rewarding}. However, instruction tuning often shows limited transferability across graph tasks. \textit{Code-based graph reasoning} treats graph reasoning as program synthesis and execution. Representative methods include direct code generation and pseudocode injection~\cite{cai2024codegraphenhancinggraphreasoning, gong2025pseudocodeinjectionmagicenablingllms}, retrieval-augmented coding with documentation or tool descriptions~\cite{li2025graphteamfacilitatinglargelanguage, wang2026graphskill, wang2025GraphToolInstruction}, and fine-tuning approaches such as GCoder~\cite{zhang2024gcoderimprovinglargelanguage}. Our benchmark supports unified evaluation of both paradigms. Detailed discussion is provided in Appendix~\ref{sec:appendix_relatedwork_method}.

\section{Graph Benchmark Construction Framework}
\label{sec:construction}


\begin{figure}[t]
  \centering
  \includegraphics[width=\textwidth]{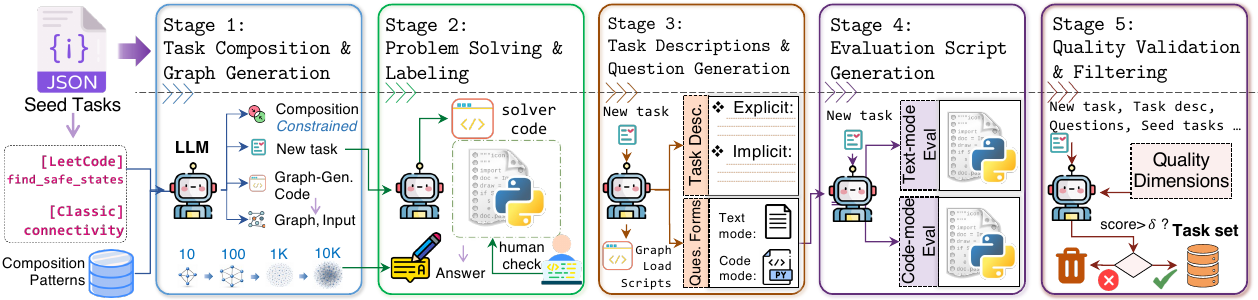}
  \vskip -0.3em
  \caption{End-to-end graph reasoning benchmark construction framework of {\dataset}. Zoomed-in views of individual stages are provided in Figures~\ref{fig:stage1_example}--\ref{fig:stage5_example} in the Appendix. 
  }
  \vskip -1.2em
  \label{fig:pipeline}
\end{figure}

To support scalable graph reasoning benchmark construction, we design {\dataset} as a \textit{semi-automatic} LLM-based framework with limited human supervision for quality control. For simplicity, we use {\dataset} to denote both the framework and the resulting dataset.
Our pipeline has two main advantages: (i) Unlike existing graph reasoning benchmarks with limited coverage of data complexity \cite{wang2024languagemodelssolvegraph, guo2023gpt4graph, wu2025grapheval36k, hu2025rethinking, xu2026graphomni, zhang2026exposing}, {\dataset} systematically covers five data-complexity dimensions. (ii) It generates not only graph tasks and data, but also an automated evaluation pipeline, including graph-loading scripts and task-specific evaluation scripts.
As shown in Figure~\ref{fig:pipeline}, the framework consists of five stages, which we describe below. More details are provided in Appendix~\ref{sec:appendix_construction}.

\textbf{Preliminary Stage: Defining Seed Tasks and Composition Patterns.}
We introduce compositional task complexity by first defining a set of seed tasks as atomic units for task composition. Existing graph reasoning benchmarks mainly focus on classical graph algorithms, with their collected tasks shown in Table~\ref{tab:graph_benchmark_tasks}. To cover prior task scopes, we collect $40$ classical graph tasks, such as shortest path. Since such tasks may be directly solved with existing Python graph libraries in coding settings, we further include $60$ open-ended graph tasks from LeetCode, an online assessment platform, to increase task diversity. Examples are shown in Figure~\ref{fig:seed_tasks}, with the full seed tasks provided in our code repository. Therefore, \textit{\textcolor{CASpurple}{we cover the \textbf{Task Source} dimension by including both classical and OA-style seed problems}}.
We also predefine seven composition patterns: \emph{sequential}, \emph{constrained}, \emph{hierarchical}, \emph{counterfactual}, \emph{map-reduce}, \emph{aggregate}, and \emph{logical-comparative}, as detailed in Figure~\ref{fig:composition}.

\textbf{Stage 1: Task Composition and Graph Generation.}
Stage~1 constructs composite graph reasoning tasks and their corresponding graph instances. Given a \texttt{combo size}, sampled seed tasks, and seven candidate composition patterns, the LLM-based data generator selects a composition pattern, a new task definition, and task-adaptive graph generation code. A single seed task is treated as a degenerate composite task. The \texttt{combo size} controls the number of seed tasks in the composition and takes values in $\{1,2,3,4\}$ in our dataset, although the framework is extensible to larger values. In Figure~\ref{fig:stage1_example} and Prompt~\ref{prompt_stage1}, the prompt provides seed-task definitions, candidate composition patterns, and the expected output format, and the LLM produces the code for generating graphs, which is then executed to produce instances at four graph sizes: $10$, $100$, $1{,}000$, and $10{,}000$ nodes. Thus, \textit{\textcolor{CASpurple}{this stage covers data complexity through compositionality \textbf{Combo Size} and graph scale through controlled \textbf{Graph Size}.}} Details are provided in \S~\ref{sec:appendix_stage1}.

\textbf{Stage 2: Solver Generation and Label Annotation.}
After constructing tasks and graph instances, the LLM-based data generator generates reference solution programs and uses them to annotate gold labels. This stage is necessary because accurate labels are essential for reliable evaluation, while fully manual annotation is costly and error-prone for large-scale compositional graph reasoning tasks.
Given the task description and graph-data format in Prompt~\ref{prompt_stage2}, the LLM generates task-specific executable Python solver code, as shown in Figure~\ref{fig:stage2_example}.
To improve reliability, human annotators create functional test cases to validate the generated solution. If the solution code fails, error messages and human feedback are provided to the LLM for revision until the code passes validation. This validation step is essential. In our human validation, approximately $80\%$ of the generated solver scripts are correct, but the remaining $20\%$ still contain errors that could compromise benchmark reliability. Since dataset construction requires perfect label accuracy, moderate human involvement can substantially improve quality while remaining manageable in terms of annotation cost. The validated solution code is then executed on the graph instances and input parameters produced in the previous stage to obtain gold labels. This stage reduces manual construction cost while maintaining human oversight over label correctness. Details are provided in \S~\ref{sec:appendix_stage2}. 

\textbf{Stage 3: Task Description and Question Generation.}
In this stage, for each task in Stage 2, the LLM-based data generator creates multiple task descriptions and question formats. Directly using the task definition as the task description, as in prior graph reasoning benchmarks~\cite{wang2024languagemodelssolvegraph, fatemi2023talklikegraphencoding}, is too explicit to meaningfully challenge LLMs' task understanding. To vary the task-description dimension, we therefore construct two types of descriptions: \textit{explicit} descriptions, which directly state the graph problem and structure, and \textit{implicit} descriptions, which present a real-world narrative that must be mapped to the underlying graph problem.
To support both text- and code-mode reasoning, we provide questions in both formats, enabling controlled comparisons across reasoning modes. For code-mode questions, we support two graph-loading settings: \emph{inline}, where graph data are included in the prompt, and \emph{file}, where only a file path is provided and the program loads the graph. Because graph formats vary across tasks and implicit descriptions may require flexible graph loading, we further generate task-specific loading scripts for both modes. These outputs, including two task descriptions, two question formats, and two graph-loading scripts, are generated by the LLM data generator using a prompt in \S \ref{prompt_stage3} that includes the task definition and reference solution program. Thus, \textit{\textcolor{CASpurple}{this stage operationalizes the \textbf{Task Description} and \textbf{Graph Loading} dimensions.}} Details are provided in \S~\ref{sec:appendix_stage3}.

\textbf{Stage 4: Evaluation Script Generation.}
Stage~4 generates task-specific evaluation scripts for fully automated scoring. This stage is necessary because graph reasoning tasks produce heterogeneous outputs, including Boolean values, numerical values, sets, lists, dictionaries, and multi-part results in composite tasks. A single generic evaluator is therefore insufficient. We therefore use the LLM-based data generator to generate task-specific evaluation scripts. For each task, given the task definition, question format, and reference solution code, the LLM uses the prompt in \S~\ref{prompt_stage4} to generate two evaluation scripts: one for text-based reasoning and one for code-based reasoning. The text evaluator extracts the final answer from the model response, parses it into a machine-readable Python object, and compares it with the gold label using type-aware rules. The code evaluator extracts the generated \texttt{solve} function, inserts it into an evaluation template, executes it on benchmark graph instances, and compares the output with the gold label. 
Details of evaluation procedure are provided in \S~\ref{sec:appendix_stage4}.

\textbf{Stage 5: Quality Validation and Filtering.}
Stage~5 filters automatically generated tasks before they are included in the final benchmark. Although earlier stages generate tasks, labels, prompts, and evaluators, they do not guarantee that every instance is clear, structurally appropriate, natural, and uniquely answerable. Therefore, we use an LLM-based quality judge to score each task on a $1$--$5$ scale along four dimensions: \emph{clarity}, \emph{graph suitability}, \emph{naturalness}, and \emph{answer uniqueness}. Given the task descriptions, question formats, and reference solutions, the LLM-based quality judge uses the prompt in \S~\ref{prompt_stage5} to output scores for each dimension. Tasks with a score below $3$ in any dimension are removed to avoid ambiguous or unreliable evaluation instances. This stage serves as the final quality-control step of the construction pipeline. Details are provided in \S~\ref{sec:appendix_stage5}.



\textbf{Human Validation of LLM Quality Judgments.}
To assess the reliability of the Stage~5 LLM judge, we compare its quality scores with human annotations. Two human annotators, H1 and H2, with strong expertise in graph algorithms and formal annotation training, independently score $20$ randomly sampled composite tasks using the same $1$--$5$ rubric. We measure agreement using Cohen's $\kappa$~\cite{cohen1960coefficient} and use quadratic-weighted $\kappa_q$ as the primary metric for ordinal ratings. The LLM judge achieves comparable agreement with humans, with $\kappa_q=0.887$ against H1 and $\kappa_q=0.880$ against H2. These results support the reliability of LLM-based quality filtering. Details are provided in \S~\ref{sec:appendix_human}.



\begin{figure}[t]
    \centering
    \includegraphics[width=0.9\linewidth]{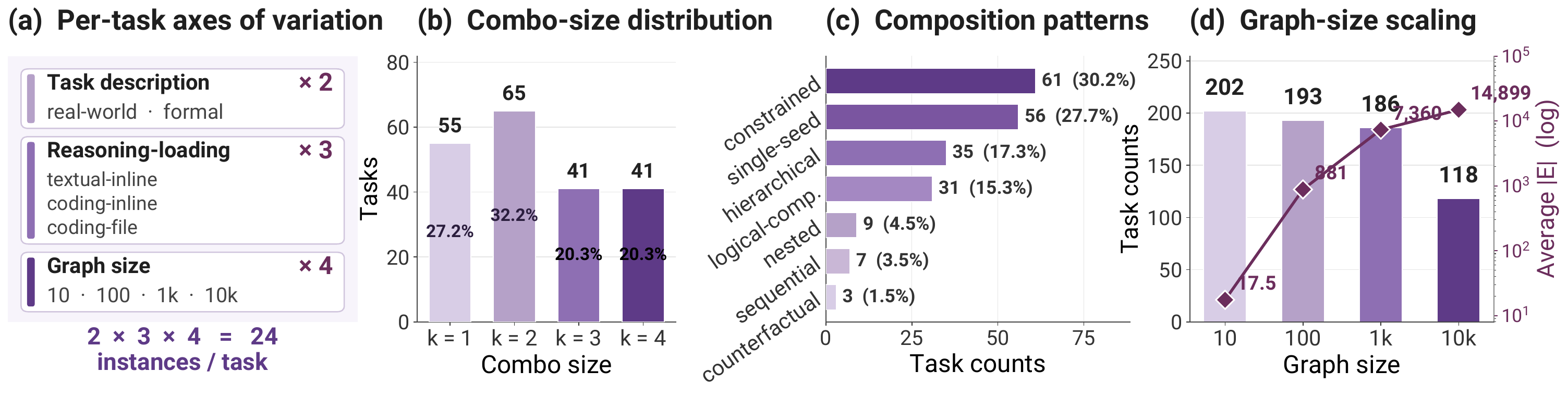}
    \vskip -1em
    \caption{Distribution of {\dataset} across key benchmark dimensions. (a)--(d) summarize the per-task instance, combo size, composition pattern, and graph-size distribution, respectively.}
    \label{fig:benchmark_dist}
    \vskip -1em
\end{figure}

\textbf{Benchmark Distribution.}
Figure~\ref{fig:benchmark_dist} summarizes the distribution of the resulting {\dataset} across several axes, including task-instance construction, graph size, combo size, and composition pattern.
First, for instances under each task (Fig.~\ref{fig:benchmark_dist}(a)), each task is instantiated with two task descriptions (implicit and explicit), three reasoning-loading scenarios (textual-inline, coding-inline, and coding-file), and four graph sizes, resulting in up to $24$ evaluation instances per task and $4{,}848$ instances in total. Some instances are discarded due to issues such as program execution timeout.
Second, for the distribution of composition depths (Fig.~\ref{fig:benchmark_dist}(b)), tasks span varying composition depths: $27.2\%$ are single-seed tasks, while $72.8\%$ combine two to four seed tasks, with a balanced distribution across depths.
Among composition patterns (Fig.~\ref{fig:benchmark_dist}(c)), \emph{constrained} dominates ($30.2\%$), followed by \emph{hierarchical} and \emph{logical-comparative}, while other patterns form a long tail due to stricter construction requirements.
Finally, graph generators are applied at four scales, from $n=10$ to $n=10{,}000$ (Fig.~\ref{fig:benchmark_dist}(d)), with full coverage at small sizes and partial coverage at larger scales due to graph-generation time budget constraints; generation is terminated if it exceeds 10 minutes.
Overall, the distribution ensures diversity across reasoning-loading scenarios, composition structures, and graph scales. 
Detailed statistics, evaluation pipeline and cost analysis are in \S \ref{sec:appendix_benchmark_dist},\ref{sec:appendix_evaluation},\ref{sec:cost-analysis}, respectively.

\section{Experiments}
\label{sec:experiments}

We evaluate diverse LLMs on {\dataset} to establish baselines and analyze how each dimension affects performance. We study two research questions: (\textbf{RQ1}) How do {\dataset} dimensions affect LLM graph reasoning? (\textbf{RQ2}) Do existing reasoning methods, including RAG, finetuning, and task-specific GNNs, improve performance in complex graph-reasoning settings? We evaluate zero-shot CoT under both code- and text-based reasoning, covering three modes: \textit{coding-file} (file path provided), \textit{coding-inline} (graph in prompt), and \textit{textual} reasoning (graph in prompt). Detailed setup, expanded results, and findings are provided in \S~\ref{sec:appendix_graph_sizes}--\S~\ref{sec:appendix_leetcode} for \textbf{RQ1} and \S~\ref{sec:appendix_rq_rag}--\S~\ref{sec:appendix_gnn} for \textbf{RQ2}.



\noindent\textbf{Language Models and Metrics.} We select a diverse set of LLMs across multiple families (e.g., \texttt{Llama}~\cite{grattafiori2024llama}, \texttt{DeepSeek}~\cite{liu2024deepseek}, and \texttt{Qwen}~\cite{bai2023qwentechnicalreport}) and parameter sizes (7B--70B), including coding-specialized models (\texttt{Qwen3-Coder-30B} \cite{yang2025qwen3}) and closed-source models (\texttt{o4-mini}~\cite{openai2024o4mini}). This selection allows us to provide a cross-sectional view of the current LLM landscape across open-source, closed-source, general-purpose, and coding-specialized models. 
We use exact match (EM) and partial credit (PC) as metrics where EM requires full correctness and PC assigns fractional credit to partially correct responses.
Detailed LLM list and metrics are in \S~\ref{sec:appendix_experiment_setup} and \S~\ref{sec:appendix_stage4} respectively.

\subsection{RQ1: Impact of {\dataset} Dimensions}
To address RQ1, we evaluate the impact of each of the five dimensions of {\dataset} in isolation.

\textbf{\blackcircnum{1} Graph Reasoning across Graph Sizes.} To understand the impact of different graph size, we evaluate graph reasoning across four graph sizes, $n\in\{10,100,1{,}000,10{,}000\}$. Large graph sizes provide a stress test on LLM capabilities under long-context and ultra-long-context inputs. 

\finding{finding:graph-size}
\begin{tcolorbox}[
    colback=blue!5, colframe=blue!20,
    boxsep=2pt, left=4pt, right=4pt, top=2pt, bottom=2pt,
    before skip=3pt, after skip=5pt
]
\small\textbf{Finding \thefinding:} 
Graph size is a strong complexity dimension, especially for textual and coding-inline reasoning. 
\end{tcolorbox}

In Table~\ref{tab:merged-size-combo-description-family}, EM decreases monotonically with graph size across all evaluated model-mode combinations. On average, coding-inline-based reasoning drops from $77.9$ EM at $n{=}10$ to $14.2$ at $n{=}10{,}000$, while text-based reasoning also drops sharply from $66.7$ to $10.4$. This indicates that large graph inputs substantially stress LLM graph reasoning, with natural-language reasoning becoming particularly fragile under ultra-long contexts. In contrast, the coding-file reasoning mode remains relatively robust, from $77.7$ to $46.3$ as $n$ from $10$ to $10{,}000$. This is because it delegates graph reading to offline code execution, reducing the LLM's input-processing burden. The remaining performance drop is mainly due to computational constraints, such as timeouts or memory limits on very large graphs. 

\textbf{\blackcircnum{2} Graph Reasoning across Task Complexity.}
Task complexity involves composite tasks that combine $1$--$4$ atomic seed tasks, requiring models to perform multi-step reasoning. We evaluate the impact of task complexity on both code-based and text-based reasoning.

\finding{finding:graph-size}
\begin{tcolorbox}[
    colback=blue!5, colframe=blue!20,
    boxsep=2pt, left=4pt, right=4pt, top=2pt, bottom=2pt,
    before skip=3pt, after skip=5pt
]
\small\textbf{Finding \thefinding:} 
Composition is a stable complexity factor for text- and coding-based LLM reasoning. 
\end{tcolorbox}

Table~\ref{tab:merged-size-combo-description-family} shows that, on average, EM drops by $21.2$, $17.0$, and $14.5$ points from combo size $1$ to $4$ under coding-inline, coding-file, and text-based reasoning, respectively. These results demonstrate that compositional depth is a stable complexity factor, as models must handle longer dependency chains, more seed tasks, and more complex intermediate states.


\textbf{\blackcircnum{3} Explicit vs. Implicit Task Descriptions.}
This dimension measures how task-description complexity affects performance, moving from \textit{explicit} graph-term descriptions to \textit{implicit} real-world scenarios that require models to first formulate the underlying graph problem before solving it.

\finding{finding:explicit-vs-implicit}
\begin{tcolorbox}[
    colback=blue!5, colframe=blue!20,
    boxsep=2pt, left=4pt, right=4pt, top=2pt, bottom=2pt,
    before skip=3pt, after skip=5pt
]
\small\textbf{Finding \thefinding:}
Implicit task descriptions introduce only modest additional difficulty.
\end{tcolorbox}
Table~\ref{tab:merged-size-combo-description-family} shows that explicit descriptions are slightly easier than implicit descriptions across all three reasoning modes. The average Explicit$-$Implicit EM gaps are $+3.5$, $+2.2$, and $+1.6$ for code-file, code-inline, and text-based reasoning, respectively. This suggests that implicit descriptions make task solving modestly harder, but their effect is smaller than that of graph size and task complexity, because LLMs are capable of inferring the underlying graph problem from scenario-based descriptions.


\begin{wrapfigure}[14]{r}{0.52\linewidth}
    \centering
    \vspace{-1.1\baselineskip}
    \includegraphics[width=1\linewidth]{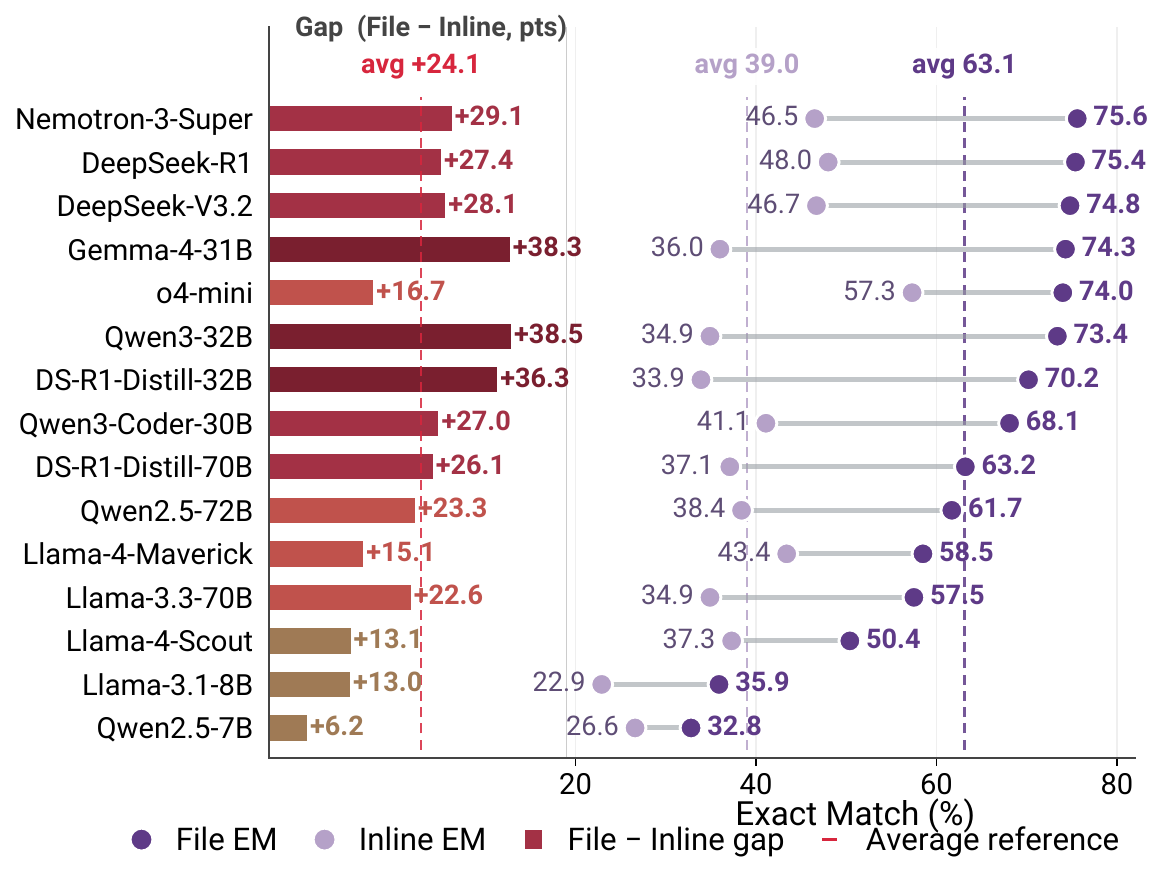}
    \vskip -1em
    \caption{
    \textbf{Coding-mode EM under file-based vs. inline graph loading, across tested LLMs.
    }
    }
\label{fig:file-vs-inline}
\vspace{-3.4\baselineskip}
\end{wrapfigure}
\textbf{\blackcircnum{4} Prompt Inline vs. Local-file Graph Loading. } 
Graph data can be loaded either directly in the prompt (\emph{inline}) or through a file path (\emph{file}). In the file-based setting, the model generates a script, such as Python code, to load the graph from disk. We evaluate the impact of file loading only under code-based reasoning, since text-based reasoning requires the graph to be included in the prompt.


\begin{table*}[t]
\centering
\caption{Each cell averages the per-model EM values across all models within that family (DeepSeek, Llama, Qwen each combine four models; Gemma, Nemotron, and OpenAI are singletons). Panels A1/A2 split coding-mode by file vs. inline graph loading; Panel B is textual-mode. The Gap column reports Explicit $-$ Implicit. \colorbox{rk1}{\textbf{bold red}} / \colorbox{rk2}{\underline{underlined yellow}} / \colorbox{rk3}{light green} mark best / second / third within each panel for Overall, Explicit, and Implicit. Detailed per-model view is in Table~\ref{tab:merged-size-combo-description}.}
\vskip -0.5em
\label{tab:merged-size-combo-description-family}
\tiny
\setlength{\tabcolsep}{2pt}
\renewcommand{\arraystretch}{0.9}
\begin{tabularx}{\linewidth}{l Y >{\columncolor{gs1}}Y >{\columncolor{gs2}}Y >{\columncolor{gs3}}Y >{\columncolor{gs4}}Y >{\columncolor{cs1}}Y >{\columncolor{cs2}}Y >{\columncolor{cs3}}Y >{\columncolor{cs4}}Y YY >{\columncolor{gapColumn}}Y}
\toprule
& \textbf{Overall}
& \multicolumn{4}{c}{\textbf{Graph size}}
& \multicolumn{4}{c}{\textbf{Combo size}}
& \multicolumn{3}{c}{\textbf{Descriptions}} \\
\cmidrule(lr){3-6}\cmidrule(lr){7-10}\cmidrule(lr){11-13}
\textbf{Family} & \textbf{EM}
& \textbf{10} & \textbf{100} & \textbf{1k} & \textbf{10k}
& \textbf{1} & \textbf{2} & \textbf{3} & \textbf{4}
& \textbf{Explicit} & \textbf{Implicit} & \textbf{Gap} \\
\midrule
\rowcolor{panelBg}
\multicolumn{13}{l}{\textit{Panel A1: Coding-mode strict EM --- file graph loading}} \\
\midrule
DeepSeek
& $70.9$ & $82.1$ & $78.2$ & $74.8$ & $48.6$ & $78.6$ & $76.0$ & $65.6$ & $54.3$ & $73.5$ & $68.3$ & $5.3$ \\
Gemma
& \second{$74.3$} & $85.6$ & $82.1$ & $78.1$ & $51.5$ & $80.0$ & $78.7$ & $71.6$ & $62.5$ & \third{$75.2$} & \second{$73.4$} & $1.8$ \\
Llama
& $50.6$ & $58.2$ & $55.5$ & $53.1$ & $35.6$ & $58.7$ & $57.6$ & $45.2$ & $33.7$ & $52.1$ & $49.1$ & $3.0$ \\
Nemotron
& \first{75.5} & $86.6$ & $83.7$ & $80.4$ & $51.5$ & $82.9$ & $79.7$ & $72.9$ & $60.6$ & \first{77.6} & \first{73.5} & $4.1$ \\
OpenAI
& \third{$74.1$} & $85.9$ & $81.9$ & $79.0$ & $49.5$ & $80.0$ & $79.4$ & $72.0$ & $59.8$ & \second{$76.1$} & \third{$72.0$} & $4.1$ \\
Qwen
& $59.1$ & $67.9$ & $64.4$ & $62.7$ & $41.1$ & $64.5$ & $65.2$ & $53.8$ & $46.7$ & $60.3$ & $57.8$ & $2.5$ \\
\cmidrule(l){2-13}
\textit{Average}
& $67.4$ & $77.7$ & $74.3$ & $71.4$ & $46.3$ & $74.1$ & $72.8$ & $63.5$ & $52.9$ & $69.1$ & $65.7$ & $3.5$ \\
\midrule
\rowcolor{panelBg}
\multicolumn{13}{l}{\textit{Panel A2: Coding-mode strict EM --- inline graph loading}} \\
\midrule
DeepSeek
& \third{$41.4$} & $78.3$ & $54.8$ & $22.2$ & $10.5$ & $48.3$ & $39.9$ & $41.3$ & $34.2$ & \third{$42.9$} & \third{$40.0$} & $2.9$ \\
Gemma
& $36.0$ & $83.5$ & $42.1$ & $11.5$ & $7.0$ & $43.6$ & $37.2$ & $33.8$ & $26.2$ & $36.5$ & $35.6$ & $0.9$ \\
Llama
& $34.7$ & $63.6$ & $44.7$ & $17.9$ & $12.5$ & $41.2$ & $37.4$ & $34.7$ & $21.4$ & $35.2$ & $34.1$ & $1.2$ \\
Nemotron
& \second{$46.5$} & $84.3$ & $55.3$ & $24.9$ & $21.4$ & $55.7$ & $44.4$ & $45.2$ & $37.7$ & \second{$47.9$} & \second{$45.0$} & $2.9$ \\
OpenAI
& \first{57.3} & $89.8$ & $69.1$ & $43.1$ & $27.2$ & $64.7$ & $58.1$ & $57.5$ & $46.0$ & \first{58.6} & \first{56.0} & $2.6$ \\
Qwen
& $35.2$ & $67.7$ & $47.9$ & $19.1$ & $6.3$ & $40.1$ & $37.6$ & $34.3$ & $25.7$ & $36.6$ & $33.9$ & $2.6$ \\
\cmidrule(l){2-13}
\textit{Average}
& $41.9$ & $77.9$ & $52.3$ & $23.1$ & $14.2$ & $48.9$ & $42.4$ & $41.1$ & $31.9$ & $43.0$ & $40.8$ & $2.2$ \\
\midrule
\rowcolor{panelBg}
\multicolumn{13}{l}{\textit{Panel B: Textual-mode strict EM}} \\
\midrule
DeepSeek
& \third{$33.9$} & $72.3$ & $38.3$ & $18.4$ & $6.8$ & $40.2$ & $37.0$ & $30.3$ & $24.0$ & \third{$34.5$} & \third{$33.3$} & $1.2$ \\
Gemma
& \second{$47.4$} & $84.8$ & $58.4$ & $29.9$ & $16.5$ & $54.4$ & $49.6$ & $43.8$ & $38.1$ & \second{$47.8$} & \second{$47.0$} & $0.8$ \\
Llama
& $18.3$ & $35.7$ & $19.9$ & $10.9$ & $6.7$ & $20.9$ & $20.0$ & $17.9$ & $12.5$ & $19.0$ & $17.6$ & $1.4$ \\
Nemotron
& $29.2$ & $70.1$ & $28.7$ & $10.0$ & $8.2$ & $35.3$ & $32.5$ & $26.3$ & $18.9$ & $30.2$ & $28.2$ & $2.0$ \\
OpenAI
& \first{48.9} & $83.5$ & $57.6$ & $34.4$ & $20.2$ & $57.8$ & $49.8$ & $46.9$ & $37.8$ & \first{50.4} & \first{47.5} & $2.9$ \\
Qwen
& $25.6$ & $53.5$ & $29.8$ & $14.9$ & $4.1$ & $28.7$ & $27.8$ & $23.8$ & $19.1$ & $26.6$ & $24.6$ & $2.0$ \\
\cmidrule(l){2-13}
\textit{Average}
& $33.9$ & $66.7$ & $38.8$ & $19.8$ & $10.4$ & $39.6$ & $36.1$ & $31.5$ & $25.1$ & $34.8$ & $33.0$ & $1.7$ \\
\bottomrule
\end{tabularx}
\vskip -2em
\end{table*}

\finding{finding:inline loading}
\begin{tcolorbox}[
    colback=blue!5, colframe=blue!20,
    boxsep=2pt, left=4pt, right=4pt, top=2pt, bottom=2pt,
    before skip=3pt, after skip=5pt
]
\small\textbf{Finding \thefinding:} 
File-based graph loading outperforms inline loading across all tested models.
\end{tcolorbox}

Figure~\ref{fig:file-vs-inline} shows all evaluated models perform better with \textit{file}-based loading than with \textit{inline} loading; additional details are provided in Table~\ref{tab:inline-file} and \S~\ref{sec:appendix_file}. Averaged across models, EM increases from $39.0$ to $63.1$.
This suggests that file-based loading substantially reduces input-parsing burden, allowing models to focus more on task modeling and algorithm implementation.

\finding{finding:inline-execution}
\begin{tcolorbox}[
    colback=blue!5, colframe=blue!20,
    boxsep=2pt, left=4pt, right=4pt, top=2pt, bottom=2pt,
    before skip=3pt, after skip=5pt
]
\small\textbf{Finding \thefinding:} 
Graph extraction is a major cause of inline reasoning failure.
\end{tcolorbox}
Table~\ref{tab:inline-file} (\S~\ref{sec:appendix_file}) shows that the average graph-extraction accuracy is only $54.5\%$, reducing average EM from $39.0\%$ under inline \texttt{solve}-only to $24.5\%$ under full inline execution, where both the generated \texttt{solve} function and graph extraction must be correct. This suggests that faithfully extracting serialized graphs imposes a burden and obscures the model's underlying algorithmic reasoning ability.

\begin{wrapfigure}[12]{r}{0.5\linewidth}
    \centering
    \vskip -1.2em
    \includegraphics[width=\linewidth]{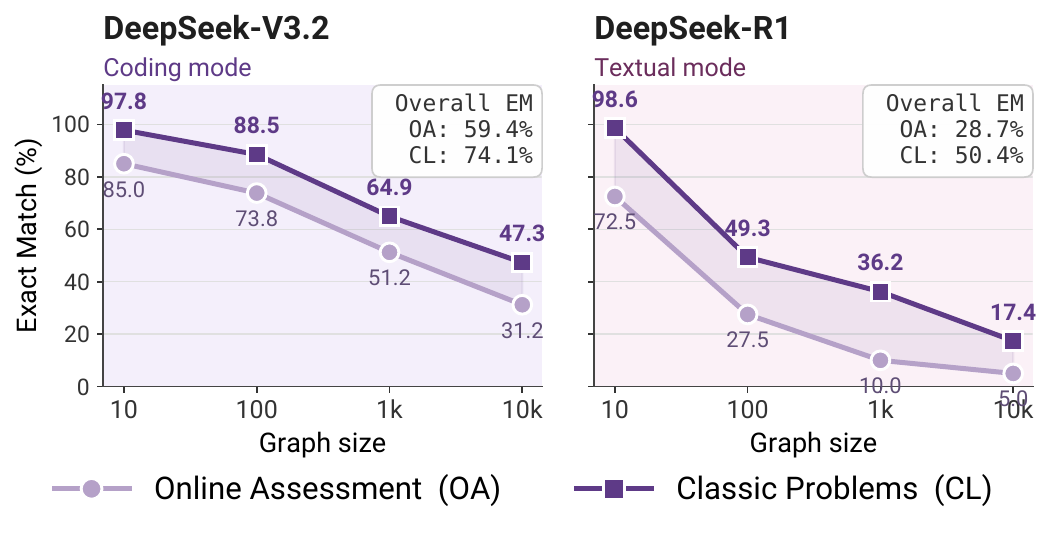}
    \vskip -1em
    \caption{
OA vs. CL EM by graph size for the top CL-perform LLM in coding (left) and text (right) modes. Insets are aggregate EM from Table~\ref{tab:leetcode-overall}.
}
\label{fig:leetcode-by-size}
\vskip -1em
\end{wrapfigure}

\textbf{\blackcircnum{5} Online Assessment vs. Classical Task Sources.}
This dimension measures the impact of shifting task sources from commonly used classical (CL) graph problems to online-assessment (OA) problems, where tasks are usually more open-ended and less likely to be covered by off-the-shelf graph libraries. We focus on combo size=$1$ tasks because their task source can be clearly identified as either CL/OA.
\finding{finding:oa-derived-tasks}
\begin{tcolorbox}[
    colback=blue!5, colframe=blue!20,
    boxsep=2pt, left=4pt, right=4pt, top=2pt, bottom=2pt,
    before skip=3pt, after skip=5pt
]

\small\textbf{Finding \thefinding: }
OA tasks are consistently more challenging than classical graph tasks across all tested models.
\end{tcolorbox}
Figure~\ref{fig:leetcode-by-size} shows that evaluated models on OA tasks consistently perform worse than on CL tasks; per-model results are provided in Table~\ref{tab:leetcode-overall} and \S~\ref{sec:appendix_leetcode}. Under code-based reasoning, the two selected models achieve EM scores of $59.4$ and $58.1$ on OA tasks, compared with $74.1$ and $80.6$ on CL tasks. Under text-based reasoning, EM similarly decreases from $50.4$ and $53.4$ on CL tasks to $28.7$ and $38.8$ on OA tasks. This consistent gap suggests that OA tasks are harder, likely because they rely less on memorized classical graph algorithms and off-the-shelf graph-library solutions.


\finding{finding:classical-tasks}
\begin{tcolorbox}[
    colback=blue!5, colframe=blue!20,
    boxsep=2pt, left=4pt, right=4pt, top=2pt, bottom=2pt,
    before skip=3pt, after skip=5pt
]
\small\textbf{Finding \thefinding:} 
CL tasks are insufficient for testing code-based graph reasoning.
\end{tcolorbox}

Fig.~\ref{fig:leetcode-by-size} shows that even at graph size $10$, strong models reach near-perfect EM on CL tasks ($97.8$) but on OA tasks with the same graph size, performance remains much lower ($72.5$--$85.0$ EM); see Table~\ref{tab:leetcode-top2-by-size} in \S~\ref{sec:appendix_leetcode}. This suggests that classical graph tasks alone, which are widely used in existing benchmarks, are insufficient because models can often invoke mature graph libraries to solve classical problems and bypass substantial reasoning. Including OA tasks mitigates this shortcut and better tests LLMs' ability to conduct open-ended graph reasoning tasks.





\subsection{RQ2: Impact of Augmentation and Adaptation}

We evaluate commonly used augmentation and adaptation methods on {\dataset}. In particular, we study the effect of domain-knowledge retrieval (RAG), instruction-tuning, and task-specific GNNs in complex graph reasoning settings. Detailed setup and expanded results are provided in \S~\ref{sec:appendix_rq_rag}-\S~ \ref{sec:appendix_gnn}. 

\textbf{\blackcircnum{1} Does Domain-Knowledge Retrieval Help {\dataset}?}
\label{sec:rq_rag}
We assess whether existing RAG methods improve model performance by comparing retrieval-augmented runs with the zero-shot CoT baseline. For coding mode, we use the GraphTeam-provided knowledge base~\cite{li2025graphteamfacilitatinglargelanguage}, which organizes NetworkX~\cite{rossi2015network} into structured entries with algorithm descriptions and examples. For text mode, we use a curated knowledge base of common graph algorithms and natural-language reasoning examples. The top-ranked retrieved entries are prepended to the model prompt. Details are provided in \S~\ref{sec:appendix_rq_rag}.

\begin{figure}[t]
    \centering
    \includegraphics[width=0.9\linewidth]{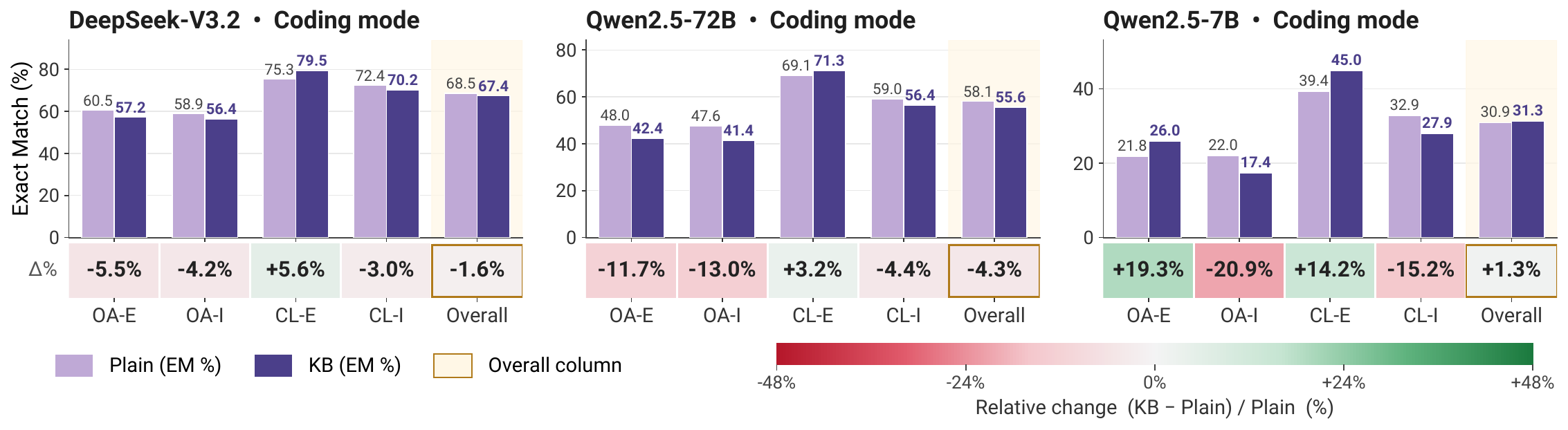}
    \vskip -0.4em
    \caption{
\textbf{KB vs. Plain EM under coding mode across LLMs and task types.} Plain denotes zero-shot CoT without retrieval. Results are reported at combo size $=1$. Task description: E = explicit, I = implicit. The bottom strip shows the relative change (red = KB hurts, green = KB helps). The ``Overall'' column aggregates across task types. Text-mode results are in Fig.~\ref{fig:rag-leetcode-merged2}.
}
    \label{fig:rag-leetcode-merged1}
    \vskip -1em
\end{figure}

\finding{finding:rag-1}
\begin{tcolorbox}[
    colback=blue!5, colframe=blue!20,
    boxsep=2pt, left=4pt, right=4pt, top=2pt, bottom=2pt,
    before skip=3pt, after skip=5pt
]
\small\textbf{Finding \thefinding:} 
Retrieval from KB improves text-mode reasoning but degrades code-inline performance.
\end{tcolorbox}

Table~\ref{tab:rag-overall} (\S~\ref{sec:appendix_rq_rag}) shows that RAG improves textual reasoning, with gains of about $+21\%$ for Qwen2.5, but hurts coding reasoning on average. Breakdowns by description and source (Figs.~\ref{fig:rag-leetcode-merged1},~\ref{fig:rag-leetcode-merged2}) show that retrieval helps coding mainly on explicit CL tasks, where relevant algorithms are easier to retrieve. In contrast, retrieved algorithms may mislead implicit tasks by anchoring models to incorrect primitives, and provide limited help for OA tasks. For textual reasoning, retrieval helps explicit tasks and implicit CL tasks, but hurts implicit OA tasks, where retrieval is less reliable. Thus, implicit OA tasks form a hard retrieval setting that requires both task modeling and open algorithm design.

\textbf{\blackcircnum{2} Can Graph Reasoning be Reasonably Learnt using Instruction-Tuning?
} 
\label{sec:rq_instruction_tuning}
We evaluate the effect of instruction tuning on {\dataset} under two reasoning modes:
(1) GCoder~\cite{zhang2024gcoderimprovinglargelanguage}, a Llama-3-8B-based code reasoning model trained on GraphWild with reinforcement learning from compiler feedback; and (2) GraphWiz~\cite{chen2024graphwizinstructionfollowinglanguagemodel}, a Llama-2-7B-based text reasoning model instruction-tuned on GraphInstruct with CoT reasoning paths and answers. Each is compared with its corresponding base.

\finding{finding:finetuning-1}
\begin{tcolorbox}[
    colback=blue!5, colframe=blue!20,
    boxsep=2pt, left=4pt, right=4pt, top=2pt, bottom=2pt,
    before skip=3pt, after skip=5pt
]
\small\textbf{Finding \thefinding:} Instruction tuning shows limited out-of-distribution generalization.
\end{tcolorbox}

\begin{figure}[t]
    \centering
    \includegraphics[width=0.92\linewidth]{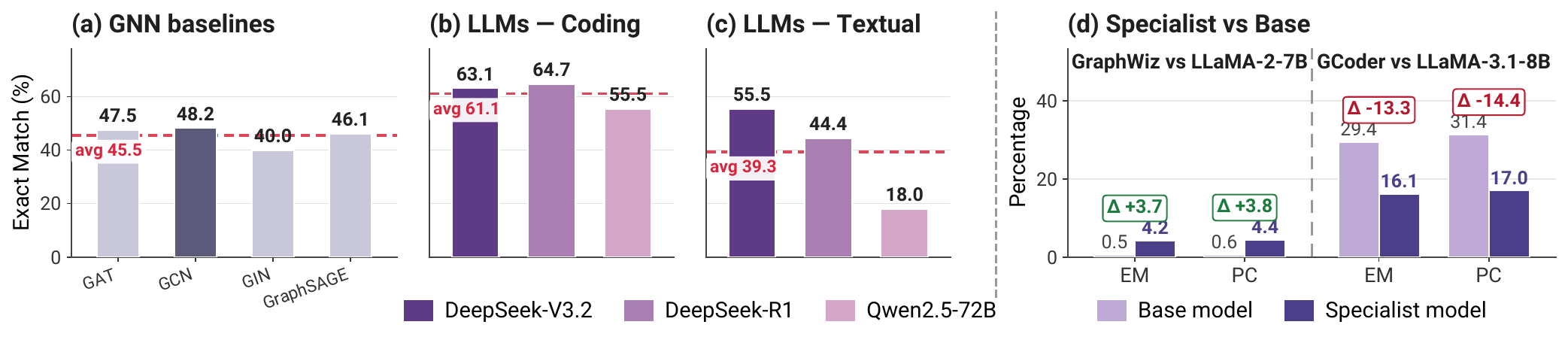}
    \vskip -1em
    \caption{
    \textbf{(a-c) EM on the 70-task GNN-eligible subset.}
The three panels compare four GNN backbones, selected LLMs in coding and textual mode, on the identical task set. 
The red dashed line indicates the panel average. 
(d) EM and PC on the instruction-tuned model vs. its base model.
}
\label{fig:gnn-vs-llms-ft-merged}
\vskip -1.3em
\end{figure}

GraphWild includes tasks similar to non-composite tasks in {\dataset}, yet GCoder fails to generalize to the distribution. In Figure~\ref{fig:gnn-vs-llms-ft-merged}(d), GCoder substantially underperforms its base model across all scenarios and graph sizes (Table~\ref{tab:gcoder-merged}, \S~\ref{sec:appendix_finetune}). 
GraphWiz shows a similar trend: it improves on CL tasks, which are closer to GraphInstruct, but does not transfer these gains to OA tasks (Table~\ref{tab:graphwiz-merged}, \S~\ref{sec:appendix_finetune}). Its gains across combo sizes are also limited to graph size $10$, whose input lengths are closest to GraphInstruct. We can conclude that the instruction tuning suffers from OOD generalization.

\textbf{\blackcircnum{3} Task-Specific GNNs vs. LLM-Based Graph Reasoning. }
\label{sec:gnn-vs-llms}
We compare task-specific GNNs and LLMs on $70$ GNN-eligible scalar-output tasks to test if specialized graph models outperform general-purpose LLM reasoning. Following GraphArena~\cite{tang2025grapharena}, we train GCN~\cite{kipf2017semisupervised}, GAT~\cite{velikov2018graph}, GIN~\cite{xu2018how}, and GraphSAGE~\cite{hamilton2017inductive} per task on generated data, and evaluate them against representative LLMs.


\finding{finding:task-specific}
\begin{tcolorbox}[
    colback=blue!5, colframe=blue!20,
    boxsep=2pt, left=4pt, right=4pt, top=2pt, bottom=2pt,
    before skip=3pt, after skip=5pt
]
\small\textbf{Finding \thefinding:} Performance ranking: LLM code-based reasoning > GNNs > LLM text-based reasoning.
\end{tcolorbox}
In Figure~\ref{fig:gnn-vs-llms-ft-merged} (a-c), the average EM is $61.1$ for LLMs under code-based reasoning, $45.5$ for GNNs, and $39.3$ for LLMs under text-based reasoning. Code-based LLM reasoning outperforms GNNs by $15.6$ percentage points. This indicates that, even on tasks whose scalar outputs are suitable for GNN modeling, generating executable Python programs and delegating computation to the interpreter provides LLMs with a clear empirical advantage over task-specific graph-supervised machine learning, while GNNs mainly serve as a stronger non-coding alternative to text-based LLM reasoning. 

\section{Conclusion}
\label{sec:conclusion}
We identify three limitations of existing graph reasoning benchmarks: limited coverage of data complexity, strong reliance on human labor, and lack of unified evaluation across text- and code-based reasoning. To address them, we propose an LLM-based semi-automatic dataset construction framework that covers five complexity dimensions: \textit{graph size}, \textit{composition depth}, \textit{task description}, \textit{graph loading}, and \textit{task source}. The framework uses limited human validation to ensure data quality and designs corresponding formats for both reasoning modes. Based on this framework, we construct {\dataset} and systematically study how data complexity affects LLM graph reasoning and existing augmentation methods. Our results reveal overlooked complexity factors and expose limitations of existing methods, offering a new testbed and empirical guidance for future research.

\textbf{Limitations.} Although {\dataset} is built through a semi-automated framework, its effectiveness depends on the backbone LLM’s capability to reliably generate and solve graph reasoning problems. Thus, standard local LLMs may be insufficient, and stronger close-source LLMs may be needed, though at modest token cost.
In addition, {\dataset} is not specifically tailored for fine-tuning; its current scope is mainly limited to evaluation and supporting the generation of new benchmark data.

\bibliographystyle{plainnat}
\bibliography{reference}


\appendix
\newpage



\section{Prompt-length Analysis for Inline Graph Serialization}
\label{sec:prompt-length}

We quantify how the size of an inline graph encoding scales with the number
of nodes $n$ and the underlying graph distribution, measured in the same
units an LLM actually consumes: \emph{tokens}. The analysis lets us identify,
for a given context window, the largest graph that fits without truncation.

\subsection{Prompt template}
\label{subsec:prompt-template}
Each task in our dataset is represented in two task description formats: explicit and implicit. For convenience, all token-length analyses are based on the explicit form. 
\begin{verbatim}
<task preamble>
Input: <parameters>
Nodes: [0, 1, 2, ..., N-1]
Edges: [[u, v]/[u, v, w], ...]
<task-specific question>
<instruction>
\end{verbatim}
Let $T_{\text{prompt}}(N, M)$ denote the token length of such a prompt for
a graph with $N$ nodes and $M$ edges. We decompose it as
\begin{equation}
  T_{\text{prompt}}(N, M)
    \;=\;
    T_{\text{fixed}}
    \;+\;
    T_{\text{nodes}}(N)
    \;+\;
    T_{\text{edges}}(M, N, \text{weighted}),
  \label{eq:decomp}
\end{equation}
where $T_{\text{fixed}}$ is the token count of the preamble, instruction
boilerplate, scalar-input lines, and section labels (\texttt{Nodes:},
\texttt{Edges:}); $T_{\text{nodes}}(N)$ is the tokens of rendering the node
list; and $T_{\text{edges}}(M, N, \text{weighted})$ is the tokens of rendering
the edge list.

\subsection{Tokenization}
\label{subsec:tokenization}

We report token counts from \emph{actual tokenization}. By default, we use OpenAI's \texttt{cl100k\_base} \cite{openai2022tiktoken} encoding
(GPT-4 / GPT-4o family).
Concretely, for a tokenizer $\phi$ and a string $s$, let $|s|_\phi$ denote
$|\phi(s)|$, the number of tokens $\phi$ emits for $s$. We compute each
term of~\eqref{eq:decomp} as follows:
\begin{align}
  T_{\text{fixed}}      &= \bigl|\, s_{\text{boilerplate}} \,\bigr|_\phi,
      \label{eq:tfixed}\\
  T_{\text{nodes}}(N)   &= \bigl|\, \texttt{str(list(range(N)))} \,\bigr|_\phi,
      \label{eq:tnodes}\\
  T_{\text{edges}}(M, N, \text{weighted})
      &= \bigl|\, \texttt{str(}E\texttt{)} \,\bigr|_\phi,
      \label{eq:tedges}
\end{align}
where $E$ is a list of $M$ edge records drawn uniformly at random from
$\{0,\dots,N-1\}^2$ (with independent weights $w \in \{1,\dots,10\}$ if
\texttt{weighted}$=\text{true}$).
reproducibility.


\subsection{Edge counts by graph distribution}
\label{subsec:edge-counts}

The only term in~\eqref{eq:decomp} that depends on the generative model is
$M$. For a target graph distribution $\mathcal{D}$ on $N$ nodes, we use the
expected edge count $M_{\mathcal{D}}(N)$:
\begin{equation}
  M_{\mathcal{D}}(N) \;=\;
  \begin{cases}
    N - 1                                 & \mathcal{D} = \text{tree}, \\
    N                                     & \mathcal{D} = \text{sparse}, \\
    2N                                    & \mathcal{D} = \text{medium}, \\
    m\,(N - m)                            & \mathcal{D} = \text{BA}(m),
                                            \text{ (Barab\'asi--Albert)} \\
    \tfrac{1}{2}\, k\, N                  & \mathcal{D} = \text{RGG}(k),
                                            \text{ (random geometric,}\ \bar d = k)\\
    \tfrac{1}{2}\, p\, N(N-1)             & \mathcal{D} = \text{ER}(p),
                                            \text{ (Erd\H{o}s--R\'enyi)} \\
    \tfrac{1}{8}\, N(N-1)                 & \mathcal{D} = \text{dense}\
                                            (\text{ER with }p=0.25), \\
    \tfrac{1}{2}\, N(N-1)                 & \mathcal{D} = \text{complete}.
  \end{cases}
  \label{eq:edge-counts}
\end{equation}
Substituting $M_{\mathcal{D}}(N)$ into~\eqref{eq:tedges} and then
into~\eqref{eq:decomp} yields the distribution-specific token cost
$T_{\mathcal{D}}(N)$.

\subsection{Results}
\label{subsec:length-results}

\begin{figure}[t]
  \centering
  \includegraphics[width=0.99\linewidth]{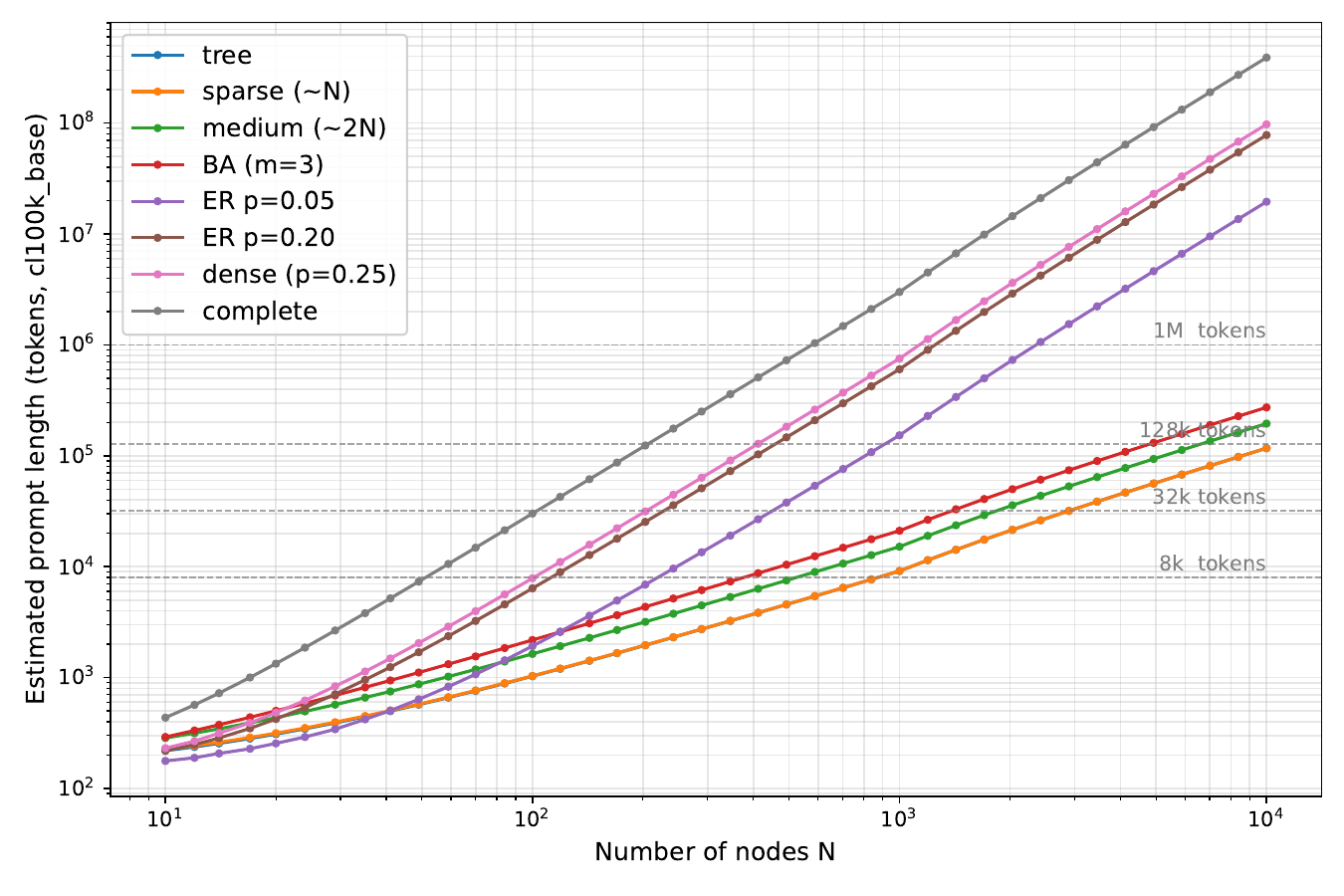}
  \caption{Estimated prompt length in tokens vs.\ number of nodes
    $N$ for eight graph distributions (\texttt{cl100k\_base} encoding) based on unweighted edges.
    Dashed lines mark LLM context windows.}
  \label{fig:prompt-length}
\end{figure}

Figure~\ref{fig:prompt-length} plots $T_{\mathcal{D}}(N)$ against $N$ on
log--log axes for eight graph distributions, with $T_{\text{fixed}} = 135$ and horizontal reference
lines marking common LLM context windows (8k, 32k, 128k, 1M tokens).
Key numerical snapshots are reproduced in
Table~\ref{tab:length-snapshot}.

\begin{table}[t]
  \centering
  \caption{Estimated total prompt length in tokens
    ($T_{\text{prompt}} = T_{\text{fixed}} + T_{\text{nodes}} +
    T_{\text{edges}}$) for unweighted edges and tokenizer
    \texttt{cl100k\_base}. $T_{\text{fixed}} = 135$. Weighted edges add roughly $30\%$ per cell.}
  \small
  \begin{tabular}{lrrrrrrrr}
    \toprule
    $N$ & tree & sparse & medium & BA($m\!=\!3$) & ER($0.05$) & ER($0.20$) & dense & complete \\
    \midrule
    10     & 219       & 225       & 285       & 291       & 177       & 219       & 231       & 435 \\
    100    & 1{,}029   & 1{,}035   & 1{,}635   & 2{,}181   & 1{,}923   & 6{,}375   & 7{,}863   & 30{,}134 \\
    1000   & 9{,}129   & 9{,}135   & 15{,}135  & 21{,}081  & 152{,}978 & 602{,}503 & 752{,}344 & 2{,}999{,}969 \\
    10000  & 117{,}105 & 117{,}113 & 195{,}089 & 272{,}996 & 19.5\,M   & 78.0\,M   & 97.5\,M   & 389.9\,M \\
    \bottomrule
  \end{tabular}
  \label{tab:length-snapshot}
\end{table}

\begin{takeawaybox}
    
Two observations follow immediately from Figure~\ref{fig:prompt-length}
and Table~\ref{tab:length-snapshot}.
\begin{itemize}[noitemsep,leftmargin=*]
  \item \textbf{Sparse distributions scale linearly.} For tree, sparse
    ($M \!\approx\! N$), medium ($M \!\approx\! 2N$), and BA($m\!=\!3$),
    $T_{\text{prompt}}$ grows as $\Theta(N \log N)$ (the $\log N$ comes
    from per-integer digit width), so a $128$k-token window fits
    $N \approx 6$--$7 \cdot 10^{3}$ nodes.
  \item \textbf{Dense distributions scale quadratically.} ER($p$), dense,
    and complete grow as $\Theta(p N^{2} \log N)$; the complete graph
    saturates the $128$k window already at $N \approx 200$, and the $1$M
    window at $N \approx 600$.
\end{itemize}

\end{takeawaybox}

\section{Benchmark Scoring Criteria and Comparative Results}
\label{sec:benchmark-scoring}

\paragraph{Scoring scheme.}
To place our benchmark in the broader landscape of graph-reasoning evaluations, we introduce a five-dimensional rubric that quantifies the task and data complexity exercised by each benchmark. Each dimension is normalized to a common $[0, 100]$ scale, so that per-dimension scores and their sum are directly comparable across benchmarks.

\begin{enumerate}[nosep, leftmargin=*]
    \item \textbf{Graph size.} This dimension evaluates whether a benchmark can systematically vary input length and thereby assess LLM performance under different long-context settings. In practice, graphs with around 100 nodes are often still manageable within the context window of modern LLMs, but they are usually insufficient for meaningful pressure testing. We therefore assign higher scores to benchmarks that include graph instances at the 1k- and 10k-node levels. Following the intuition that each order-of-magnitude increase in graph size introduces a comparable increase in difficulty, we score the maximum node count $n_{\max}$ on a logarithmic scale:
    \[
    \text{NodeScaleScore} \;=\; 100 \times \frac{\log_{10}\bigl(\min(n_{\max},\,10{,}000)\bigr)}{\log_{10}(10{,}000)}.
    \]
    Under this definition, maximum node sizes of $10$, $100$, $1{,}000$, and $10{,}000$ correspond to scores of $25$, $50$, $75$, and $100$, respectively.

    \item \textbf{Task complexity.} This dimension measures whether a benchmark covers graph reasoning problems with increasing levels of difficulty. Benchmarks restricted to elementary single-task polynomial-time (P) problems receive $33$ points; those that include more challenging single-task NP-hard problems receive $66$ points; and those that support composite tasks, for example by chaining multiple classical graph procedures, receive $100$ points.

    \item \textbf{Task description.} This dimension evaluates whether a benchmark tests only direct graph problem solving, or also requires the model to infer the underlying graph formulation from more realistic descriptions. Benchmarks that describe tasks exclusively in explicit graph-theoretic terminology receive $50$ points, whereas those that additionally provide implicit, narrative-style problem formulations receive $100$ points.

    \item \textbf{Graph loading.} This dimension evaluates whether a benchmark supports graph reasoning beyond the setting where the full graph is directly embedded in the prompt. Benchmarks that provide graph data only within the prompt receive $50$ points. Benchmarks that additionally support loading graphs from external files, thereby enabling instances larger than the model's context window and requiring more general code solutions, receive $100$ points.

    \item \textbf{Task source.} This dimension evaluates the diversity of task origins. Benchmarks drawing tasks only from classical graph algorithms receive $33$ points, since such tasks may be less challenging for code reasoning when existing graph libraries can be directly applied. Benchmarks drawing tasks only from online-assessment (OA) style problems receive $66$ points, and those combining both classical and OA-style sources receive $100$ points.
\end{enumerate}

\begin{figure}[h]
    \centering
    \includegraphics[width=\linewidth]{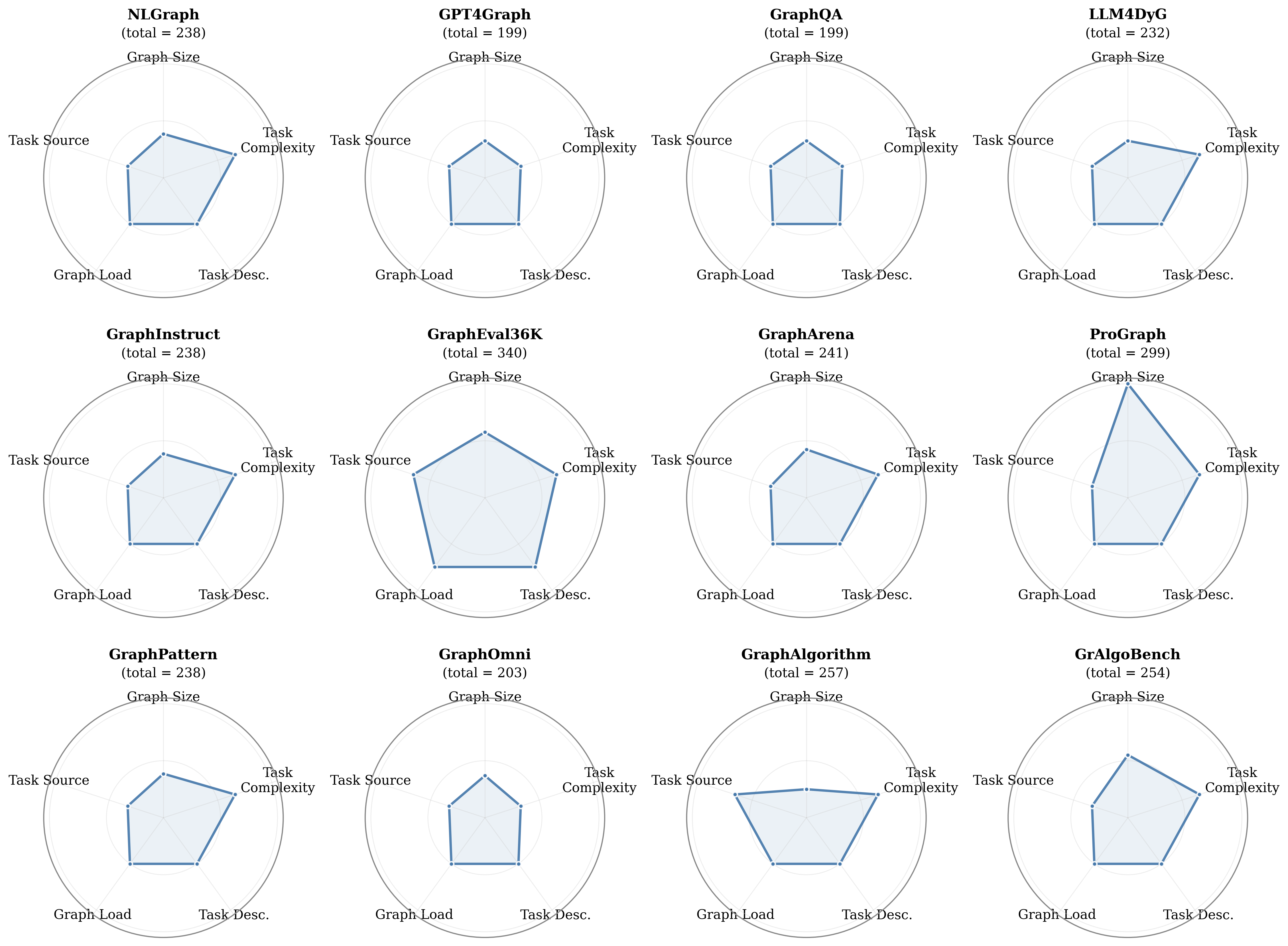}
    \caption{Per-benchmark score profiles under the proposed five-dimensional rubric (\emph{Graph size}, \emph{Task complexity}, \emph{Task description}, \emph{Graph loading}, and \emph{Task source}). Each radar displays the five per-dimension scores on a common $[0,100]$ scale; the total score of each benchmark is reported.}
    \label{fig:radar_benchmarks_multi}
\end{figure}

\paragraph{Comparative results.}
Figure~\ref{fig:radar_benchmarks_multi} applies this rubric to a representative collection of graph-reasoning benchmarks: NLGraph~\cite{wang2024languagemodelssolvegraph}, GPT4Graph~\cite{guo2023gpt4graph}, GraphQA~\cite{fatemi2023talklikegraphencoding}, LLM4DyG~\cite{zhang2024llm4dyg}, GraphInstruct~\cite{luo2024graphinstruct}, GraphEval36K~\cite{wu2025grapheval36k}, GraphArena~\cite{tang2025grapharena}, ProGraph~\cite{li2024can}, GraphPattern~\cite{dai2025how}, GraphOmni~\cite{xu2026graphomni}, GraphAlgorithm~\cite{hu2025rethinking}, and GrAlgoBench~\cite{zhang2026exposing}, alongside our benchmark. Two trends emerge. First, existing benchmarks tend to excel along only a narrow subset of the five dimensions, with most clustering at low task complexity, explicit-only descriptions, and inline-only graph loading (graph in prompts). Second, no prior benchmark simultaneously attains high scores on all dimensions, leaving a gap that our benchmark is designed to close: it offers composite tasks, both explicit and implicit descriptions, both inline and file-based graph loading, mixed task sources, and graph sizes up to $n_{\max} = 10{,}000$, achieving the maximum score on every axis.

\begin{table*}[h]
\centering
\scriptsize
\setlength{\tabcolsep}{4pt}
\caption{Meta information for representative graph reasoning benchmarks, organized first by reasoning mode (textual reasoning vs. coding reasoning) and then by goal. \#Tasks denotes the number of tasks; the exact task lists are provided in Table~\ref{tab:graph_benchmark_tasks}. ``Task Source'' indicates where the benchmark tasks are drawn from, such as classical graph problems, manually designed specific tasks, online assessment (OA) platforms, or graph libraries (usually from classical). ``Graph Loading (GL)'' indicates whether graph data are provided in the prompt, in an external file, or both. ``Task Description (TD)'' indicates whether task descriptions are explicit, implicit, or both. ``Task Complexity (TC)'' indicates whether the benchmark covers polynomial-time (P) tasks, NP-hard tasks, or composite tasks formed by combining multiple classical graph problems. ``Scalability'' indicates whether the benchmark explicitly supports automatic expansion to new tasks beyond a fixed task set.}
\begin{tabularx}{\linewidth}{X r c X X X X r c}

\toprule
\textbf{Benchmark} & \textbf{\#Tasks} & \textbf{Goal} & \textbf{Task Source} & \textbf{GL} & \textbf{Graph Size} & \textbf{TD} & \textbf{TC} & \textbf{Scalability} \\
\midrule
\multicolumn{9}{c}{\textit{Textual reasoning --- Evaluation}} \\
NLGraph~\cite{wang2024languagemodelssolvegraph} & 8 & Evaluation & Classical & In prompt & $\leq 35$ & Explicit & P, NP & No \\
GPT4Graph~\cite{guo2023gpt4graph} & 10 & Evaluation & Classical & In prompt & $\approx 10$--$20$ & Explicit & P & No \\
GraphQA~\cite{fatemi2023talklikegraphencoding} & 7 & Evaluation & Classical & In prompt & 5--20 & Explicit & P & No \\
LLM4DyG~\cite{zhang2024llm4dyg} & 9 & Evaluation & Specific & In prompt & 5--20 & Explicit & N/A & No \\
GraphArena~\cite{tang2025grapharena} & 10 & Evaluation & Classical & In prompt & 4--50 & Explicit & P, NP & No \\
GraphPattern~\cite{dai2025how} & 11 & Evaluation & Classical & In prompt & 5--35 & Explicit & N/A & No \\
GraphOmni~\cite{xu2026graphomni} & 6 & Evaluation & Classical & In prompt & 5--30 & Explicit & P & No \\
GrAlgoBench~\cite{zhang2026exposing} & 9 & Evaluation & Classical & In prompt & 8--160 & Explicit & P, NP & No \\
\midrule
\multicolumn{9}{c}{\textit{Textual reasoning --- Training}} \\
GraphWiz~\cite{chen2024graphwizinstructionfollowinglanguagemodel} & 9 & Training & Classical & In prompt & 2--100 & Explicit & P, NP & No \\
GraphInstruct~\cite{luo2024graphinstruct} & 21 & Training & Classical & In prompt & 5--35 & Explicit & P, NP & No \\
NLGift~\cite{zhang2024can} & 4 & Training & Classical & In prompt & 3--25 & Explicit & P & No \\
GraphSilo~\cite{peng2025rewarding} & 13 & Training & Classical & In prompt & 5--35 & Explicit & P & No \\
\midrule
\multicolumn{9}{c}{\textit{Coding reasoning --- Evaluation}} \\
GraphEval36K~\cite{wu2025grapheval36k} & 40 & Evaluation & LeetCode & In file & 20--200 & Both & P, NP & No \\
GraphAlgorithm~\cite{hu2025rethinking} & 239 & Evaluation & \makecell[tl]{Codeforces,\\ AtCoder,\\ CodeChef,\\ Kattis} & In prompt & $\sim$10 & Both & P, NP & No \\
ProGraph~\cite{li2024can} & 512 & Evaluation & \makecell[tl]{NetworkX,\\ igraph,\\ CDlib,\\ graspologic,\\ little ball of fur,\\ Karate Club} & In prompt & up to $10^6$ & Explicit & P, NP & No \\
\midrule
\multicolumn{9}{c}{\textit{Coding reasoning --- Training}} \\
GraphWild~\cite{zhang2024gcoderimprovinglargelanguage} & 313 & Training & NetworkX & In prompt & $\sim$10 & Both & P, NP & No \\
GTools~\cite{wang2025GraphToolInstruction} & 11 & Training & Classical & Both & 2--1000 & Explicit & P, NP & No \\
\bottomrule
\end{tabularx}
\label{tab:graph_benchmark_meta}
\end{table*}


\begin{table*}[h]
\centering
\scriptsize
\setlength{\tabcolsep}{4pt}
\caption{Exact task lists or task-collection descriptions for representative graph reasoning benchmarks.}
\begin{tabularx}{\linewidth}{l X}
\toprule
\textbf{Benchmark} & \textbf{Exact task list} \\
\midrule
\multicolumn{2}{c}{\textit{Textual reasoning --- Evaluation}} \\
NLGraph~\cite{wang2024languagemodelssolvegraph} &
Connectivity; Cycle; Topological Sort; Shortest Path; Maximum Flow; Bipartite Graph Matching; Hamilton Path; GNN. \\

GPT4Graph~\cite{guo2023gpt4graph} &
Structure understanding: Size Detection; Degree Detection; Edge Detection; Attribute Retrieval; Diameter Computing; Clustering Coefficient Computing.
Semantic understanding: KGQA; GQL Generation; Node Classification; Graph Classification. \\

GraphQA~\cite{fatemi2023talklikegraphencoding} &
Edge Existence; Node Degree; Node Count; Edge Count; Connected Nodes; Cycle Check; Disconnected Nodes. \\

LLM4DyG~\cite{zhang2024llm4dyg} &
Temporal: when link; when connect; when triadic closure.
Spatial: what neighbors at time; what neighbors in periods; check triadic closure.
Spatial-temporal: check temporal path; find temporal path; sort edge by time. \\

GraphArena~\cite{tang2025grapharena} &
P: Common Neighbor; Shortest Distance; Connected Component; Graph Diameter.
NP: Maximum Clique Problem; Maximum Independent Set; Minimum Vertex Cover; Maximum Common Subgraph; Graph Edit Distance; Traveling Salesman Problem. \\

GraphPattern~\cite{dai2025how} &
Term-based: pattern translation; graph modification; pattern detection; graph generation.
Topo-based: graph isomorphic mapping; graph modification; pattern detection.
Data-driven: dense subgraph mining; frequent subgraph extraction; discriminative pattern learning.
Real-world: molecule pattern detection; discriminative pattern learning and classification. \\

GraphOmni~\cite{xu2026graphomni} &
Connectivity; Cycle Detection; Diameter; BFS Order; Triangle Counting; Shortest Path. \\

GrAlgoBench~\cite{zhang2026exposing} &
Maximum Degree Node; Maximum Weight Triangle; Maximum Clique Problem; PathSum; Distance-k; Diameter; Maximum k-core; Minimum Spanning Tree; Distance Threshold. \\

\midrule
\multicolumn{2}{c}{\textit{Textual reasoning --- Training}} \\
GraphWiz~\cite{chen2024graphwizinstructionfollowinglanguagemodel} &
Cycle Detection; Connectivity; Bipartite Check; Topological Sort; Shortest Path; Maximum Triangle Sum; Maximum Flow; Hamilton Path; Subgraph Matching. \\

GraphInstruct~\cite{luo2024graphinstruct} &
Neighbor; Degree; Predecessor; PageRank; Clustering Coefficient; Jaccard; Common Neighbor; Edge; Shortest Path; Connectivity; Maximum Flow; DFS; BFS; Cycle; Connected Component; Diameter; Bipartite; Topological Sort; MST; Euler Path; Hamiltonian Path. \\

NLGift~\cite{zhang2024can} &
Connectivity; Shortest Path; Topological Sort; Maximum Flow. \\

GraphSilo~\cite{peng2025rewarding} &
Node-level: Degree; Clustering Coefficient; Neighbor; PageRank; Predecessor.
Node-pair-level: Jaccard; Common Neighbor; Connectivity; Maximum Flow.
Graph-level: Breadth First Search; Cycle; Diameter; MST. \\

\midrule
\multicolumn{2}{c}{\textit{Coding reasoning --- Evaluation}} \\
GraphEval36K~\cite{wu2025grapheval36k} &
40 LeetCode graph problems. \\

ProGraph~\cite{li2024can} &
Basic Graph Theory; Graph Statistical Learning; Graph Embedding. \\

GraphAlgorithm~\cite{hu2025rethinking} &
239 graph problems from OA platforms. \\

\midrule
\multicolumn{2}{c}{\textit{Coding reasoning --- Training}} \\
GraphWild~\cite{zhang2024gcoderimprovinglargelanguage} &
313 graph coding problems derived from NetworkX algorithms. \\

GTools~\cite{wang2025GraphToolInstruction} &
Cycle Detection; Triangle; Edge Count; Node Count; Topo; Degree Count; Edge Existence; Node Existence; Maximum Flow; Path Existence; Shortest Path. \\
\bottomrule
\end{tabularx}
\label{tab:graph_benchmark_tasks}
\end{table*}

\section{Detailed Related Work}

\subsection{Benchmarks for LLM-based Graph Reasoning}
\label{sec:appendix_benchmarks4graph}
Graph reasoning tasks are widely considered an effective testbed for evaluating the reasoning capabilities of large language models (LLMs). They offer controllable difficulty, can be generated automatically at scale, and are less likely than standard reasoning benchmarks to overlap with public pretraining corpora, thereby partially alleviating data contamination concerns. Existing benchmarks for graph reasoning can be broadly grouped into two categories: those built from classical graph tasks or graph algorithmic problems \cite{wang2024languagemodelssolvegraph, guo2023gpt4graph, liu2023evaluating, fatemi2023talklikegraphencoding, tang2025grapharena, li2024can, xu2026graphomni, zhang2026exposing}, and those built from graph programming problems collected from online assessment (OA) platforms \cite{wu2025grapheval36k, hu2025rethinking}.

\paragraph{(i) Benchmarks based on classical graph tasks.}
This line of work mainly draws on classical graph problems studied in algorithm textbooks and the broader algorithms literature, such as connectivity, cycle detection, topological sorting, shortest paths, and the traveling salesman problem. We summarize the tasks covered by existing work in Tables~\ref{tab:graph_benchmark_meta} and \ref{tab:graph_benchmark_tasks}.

\textbf{Evaluation benchmarks.}
NLGraph \cite{wang2024languagemodelssolvegraph} is among the earliest studies to systematically examine graph reasoning in LLMs. It selects eight tasks, such as connectivity, cycle detection, and topological sorting, and divides instances into easy, medium, and hard levels based on graph size and sparsity. Graph instances and ground-truth answers are generated by manually implemented programs, and the models are evaluated under different prompting strategies. GPT4Graph \cite{guo2023gpt4graph} contains 10 tasks, including 6 structural understanding tasks, such as edge existence, and 4 semantic understanding tasks, such as node classification. Its structural tasks are built from small subgraphs sampled from ogbn-arxiv \cite{hu2020open} and Aminer \cite{tang2008arnetminer}, with answers produced by manually written programs. LLMtoGraph \cite{liu2023evaluating} also considers basic classical graph problems, but introduces a more fine-grained evaluation protocol. Beyond final-answer accuracy, it includes \emph{comprehension}, \emph{correctness}, \emph{fidelity}, and \emph{rectification} to assess reasoning quality, answer accuracy, answer completeness in multi-answer settings, and self-correction ability. GraphQA \cite{fatemi2023talklikegraphencoding} covers 12 basic tasks, such as node degree and triangle counting, using programmatically generated random graphs and manually implemented algorithms to compute ground-truth answers. GraphArena \cite{tang2025grapharena} differs from prior work by emphasizing real-world graphs, including DBLP \cite{ley2002dblp}, DBpedia \cite{bizer2009dbpedia}, OpenFlights, and PubChemQC \cite{nakata2017pubchemqc}. It covers four classes of P problems and six classes of NP problems, and uses automated scripts to verify intermediate results, reducing the chance of correct answers by guessing. ProGraph \cite{li2024can} evaluates whether LLMs can solve classical graph tasks by writing programs with the help of graph analysis documentation and toolkits. It covers graph-theoretic, graph statistical learning, and graph embedding tasks, and contains 512 human-annotated QA pairs. GraphOmni \cite{xu2026graphomni} systematically studies textual graph reasoning along three dimensions: graph type, serialization format, and prompting mode, and provides detailed error analysis. GrAlgoBench \cite{zhang2026exposing} targets large reasoning models and constructs nine tasks from real-world graphs such as DBLP \cite{ley2002dblp}, Street Network \cite{boeing2025modeling}, OpenFlights \cite{openflights}, Wikipedia \cite{yin2017local}, and DBpedia \cite{bizer2009dbpedia}, organizing them into three categories: Enumeration, Exploration, and Intuition.

\textbf{Generalization after fine-tuning.}
NLGift \cite{zhang2024can} studies whether capabilities learned from in-distribution graph reasoning data transfer to out-of-distribution (OOD) settings. It fine-tunes models on four classical tasks, connectivity, shortest path, topological sort, and maximum flow, using programmatically generated graph instances and ground-truth answers, and evaluates them under five types of shift: semantic, numerical, structural, reasoning, and real-world. The results show moderate transfer under semantic, numerical, and structural shifts, but weak generalization under reasoning and real-world shifts.

Overall, this line of work mainly focuses on classical graph tasks, covering both capability evaluation and post-fine-tuning generalization. However, for many such tasks, models can often solve the problem by directly calling existing functions in Python graph libraries, which limits the challenge posed to more complex code-based reasoning.

\paragraph{(ii) Benchmarks based on graph problems from online assessment platforms.}
Another line of work collects more complex, engineering-oriented graph problems from online assessment platforms to evaluate LLMs' code reasoning and code generation. These benchmarks are closer to competitive programming and practical algorithm training than to textbook-style graph tasks. GraphEval36K \cite{wu2025grapheval36k} collects 40 graph problems from LeetCode \cite{leetcode_graph}. Each problem includes a statement, examples, constraints, and a code skeleton. The benchmark uses NetworkX to generate test cases and manually written reference solutions to produce ground-truth outputs. Evaluation is based on whether the generated code passes the test cases. GraphAlgorithm \cite{hu2025rethinking} further expands this direction by collecting 239 graph problems from Codeforces, AtCoder, CodeChef, and Kattis.

\paragraph{(iii) Benchmarks for special graph settings.}
Beyond the two main directions above, some work studies more specialized graph reasoning settings. LLM4DyG \cite{zhang2024llm4dyg} focuses on dynamic graph reasoning with temporal and spatial information, where answers are first determined programmatically and questions are then generated from templates. GraphPattern \cite{dai2025how} evaluates whether LLMs understand graph patterns through tasks such as generating a graph with exactly one triangle, as well as graph modification, pattern detection, isomorphic mapping, dense subgraph mining, frequent subgraph extraction, and discriminative pattern learning. Its graph instances and question templates are generated programmatically, and answers are automatically derived from the construction constraints.

\paragraph{Limitations and our motivation.}
Despite substantial progress, three key limitations remain. 
First, existing benchmarks typically rely on classical graph tasks to evaluate textual reasoning, while using OA-style graph problems to assess code reasoning, leaving the field without a unified benchmark that supports both paradigms. 
Second, most existing benchmarks are built on predefined tasks and require human involvement in deriving ground-truth solutions or implementing task-specific programs for labeling and evaluation, which limits their scalability to new graph tasks. 
Third, task difficulty is still characterized rather coarsely, without systematically covering larger graph scales, compositional graph reasoning, or implicit problem descriptions. 
To address these limitations, we propose a semi-automatic graph problem construction and evaluation framework that can generate graph reasoning tasks with diverse difficulty levels and provide automated evaluation for both textual reasoning models and code-based reasoning models.

\subsection{Methods for LLM-based Graph Reasoning}
\label{sec:appendix_relatedwork_method}
LLMs still perform unsatisfactorily on graph reasoning tasks, so recent studies have increasingly focused on improving their graph reasoning capabilities. From the perspective of final reasoning execution, existing methods can be broadly divided into two categories: text-based graph reasoning \cite{fatemi2023talklikegraphencoding, wang2024languagemodelssolvegraph, chen2024graphwizinstructionfollowinglanguagemodel, luo2024graphinstruct} and code-based graph reasoning \cite{cai2024codegraphenhancinggraphreasoning, zhang2024gcoderimprovinglargelanguage, wang2025GraphToolInstruction, li2025graphteamfacilitatinglargelanguage}. The former solves graph tasks through natural language reasoning over graph structures serialized as text in the prompt, whereas the latter realizes graph algorithmic computation by generating and executing code.

\paragraph{(i) Text-based graph reasoning.}
Text-based graph reasoning treats graph problems as natural language reasoning tasks by serializing graph structures into textual inputs and requiring LLMs to infer answers directly from the prompt. Existing efforts mainly improve this paradigm through better prompting strategies or fine-tuning methods that strengthen the model's ability to perform structured reasoning over textualized graphs.

\textbf{Prompting-based reasoning enhancement.}
This paradigm relies on LLMs' ability to parse and reason over textualized graph structures, and typically improves performance through carefully designed prompting strategies. For example, Talk-like-a-Graph \cite{fatemi2023talklikegraphencoding} and NLGraph \cite{wang2024languagemodelssolvegraph} systematically evaluate the effectiveness of standard prompting methods for graph reasoning tasks, including zero-shot, few-shot, chain-of-thought \cite{wei2022chain}, and self-consistency \cite{wang2023selfconsistency}. Building on this, NLGraph further proposes Build-a-Graph, which explicitly requires the model to output a conceptual representation of the graph to enhance structural awareness, as well as Algorithmic Prompting, which injects algorithm-level hints into prompts to facilitate reasoning. Similarly, Simple-RTC (Simple-Reasoning-Then-Coding) \cite{hu2025rethinking} adopts a two-stage ``reason-then-code'' paradigm: it first guides the model to analyze the problem and design an algorithm, and then delegates the precise, repetitive, and error-prone execution steps to code, thereby reducing the burden of purely natural language reasoning.

\textbf{Enhancing text-based reasoning through fine-tuning.}
Purely prompting-based methods remain limited by LLMs' insufficient capacity for deep algorithmic understanding and structured derivation, and their overall performance is therefore often unsatisfactory. To mitigate these limitations, several works further introduce instruction tuning to strengthen graph reasoning ability. For example, GraphWiz \cite{chen2024graphwizinstructionfollowinglanguagemodel} uses instruction data optimized by direct preference optimization to guide models toward more suitable reasoning paths for graph computation; GraphInstruct \cite{luo2024graphinstruct} proposes GraphLM, which applies LoRA fine-tuning with final-answer prediction as the training objective to improve graph understanding; it further introduces GraphLM+, which jointly optimizes key tokens in multi-step reasoning and the final answer to further improve graph reasoning performance. Another representative work is GraphSilo \cite{peng2025rewarding}, which constructs a training dataset with step-wise labels to train a process reward model, GraphPRM. GraphPRM scores candidate reasoning processes and supports two reasoning enhancement strategies: \emph{Best-of-N}, which first samples $N$ complete candidate solutions and then selects the best one using GraphPRM; and \emph{Guided Beam Search}, which generates multiple candidates at each step and incrementally scores and filters them with GraphPRM to guide the search trajectory. In addition, GraphPRM can be further used to construct DPO preference pairs, which in turn support preference optimization fine-tuning of LLMs.
Nevertheless, prior work shows that instruction tuning still exhibits limited transferability on graph reasoning tasks \cite{zhang2024can}. Achieving robust generalization across different graph tasks remains a major challenge in this line of research.

\paragraph{(ii) Code-based graph reasoning.}
Code-based graph reasoning treats graph problems as program synthesis and execution tasks, where LLMs solve graph reasoning problems by generating executable code rather than relying solely on natural language inference over serialized graphs. Existing methods in this paradigm mainly improve performance through direct code generation, retrieval-augmented coding with external documentation or prior solutions, and fine-tuning on graph-specific coding data.

\textbf{Direct code generation for graph computation.}
This paradigm equips LLMs with graph algorithmic capability through code generation and execution, without relying on explicit step-by-step natural language reasoning. For example, CodeGraph \cite{cai2024codegraphenhancinggraphreasoning} uses a small number of examples to guide models to generate graph algorithm code, and then executes the code with a program interpreter to obtain the final answer. PIE \cite{gong2025pseudocodeinjectionmagicenablingllms}, in contrast, injects expert-level pseudocode derived from recent research papers to guide code generation, and uses test cases to debug and select among candidate implementations. However, this method depends heavily on manually provided pseudocode that is tightly coupled to specific tasks, which limits its scalability and general applicability in practice.

\textbf{Retrieval-augmented code generation.}
To further improve code-based graph reasoning, some studies incorporate retrieval augmentation and use external documents or prior experience as auxiliary information. GraphTeam \cite{li2025graphteamfacilitatinglargelanguage} adopts a multi-agent collaboration framework in which a retrieval agent uses semantic search to obtain relevant information from NetworkX documentation and prior experience, while a coding agent is responsible for code generation and compilation-based verification. GraphSkill \cite{wang2026graphskill} also uses NetworkX documentation as an external knowledge source, but further improves retrieval efficiency and accuracy through a retrieval agent built over hierarchically structured documents, and introduces a self-debugging mechanism to automatically correct both syntactic and logical errors in generated code. Similarly, GraphTool-Instruction \cite{wang2025GraphToolInstruction} explicitly provides a complete set of graph tool descriptions in the prompt to guide LLMs to generate tool-invoking code; however, this approach depends on manually constructed tool sets and requires enumerating all tool descriptions in the prompt, making it difficult to scale to large tool collections.

\textbf{Enhancing code-based reasoning through fine-tuning.}
Beyond retrieval augmentation, few works attempt to improve code-based graph reasoning through fine-tuning. For example, GCoder \cite{zhang2024gcoderimprovinglargelanguage} constructs a code reasoning dataset, GraphWild. Built around classical graph computation problems, this dataset converts 313 algorithm documents from NetworkX into problem-coding pairs and, after joint filtering by humans and LLMs, yields 49,224 training instances. GCoder, fine-tuned on this dataset, substantially outperforms general-purpose LLMs. At inference time, the method can be further combined with retrieval augmentation by retrieving relevant implementations from an offline code repository based on embedding similarity, thereby improving generalization to new tasks.


\section{Detailed Benchmark Construction Framework}
\label{sec:appendix_construction}

To support the continuous generation of fresh graph reasoning data, we design a semi-automatic LLM-based benchmark construction framework, {\dataset}. For simplicity, we use {\dataset} to refer to both the generated dataset and the construction framework, although they are not strictly identical. The framework introduces limited human verification at key stages to ensure data quality. It supports the expansion of new graph reasoning tasks, automatically generates supporting task description generation scripts for both file-based and inline graph loading, constructs prompts for textual and code-based reasoning, and enables automatic evaluation for both reasoning modes. Specifically, the construction pipeline consists of five stages: task composition, label generation, task description and question generation, evaluation-script generation, and quality filtering, as shown in Figure~\ref{fig:pipeline}, and uses DeepSeek-V3 \cite{liu2024deepseek} as the data generator. We describe each stage in detail below.

\subsection{Preliminary Stage: Seed Tasks and Composition Pattern Definition}

\begin{figure}[!htbp]
    \centering
    \includegraphics[width=\linewidth]{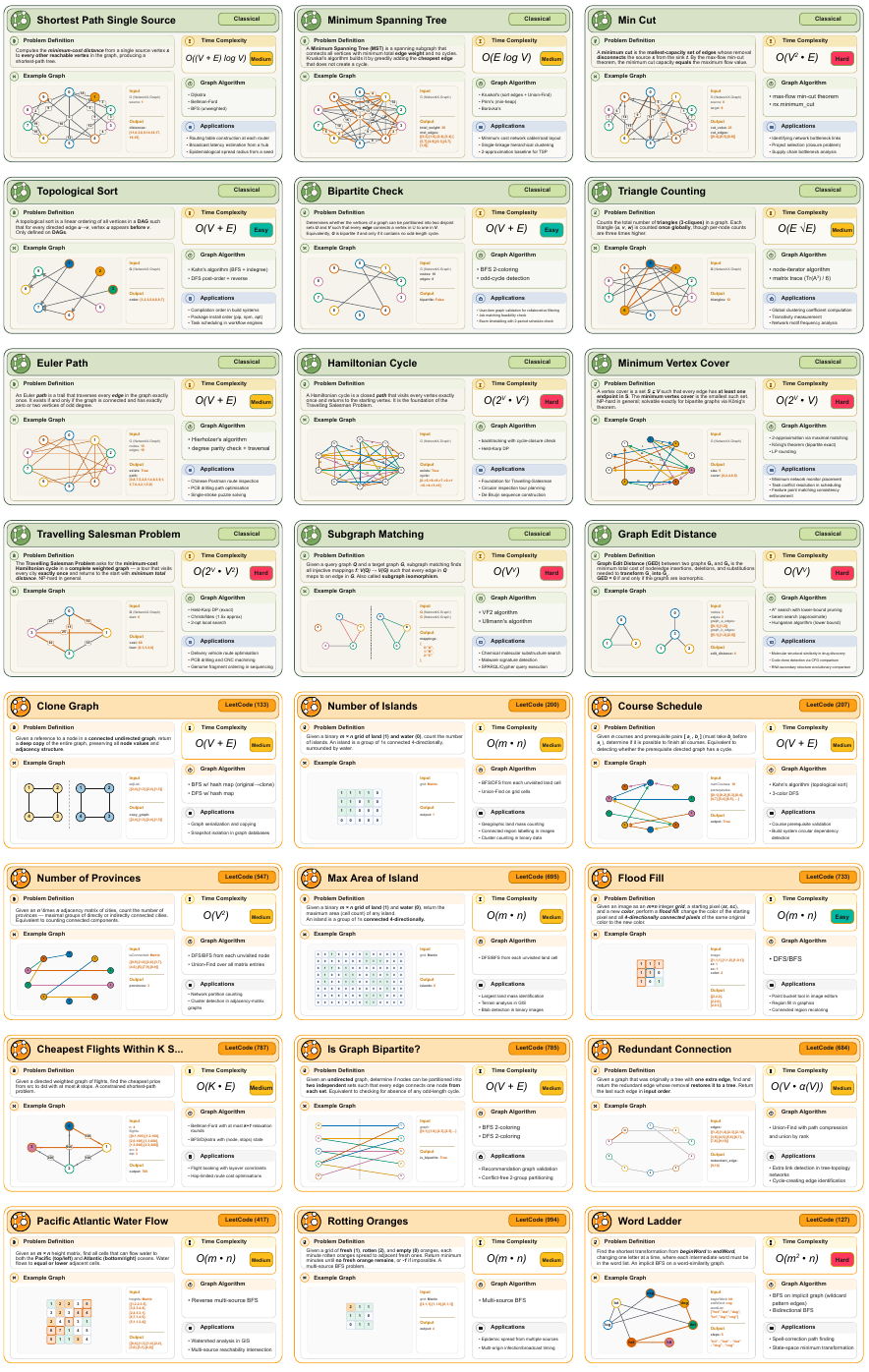}
    \caption{Representative examples of seed graph tasks used for composite task construction.}    
    \label{fig:seed_tasks}
\end{figure}

\textbf{Seed tasks.}
Composite tasks are typically more challenging than single tasks because models must solve multiple subtasks, maintain intermediate results, and combine these results according to the correct rule. Moreover, real-world graph problems are often not limited to simple classical graph algorithms; instead, they frequently involve multiple interdependent operations. Therefore, we explicitly incorporate compositional complexity into our benchmark construction. To generate diverse composite tasks, we first define a set of seed tasks, which serve as atomic units for further composition and extension. Existing graph reasoning benchmarks usually focus on classical graph algorithm tasks and provide limited coverage of composite tasks, as shown in Table~\ref{tab:graph_benchmark_tasks}. To cover the evaluation scope of prior benchmarks, we collect $40$ common classical graph tasks, such as connectivity testing, cycle detection, and shortest-path computation. However, classical graph tasks alone are often insufficient to fully challenge code-based reasoning, especially when models can call existing Python graph libraries. Therefore, we further introduce $60$ open-ended graph tasks from LeetCode to increase the diversity of task forms and solution requirements. Figure~\ref{fig:seed_tasks} shows representative examples of seed tasks. The complete task set is provided in our code repository.

\begin{figure}[t]
    \centering
    \includegraphics[width=\linewidth]{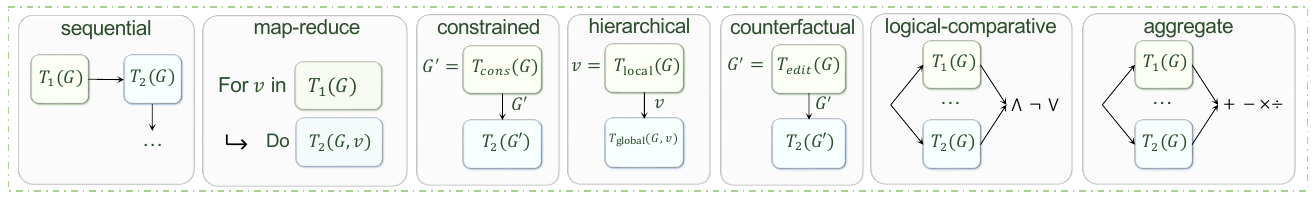}
    \caption{Composition patterns used for constructing composite graph reasoning tasks.}
    \label{fig:composition}
\end{figure}

\textbf{Composition pattern definition.}
When constructing composite tasks, we also need to specify how subtasks are combined. To this end, we define seven composition patterns based on common multi-step graph reasoning structures: \emph{sequential}, \emph{map-reduce}, \emph{constrained}, \emph{hierarchical}, \emph{counterfactual}, \emph{logical-comparative}, and \emph{aggregate}. These patterns capture subtask dependencies, structural constraints, hierarchical relations, graph modifications, repeated subproblem solving, and multi-quantity aggregation. 
We define seven composition patterns to characterize common subtask dependencies and reasoning challenges in composite graph tasks. Their execution logic is illustrated in Figure~\ref{fig:composition}. Such compositional patterns also reflect common graph reasoning needs in real-world applications. For example, transportation planning may require first filtering feasible routes under constraints and then optimizing travel time; social network analysis may require computing local user properties and aggregating them across communities; and infrastructure analysis may require simulating edge failures before reassessing connectivity. Thus, these patterns are not only synthetic combinations of graph algorithms, but also abstractions of multi-step decision processes over relational data.

\begin{itemize}[nosep, leftmargin=*]
    \item \textbf{\emph{Sequential}.} One subtask must be completed before the next subtask can be executed, and its output serves as the input to the following step. For example, the model may first identify a target subgraph and then compute its diameter or shortest path. This pattern evaluates whether the model can preserve and correctly use intermediate results across steps.
    
    \item \textbf{\emph{Map-reduce}.} The same operation is repeatedly applied to multiple subgraphs, and the resulting outputs are then aggregated. For example, the model may compute the diameter of each connected component and return the maximum value. This pattern evaluates iterative consistency across repeated subproblems and correct final aggregation.

    \item \textbf{\emph{Constrained}.} One subtask first defines a feasible set of nodes, edges, or subgraphs, and subsequent reasoning must be performed only within this restricted region. For example, the model may compute a maximum-flow value using only edges whose capacities exceed a given threshold, or find the shortest path restricted to nodes with degree above a specified value. This pattern evaluates whether the model can enforce structural constraints before performing optimization or reasoning under those constraints.
    
    \item \textbf{\emph{Hierarchical}.} The task involves both local and global graph structures. For example, the model may first identify the node with the highest local clustering coefficient and then determine whether that node lies on a global shortest path. This pattern evaluates whether the model can correctly connect neighborhood-level properties with graph-level properties.
    
    \item \textbf{\emph{Counterfactual}.} The graph is first modified, and the target property is then recomputed. For example, the model may remove bridge edges and then recount the connected components. This pattern evaluates whether the model can reason over the modified graph rather than reusing conclusions from the original graph.
    
    \item \textbf{\emph{Logical-comparative}.} The final answer is determined by logical or comparative relations among the outputs of multiple subtasks. For example, the model may determine whether a graph is both connected and bipartite. This pattern evaluates whether the model can apply the correct Boolean or comparison rule after solving the individual subtasks.
    
    \item \textbf{\emph{Aggregate}.} Multiple heterogeneous graph quantities are computed separately and then fused into a final answer. For example, the model may combine the size and diameter of the largest connected component into a structural score. This pattern evaluates whether the model can track multiple intermediate quantities with different meanings and combine them according to the specified rule.
    
\end{itemize}

\subsection{Stage 1: Task Composition and Graph Data Generation}
\label{sec:appendix_stage1}
Stage 1 constructs new tasks and their corresponding graph instances. Its goal is to compose seed tasks into composite graph-reasoning tasks and to generate graph data that satisfy each task's requirements. A single atomic task is treated as a degenerate composite task. The inputs to this stage are combo size, randomly sampled seed tasks, and the seven candidate composition patterns. The combo size denotes the number of seed tasks included in the composition; in our dataset, it takes values in $\{1,2,3,4\}$, although the framework is not restricted to this range. The outputs are the selected composition pattern, the new task definition, and the task-adaptive graph generation code. We generate task-adaptive graph generation code because different graph tasks impose different requirements on graph type, directionality, edge weights, graph scale, and input parameters.

\begin{wrapfigure}[17]{r}{0.7\columnwidth}
    \centering
    \includegraphics[width=\linewidth]{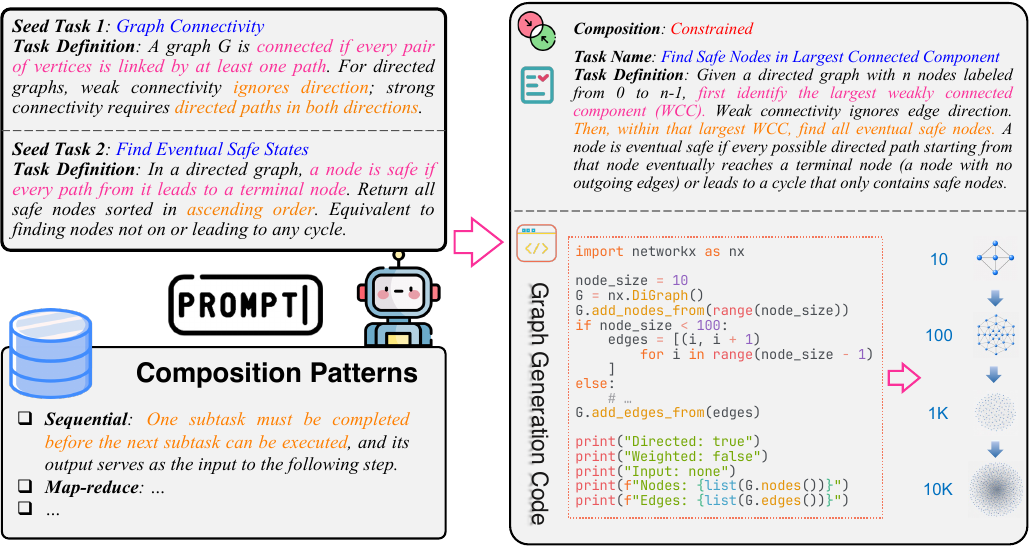}
    \caption{Stage 1 example of composite task construction and task-adaptive graph generation.}
    \label{fig:stage1_example}
\end{wrapfigure}
As shown in Figure~\ref{fig:stage1_example}, given two subtasks such as \texttt{graph connectivity} and \texttt{find eventual safe states}, we organize their task definitions, candidate composition patterns, and expected output format into a prompt, and ask the LLM to generate a new composite task definition together with the corresponding graph generation code; the complete prompt template is provided in Appendix~\ref{prompt_stage1}. During task sampling, we aim to maintain a balanced distribution across combo sizes and seed-task difficulty levels. We define seed-task difficulty according to algorithmic time complexity, including linear-time tasks $O(|V|)$, polynomial-time tasks $O(|V|^k), k>1$, and NP-hard tasks. After obtaining the graph generation code, we execute it automatically to produce graph instances and input parameters at four fixed graph sizes: $10$, $100$, $1{,}000$, and $10{,}000$ nodes. The same code can also be reused to generate model-specific graph sizes calibrated to context-window budgets, which supports the stress tests described later.
Overall, this stage covers two benchmark dimensions: \textbf{task complexity} and \textbf{graph scale}. Task complexity is controlled by varying combo size from $1$ to $4$, while graph scale is controlled by increasing the number of nodes from $10$ to $10{,}000$. The final outputs of this stage are the \textit{composition pattern}, the composite \textit{task definition}, the \textit{graph generation code}, and the generated \textit{graph instances} with their input parameters.

\begin{wrapfigure}[12]{r}{0.7\columnwidth}
    \centering
    \vspace{-2.4\baselineskip}
    \includegraphics[width=\linewidth]{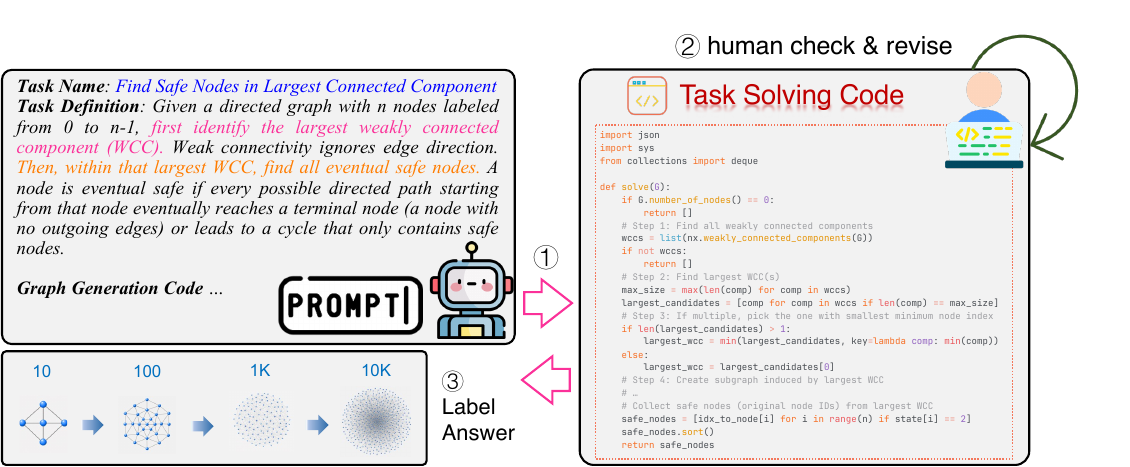}
    \caption{Stage 2 example of LLM-assisted reference solver generation and human validation.}
    \label{fig:stage2_example}
    \vspace{-0.8\baselineskip}
\end{wrapfigure}
\subsection{Stage 2: Solver Generation, Label Annotation, and Human Verification}
\label{sec:appendix_stage2}
After constructing new tasks and generating the corresponding graph instances, Stage 2 aims to produce correct answers for each graph instance. This stage is essential because the benchmark is only meaningful when its labels are accurate and reliable. Since our benchmark covers many tasks, multiple graph sizes, and composite reasoning workflows, fully manual answer derivation would be costly, inefficient, and error-prone, especially for multi-step graph reasoning. Therefore, we generate a Python-based reference solver for each task and use it to compute labels automatically.

Unlike prior work that mainly relies on manually written reference solvers~\cite{cai2024codegraphenhancinggraphreasoning, zhang2024gcoderimprovinglargelanguage}, we use LLMs to generate task-specific reference solvers and introduce human involvement only for validation, thereby reducing construction cost. Specifically, as shown in Figure~\ref{fig:stage2_example}, we organize the task description and graph-data format into a prompt and ask the LLM to generate executable Python solver code; the complete prompt template is provided in \S~\ref{prompt_stage2}. To ensure solver reliability, human annotators design simple test cases to validate the generated code functionally. If the code fails these tests, the error messages and human feedback are returned to the LLM, which then revises the solver until it passes validation. Finally, we apply the validated reference solver to the graph data and input parameters generated in the previous stage, producing gold labels for each graph instance.

\subsection{Stage 3: Task Description and Question Generation}
\label{sec:appendix_stage3}

After obtaining graph instances and verified answers, Stage 3 generates multiple task descriptions and question formats for each task. For task descriptions, we explicitly consider the complexity introduced by the presentation style. In addition to explicit descriptions that directly state the graph task and structural information, we also generate implicit descriptions, where the model must first map a real-world scenario to a graph problem before performing graph reasoning. This corresponds to the \textbf{task description} dimension in our data-complexity design.

Existing benchmarks typically focus on a single reasoning mode, either text-based reasoning or code-based reasoning, rather than evaluating both within the same task set. We therefore provide both text-mode and code-mode question formats for each task, enabling controlled comparisons between different reasoning paradigms. For code-based reasoning, many existing benchmarks still place the full graph data inside the prompt. This design weakens a key advantage of code-based reasoning: the ability to delegate graph loading and computation to executable programs, thereby reducing the LLM's context burden and unnecessary token consumption. To address this issue, we support two graph-loading modes for code-based reasoning: \emph{file} and \emph{inline}. In the file setting, the prompt only provides the graph file path; in the inline setting, the graph data are inserted directly into the task description. Because different tasks require different data formats, and implicit descriptions often require more flexible graph injection, we further use LLMs to generate task-specific graph-loading scripts for both loading modes. This corresponds to the \textbf{graph loading} dimension in our data-complexity design.

\begin{wrapfigure}{r}{0.7\columnwidth}
    \centering
    \vspace{-1.8\baselineskip}
    \includegraphics[width=\linewidth]{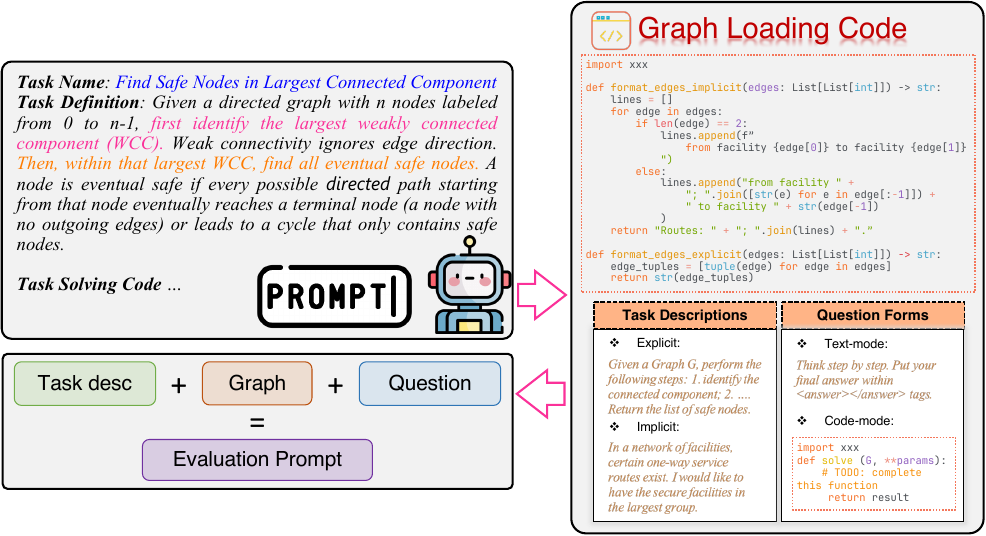}
    \caption{Stage 3 example of task description, question, and graph-loading script generation.}
    \label{fig:stage3_example}
    \vspace{-0.8\baselineskip}
\end{wrapfigure}
Figure~\ref{fig:stage3_example} illustrates this process. Given the task definition and the validated reference solver, we prompt the LLM to generate explicit and implicit task descriptions, text-mode and code-mode question formats, and graph-loading scripts; the complete prompt template is provided in \S~\ref{prompt_stage3}. The graph-loading scripts are used to inject graph instances into task descriptions in either file or inline form. Compared with file-based graph loading, inline loading is more complex because it must serialize nodes, edges, weights, and auxiliary parameters in a task-consistent format. These scripts allow the same task-description template to be reused across different graph instances, avoiding the need to regenerate a complete description for every graph.
Overall, this stage outputs \textit{explicit and implicit task descriptions}, \textit{graph-loading scripts for file and inline graph injection}, and \textit{question forms for text-based and code-based reasoning}. These components allow us to evaluate the same underlying graph task under controlled variations in description style, reasoning mode, and graph-loading mechanism. As a result, we can analyze not only whether a model solves a task, but also how its performance changes under different presentation conditions.

\subsection{Stage 4: Evaluation Script Generation}
\label{sec:appendix_stage4}
After the task descriptions, graph-loading scripts, and question formats are generated, Stage 4 constructs task-specific evaluation scripts to enable fully automated scoring of downstream model outputs. This stage is necessary because graph reasoning tasks produce heterogeneous answer types: some return Boolean values, some return numerical values, and others return structured objects such as sets, lists, or dictionaries. This heterogeneity is further amplified in composite tasks, where the final answer may combine multiple intermediate results. As a result, a single generic evaluator cannot reliably cover all task types. Existing benchmarks often rely on manually written evaluation code, which limits scalability when new tasks are added. In contrast, our framework uses LLMs to generate task-specific evaluation scripts, enabling scalable and reproducible evaluation.

For each task, we generate two evaluators, one for text-based reasoning and one for code-based reasoning. The textual evaluator is designed for natural-language responses: it extracts the final answer, parses it into a machine-readable Python object, and compares it with the gold label. This allows evaluation to focus on semantic correctness rather than surface-level wording, explanations, or formatting variations. The code evaluator is designed for executable-code responses: it extracts the model-generated \texttt{solve} function, inserts it into the evaluation boilerplate, executes the resulting program on benchmark graph instances, and compares the produced output with the gold labels.

\begin{wrapfigure}{r}{0.7\columnwidth}
    \centering
    \vspace{-0.8\baselineskip}
    \includegraphics[width=\linewidth]{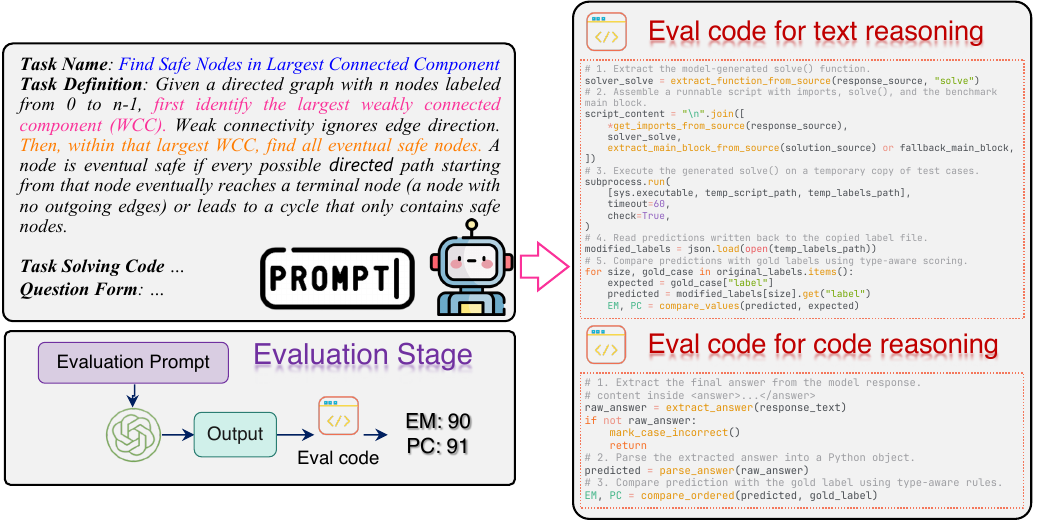}
    \caption{Stage 4 example of evaluation script generation.}
    \label{fig:stage4_example}
    \vspace{-0.8\baselineskip}
\end{wrapfigure}
In Figure~\ref{fig:stage4_example}, the LLM receives the task definition, question form, and reference solution code, and then generates the corresponding evaluation script; the complete prompt template is provided in \S~\ref{prompt_stage4}. During evaluation, the target LLM first produces an answer to the evaluation prompt. For text-based reasoning, the evaluator extracts the content inside \texttt{<answer>...</answer>}, parses it into a Python object, and applies type-aware comparison against the gold label. For example, floating-point outputs are compared using numerical tolerances, whereas set-valued outputs are compared without regard to element order. For code-based reasoning, the evaluator extracts the generated \texttt{solve} function from the response, combines it with a standard code-question template, and executes the assembled program in a Python environment to obtain the predicted output.

\textbf{Metrics.}
We report the Exact Match (EM) and Partial Credit (PC) scores. For the $i$-th test instance, let the model prediction be $\hat{y}_i$ and the gold label be $y_i$. EM measures whether the prediction exactly matches the gold label:
\[
\mathrm{EM}_i = \mathbb{I}\big[\hat{y}_i = y_i\big],
\]
where $\mathbb{I}[\cdot]$ is the indicator function. For structured outputs, equality is defined according to the output type. For example, set-valued outputs ignore element order, and floating-point outputs allow small numerical tolerance.

PC measures the degree of partial agreement between the prediction and the gold label:
\[
\mathrm{PC}_i = s(\hat{y}_i, y_i), \qquad s(\hat{y}_i,y_i)\in[0,1],
\]
where $s(\cdot,\cdot)$ is a type-aware similarity function. For scalar outputs, PC is identical to EM. For ordered lists, PC is computed as the proportion of elements that match at the same positions:
\[
s(\hat{y},y)=
\frac{\sum_{j=1}^{\min(|\hat{y}|,|y|)} \mathbb{I}[\hat{y}_j=y_j]}
{\max(|\hat{y}|,|y|,1)}.
\]
For set-valued outputs, we use Jaccard similarity:
\[
s(\hat{y},y)=\frac{|\hat{y}\cap y|}{|\hat{y}\cup y|}.
\]
For dictionary outputs, we compute the proportion of correctly matched key--value pairs:
\[
s(\hat{y},y)=
\frac{|\{k \mid k\in \hat{y}\cap y,\ \hat{y}[k]=y[k]\}|}
{\max(|\hat{y}|,|y|,1)}.
\]
Dataset-level EM and PC are then obtained by averaging over all $N$ test instances:
\[
\mathrm{EM}=\frac{1}{N}\sum_{i=1}^{N}\mathrm{EM}_i,
\qquad
\mathrm{PC}=\frac{1}{N}\sum_{i=1}^{N}\mathrm{PC}_i.
\]
EM captures fully correct predictions, while PC captures partial correctness for structured outputs, preventing near-correct answers with localized errors from being treated as entirely incorrect.

Overall, Stage 4 turns generated graph tasks into an executable evaluation framework. Once downstream model outputs are obtained, the corresponding evaluation scripts can be invoked directly to compute EM and PC, without manually writing task-specific evaluators. This design allows the benchmark to remain unified, reproducible, and scalable as new tasks are continuously added.

\begin{table*}[t]
\centering
\scriptsize
\caption{Quality validation dimensions used to score generated benchmark tasks. Each dimension is rated on a $1$--$5$ scale.}
\label{tab:quality_validation_dimensions}
\setlength{\tabcolsep}{3pt}
\renewcommand{\arraystretch}{1.15}
\begin{tabularx}{\linewidth}{p{0.18\linewidth} *{5}{>{\centering\arraybackslash}X}}
\toprule
\textbf{Dimension} & \textbf{5} & \textbf{4} & \textbf{3} & \textbf{2} & \textbf{1} \\
\midrule
\textbf{$Q_1$: Clarity} &
Immediately clear; all edge cases addressed. &
Clear format; no guessing needed. &
Generally understandable; minor ambiguities. &
Ambiguous key terms; guessing needed. &
Vague or contradictory. \\

\textbf{$Q_2$: Graph suitability} &
Ideal; exercises key algorithmic behaviors. &
Well-suited to all subtasks. &
Acceptable with minor mismatches. &
Correct type but degenerate structure. &
Graph contradicts the task. \\

\textbf{$Q_3$: Naturalness} &
Highly realistic practical scenario. &
Convincing and domain-appropriate. &
Plausible but generic. &
Contrived or implausible. &
Forced or contradictory. \\

\textbf{$Q_4$: Answer uniqueness} &
Provably unique answer. &
Unique for all tested inputs. &
Usually unique; rare ambiguity. &
Some edge cases allow alternatives. &
Multiple valid answers without tie-breaking. \\
\bottomrule
\end{tabularx}
\end{table*}

\begin{table}[t]
\centering
\caption{Cohen's $\kappa$ agreement on task-quality dimensions in the composite-task construction pipeline. We evaluate three rater pairs: two human annotators (H1, H2) and one LLM judge, over $20$ composite tasks rated by all three. Bold marks the highest $\kappa_q$ within each row; \emph{interp.} reports the Landis--Koch interpretation for H2 vs.\ LLM based on $\kappa_q$.}
\label{tab:human-llm-kappa}
\footnotesize
\setlength{\tabcolsep}{4pt}
\begin{tabularx}{\linewidth}{l YY YY YY l}
\toprule
& \multicolumn{2}{c}{H1 vs.\ H2}
& \multicolumn{2}{c}{H1 vs.\ LLM}
& \multicolumn{2}{c}{H2 vs.\ LLM}
& \emph{interp.} \\
\cmidrule(lr){2-3}\cmidrule(lr){4-5}\cmidrule(lr){6-7}
Dimension & $\kappa$ & $\kappa_q$ & $\kappa$ & $\kappa_q$ & $\kappa$ & $\kappa_q$ & (H2 vs.\ LLM) \\
\midrule
$Q_{1}$ clarity           & $0.776$ & $0.791$ & $\mathbf{1.000}$ & $\mathbf{1.000}$ & $0.776$ & $0.791$ & substantial \\
$Q_{2}$ graph suitability & $0.746$ & $\mathbf{0.943}$ & $0.831$ & $0.886$ & $0.570$ & $0.820$ & almost perfect \\
$Q_{3}$ naturalness       & $0.643$ & $0.576$ & $0.554$ & $0.598$ & $0.582$ & $\mathbf{0.701}$ & substantial \\
$Q_{4}$ answer uniqueness & $\mathbf{1.000}$ & $\mathbf{1.000}$ & $0.732$ & $0.966$ & $0.732$ & $0.966$ & almost perfect \\
\midrule
\textit{Pooled (4 dims)}  & $0.836$ & $\mathbf{0.903}$ & $0.835$ & $0.887$ & $0.772$ & $0.880$ & almost perfect \\
\bottomrule
\end{tabularx}
\end{table}

\subsection{Stage 5: Quality Validation and Filtering}
\label{sec:appendix_stage5}

Stage 5 verifies whether the automatically generated tasks are suitable for inclusion as benchmark instances. Its purpose is to remove low-quality samples and improve the reliability of the final evaluation set.
Although the preceding stages automatically generate tasks, labels, prompts, and evaluators, this process does not guarantee that every generated instance is well defined, structurally appropriate, and uniquely answerable. A generated task may contain semantic ambiguity, rely on a graph instance that is poorly matched to the target reasoning problem, or allow multiple equally valid outputs. Retaining such instances would make model errors difficult to interpret: a low score could reflect defects in the benchmark instance rather than genuine limitations in graph reasoning. Therefore, before finalizing the benchmark, we introduce quality validation as the final filtering step.
Each task is evaluated along four dimensions using a $1$--$5$ scale, where $1$ denotes the lowest quality and $5$ denotes the highest quality. The detailed rubric is provided in Table~\ref{tab:quality_validation_dimensions}.

\begin{itemize}
    \item \textbf{\emph{Q1: Clarity}.} This dimension evaluates whether the task statement is precise and easy to understand. For example, the task should clearly specify whether the graph is directed or undirected and define the expected output format.

    \item \textbf{\emph{Q2: Graph suitability}.} This dimension evaluates whether the generated graph is appropriate for the target reasoning problem. An unsuitable graph may make the task trivial or ill-defined, such as using a graph with little or no branching for a path-selection problem.
    
    \item \textbf{\emph{Q3: Naturalness}.} This dimension evaluates whether the real-world task formulation resembles a plausible practical application rather than an artificial narrative wrapped around a graph algorithm.
    
    \item \textbf{\emph{Q4: Answer uniqueness}.} This dimension evaluates whether the task has a clearly defined unique target answer rather than multiple equally valid outputs. For example, if the benchmark expects a single path, the task should avoid cases with multiple equally valid shortest paths unless an explicit tie-breaking rule is specified.
\end{itemize}

\begin{wrapfigure}{r}{0.7\columnwidth}
    \centering
    \vspace{-0.8\baselineskip}
    \includegraphics[width=\linewidth]{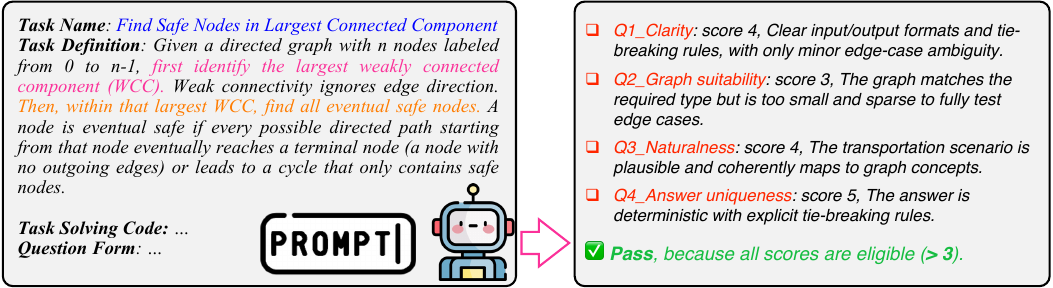}
    \caption{Stage 5 example of quality validation and filtering.}
    \label{fig:stage5_example}
    \vspace{-0.8\baselineskip}
\end{wrapfigure}
In implementation, we provide the task description, question format, and reference solver to the LLM judge and construct a quality-assessment prompt; the complete prompt template is provided in \S~\ref{prompt_stage5}. The LLM judge outputs both scores and justifications for the four dimensions. Figure~\ref{fig:stage5_example} presents a concrete example. If a task receives a score below $3$ on any dimension, it is removed from the benchmark. In the illustrated example, all dimension scores are at least $3$, so the task is retained. Thus, Stage 5 functions as the final quality-control mechanism in the automatic construction pipeline: it does not generate new benchmark content, but determines which generated tasks are sufficiently clear, appropriate, natural, and reliable for inclusion in the final evaluation set.

\subsection{Human validation of LLM quality judgments.}
\label{sec:appendix_human}

Because Stage 5 relies on an LLM judge for large-scale filtering, we further examine whether its quality judgments are consistent with human annotations. Specifically, two human annotators with graph-algorithm knowledge, one undergraduate student and one graduate student in computer science, independently score $20$ randomly sampled composite tasks using the same $1$--$5$ Likert scale as the LLM judge. During annotation, the annotators cannot see each other's scores or the LLM scores.

We measure agreement using Cohen's $\kappa$~\cite{cohen1960coefficient}, where larger values indicate stronger agreement. Since the ratings are ordinal Likert-scale scores, we use quadratic-weighted $\kappa_q$ as the primary metric, because it penalizes severe disagreements, such as $1$ vs.\ $5$, more heavily than minor disagreements, such as $4$ vs.\ $5$. We also report unweighted $\kappa$ for completeness. All four dimensions follow the shared rubric in Table~\ref{tab:quality_validation_dimensions}. Table~\ref{tab:human-llm-kappa} reports the agreement among the two human annotators and the LLM judge.

Across the four dimensions, the two human annotators achieve a pooled $\kappa = 0.836$ and a pooled $\kappa_q = 0.903$. According to the Landis--Koch scale, this corresponds to ``almost perfect'' agreement. This high inter-human agreement suggests that the four quality dimensions are sufficiently well defined for annotators with relevant background knowledge to assess task quality consistently.

Overall, the LLM judge achieves agreement close to this human agreement ceiling. Pooled across the four dimensions, the LLM obtains $\kappa_q = 0.887$ with H1 and $\kappa_q = 0.880$ with H2, both within approximately $0.02$ of the human-human agreement level of $\kappa_q = 0.903$. The unweighted $\kappa$ values show a similar pattern.
These results indicate that, under a clearly specified rubric, the LLM judge aligns strongly with human annotators and approaches the level of inter-human agreement. Therefore, using the LLM judge for quality filtering of automatically generated tasks is a reliable choice.

\subsection{{\dataset} Distribution}
\label{sec:appendix_benchmark_dist}
{\dataset} contains $202$ composite graph-reasoning tasks, instantiated along four axes of variation summarized in Table~\ref{tab:benchmark-dist-axes}.

\begin{wraptable}[8]{r}{0.6\linewidth}
\centering
\vskip -2.9em
\caption{Per-task axes of variation in {\dataset}.}
\vskip -0.6em
\label{tab:benchmark-dist-axes}
\footnotesize
\setlength{\tabcolsep}{4pt}
\renewcommand{\arraystretch}{1.08}
\begin{tabular}{l l r}
\toprule
Axis & Levels & Card. \\
\midrule
Task description & real-world, formal & $2$ \\
Reasoning scenario & textual-inline, coding-inline, coding-file & $3$ \\
Graph size & $n \in \{10,100,1{,}000,10{,}000\}$ & $4$ \\
\midrule
Per task & $2 \times 3 \times 4$ & $\mathbf{24}$ \\
Total instances & $(202+193+186+118) \times 2 \times 3$ & $\mathbf{4{,}198}$ \\
\bottomrule
\end{tabular}
\vskip 0.5em
{\footnotesize 
\textit{Note:} Some large-size instances are missing due to timeouts (Tab.~\ref{tab:benchmark-dist-graph-size}).
}
\vskip -1em
\end{wraptable}

\textbf{Per-task axes.}
Each task carries (i)~\emph{two task descriptions} -- an \emph{implicit} real-world narrative that situates the graph problem in a domain context and a \emph{explicit} formal description that states the same problem in mathematical terms; (ii)~\emph{two question forms}, a \emph{text-based} question that asks the model to reason over the graph in natural language and a \emph{code-based} question that asks the model to write a Python program that solves the task; and (iii)~\emph{two graph-loading modes}, \emph{inline} (the full graph is serialized into the prompt) and \emph{file} (the prompt only contains a path and the generated program reads the graph from disk). By construction, text-based questions only use the inline mode (the model has no execution environment in which to read a file), whereas code-based questions use both modes. The combination $\{\text{textual-inline}, \text{coding-inline}, \text{coding-file}\}$ therefore yields three reasoning scenarios per task. Each task is finally instantiated at \emph{four graph sizes} $n \in \{10, 100, 1{,}000, 10{,}000\}$ to stress-test scaling. The Cartesian product $2 \times 3 \times 4$ gives $24$ evaluation instances per task and 
${4{,}198}$ instances due to generation timeous for some large-size graphs.

\begin{wraptable}[8]{r}{0.32\linewidth}
\centering
\vskip -1.8em
\caption{Distribution of combo size in {\dataset}.}
\vskip -0.6em
\label{tab:benchmark-dist-combo}
\small 
\setlength{\tabcolsep}{6pt}
\renewcommand{\arraystretch}{1.08}
\begin{tabular}{c r r}
\toprule
Combo size & Tasks & \% \\
\midrule
$1$ & $55$ & $27.2$ \\
$2$ & $65$ & $32.2$ \\
$3$ & $41$ & $20.3$ \\
$4$ & $41$ & $20.3$ \\
\midrule
Total & $\mathbf{202}$ & $100.0$ \\
\bottomrule
\end{tabular}
\vskip -1em
\end{wraptable}
\textbf{Combo size (Table~\ref{tab:benchmark-dist-combo}).}
Tasks are stratified by composition depth, the number of seed primitives composed together. Combo-size-$1$ tasks are direct lifts of the seed primitives ($55$ tasks; $27.2\%$). The remaining $147$ tasks ($72.8\%$) chain two to four seed problems via explicit composition operators: $32.2\%$ are pairs (combo $2$, $65$ tasks), and combo sizes $3$ and $4$ each contribute $20.3\%$ ($41$ tasks). The composition mix is approximately balanced across depths; the slight skew toward combo $2$ reflects the larger combinatorial pool available at depth $2$.

\begin{wraptable}[13]{r}{0.38\linewidth}
\centering
\vskip -1em
\caption{Distribution of composition patterns in {\dataset}.}
\vskip -0.6em
\label{tab:benchmark-dist-pattern}
\small
\setlength{\tabcolsep}{6pt}
\renewcommand{\arraystretch}{1.08}
\begin{tabular}{l r r}
\toprule
Pattern & Tasks & \% \\
\midrule
constrained & $61$ & $30.2$ \\
none (single-seed) & $56$ & $27.7$ \\
hierarchical & $35$ & $17.3$ \\
logical-comparative & $31$ & $15.3$ \\
map-reduce & $9$ & $4.5$ \\
sequential & $7$ & $3.5$ \\
counterfactual & $3$ & $1.5$ \\
\midrule
Total & $\mathbf{202}$ & $100.0$ \\
\bottomrule
\end{tabular}
\end{wraptable}
\textbf{Composition pattern (Table~\ref{tab:benchmark-dist-pattern}).}
Tasks with combo size $\geq 2$ chain seeds via one of six explicit operators. \emph{Constrained} composition dominates at $30.2\%$, followed by \emph{hierarchical} composition ($17.3\%$) and \emph{logical-comparative} composition ($15.3\%$). The rarer operators -- \emph{map-reduce}, \emph{sequential}, and \emph{counterfactual} -- account for $4.5\%$, $3.5\%$, and $1.5\%$ respectively. Single-seed pass-through tasks (\emph{none}, $27.7\%$) coincide almost exactly with combo size $1$. The long-tail distribution reflects the relative ease of generating valid task instances under each operator: \emph{constrained} compositions have the broadest applicability, while \emph{counterfactual} compositions require carefully matched seed pairs.

\begin{table}[t]
\centering
\caption{Distribution of graph sizes in {\dataset}.}
\label{tab:benchmark-dist-graph-size}
\small 
\setlength{\tabcolsep}{4pt}
\renewcommand{\arraystretch}{1.08}
\begin{tabular}{r r r r r r}
\toprule
Graph size & Realized tasks & Avg. $|V|$ & Avg. $|E|$ & Directed & Weighted \\
\midrule
$10$ & $202$ & $10$ & $17.5$ & $35\%$ & $22\%$ \\
$100$ & $193$ & $100$ & $881.2$ & $32\%$ & $22\%$ \\
$1{,}000$ & $186$ & $1{,}000$ & $7{,}359.8$ & $32\%$ & $22\%$ \\
$10{,}000$ & $118$ & $10{,}000$ & $14{,}898.6$ & $26\%$ & $15\%$ \\
\bottomrule
\end{tabular}
\end{table}
\textbf{Graph size (Table~\ref{tab:benchmark-dist-graph-size}).}
For each task, the underlying graph generator is re-run at four target node sizes $n \in \{10, 100, 1{,}000, 10{,}000\}$. All $202$ tasks realize the smallest size; coverage drops to $193 / 186 / 118$ tasks at the larger three sizes, where the per-task generator either times out on intractable code or refuses to produce a valid instance. 
Roughly one in three graphs is directed and one in five is weighted; both fractions decline slightly at $n = 10{,}000$ because the task pool that survives to that scale is a slightly easier subset.

\begin{takeawaybox}
\small 
The five-stage pipeline converts graph task generation into an executable, diverse, and systematically controlled framework. The resulting benchmark is not merely a collection of graph tasks, but a complete evaluation framework: each task is associated with graph instances at specified sizes, solution scripts for deriving ground-truth labels, multiple task descriptions and question forms, file-based and inline graph-loading scripts, task-specific evaluation scripts, and quality-assessment results. This design enables controlled analysis of model performance across five dimensions of data complexity. The pipeline uses LLMs as task generators and relies primarily on human validation, thereby reducing the cost of large-scale manual construction. This design also improves extensibility: when new tasks need to be added, the same pipeline can be applied with limited additional human effort.
    
\end{takeawaybox}




\subsection{Evaluation Pipeline.}
\label{sec:appendix_evaluation}
Each benchmark instance is defined by a task, graph size, task description, and reasoning mode. The evaluation process is fully automated. For textual reasoning, the graph-loading script first combines the selected task description with the corresponding graph instance to produce a graph-grounded prompt, which is then concatenated with the textual question and sent to the evaluated LLM. The model prediction is then compared with the ground-truth answer by the text-mode evaluation script, which reports exact match (EM) and partial credit (PC). Coding-mode evaluation follows a similar pipeline but differs in execution logic. The model is prompted to generate executable Python code, from which the \texttt{solve} function is extracted and combined with task-specific boilerplate code for graph loading, either from inline graph data or a file path. The resulting program is executed to obtain the model prediction, which is then compared with the ground truth to compute EM and PC scores.

\subsection{Benchmark Construction Cost Analysis}
\label{sec:cost-analysis}

We analyze the construction cost of {\dataset} based on the actual LLM calls used in the final benchmark generation process. The final benchmark contains $202$ composite tasks, each associated with multiple task descriptions, question formats, graph generators, reference solvers, graph-loading scripts, and evaluation scripts. All data-generation calls were conducted using DeepSeek-V3.2. The cost is computed from recorded input and output token counts using the provider pricing at the time of construction: $\$0.14$ per million input tokens and $\$0.28$ per million output tokens.

\paragraph{Pipeline-level cost accounting.}
The five-stage construction pipeline involves LLM calls in stages S1, S3, S4, and S5, while S2 is based on program execution and does not require LLM calls. In S1, the model generates graph-construction code and task specifications for candidate composite tasks. In S2, candidate generators and reference solvers are executed to produce ground-truth labels, and invalid or failed tasks are removed. In S3, the model generates scenario-specific task descriptions, question formats, and graph-loading scripts. In S4, it generates evaluation scripts for textual and coding modes, which are then validated against reference solutions. In S5, an LLM-based quality check evaluates each generated task along predefined quality criteria.
The practical number of tasks across stages is: \[
400 \xrightarrow{\text{S1}} 400
\xrightarrow{\text{S2}} 268
\xrightarrow{\text{S3}} 268
\xrightarrow{\text{S4}} 265
\xrightarrow{\text{S5}} 
202 .
\]

\paragraph{Measured token usage and construction cost.}
Table~\ref{tab:construction-cost} reports the measured token usage and corresponding cost for each construction stage. In total, the final construction process used $3.71$M input tokens and $2.23$M output tokens, resulting in a total LLM API cost of $\$1.143$.

\begin{table}[t]
\centering
\caption{Measured LLM cost for constructing {\dataset}. Token counts are recorded from the final construction process. Costs are computed using the DeepSeek-V3.2 pricing at the time of construction: $\$0.14$/M input tokens and $\$0.28$/M output tokens.}
\label{tab:construction-cost}
\small
\setlength{\tabcolsep}{4pt}
\renewcommand{\arraystretch}{1.15}
\begin{tabularx}{\linewidth}{l X r r r r}
\toprule
\textbf{Stage} & \textbf{Generated Artifact}
& \textbf{Calls} & \textbf{Input Tokens} & \textbf{Output Tokens}
& \textbf{Cost} \\
\midrule
S1 & Graph-generation code and task specifications
   & $426$ & $545{,}170$ & $302{,}129$ & $\$0.161$ \\
S2 & Graph labeling by program execution
   & $0$   & --- & --- & $\$0.000$ \\
S3 & Task descriptions, question formats, and graph-loading scripts
   & $825$ & $1{,}155{,}915$ & $675{,}966$ & $\$0.351$ \\
S4 & Textual and coding evaluation scripts
   & $533$ & $1{,}222{,}384$ & $1{,}164{,}400$ & $\$0.497$ \\
S5 & LLM-based quality validation
   & $265$ & $785{,}262$ & $86{,}551$ & $\$0.134$ \\
\midrule
\multicolumn{2}{l}{\textbf{Total}}
   & $2{,}049$ & $3{,}708{,}731$ & $2{,}229{,}046$
   & $\boldsymbol{\$1.143}$ \\
\bottomrule
\end{tabularx}
\end{table}

\paragraph{Amortized construction cost.}
The measured total cost corresponds to $\$0.0057$ per final composite task. Since the benchmark contains $202$ tasks, $4$ graph sizes, and $6$ evaluation settings across coding and textual modes, it yields $4{,}198$ canonical evaluation instances, corresponding to $\$2.4\times10^{-4}$ per instance. Each task also contains $11$ main generated artifacts, including the graph generator, reference solver, graph-loading script, evaluation scripts, task descriptions, and question formats. This gives an amortized cost of $\$5.1\times10^{-4}$ per generated artifact.


\section{Detailed Experimental Setup, Results, and Analysis}
\label{sec:appendix_experiment}

\subsection{Experimental Setup}
\label{sec:appendix_experiment_setup}

\textbf{LLM Selection.} We select a diverse set of LLMs across multiple families and parameter sizes, covering open-source, closed-source, general-purpose, reasoning-oriented, and coding-specialized models. Specifically, we evaluate \texttt{DeepSeek-V3.2} \cite{liu2024deepseek}, \texttt{DeepSeek-R1} \cite{guo2025deepseek}, \texttt{DeepSeek-R1-Distill-32B} \cite{guo2025deepseek}, \texttt{DeepSeek-R1-Distill-70B} \cite{guo2025deepseek}, \texttt{Gemma-4-31B} \cite{gemma42025}, \texttt{Llama-3.1-8B} \cite{dubey2024llama}, \texttt{Llama-3.3-70B}, \texttt{Llama-4-Maverick} \cite{llama42025}, \texttt{Llama-4-Scout} \cite{llama42025}, \texttt{Nemotron-3-Super} \cite{blakeman2025nvidia}, \texttt{o4-mini}~\cite{openai2024o4mini}, \texttt{Qwen2.5-7B} \cite{yang2025qwen3}, \texttt{Qwen2.5-72B} \cite{yang2025qwen3}, \texttt{Qwen3-32B} \cite{yang2025qwen3}, \texttt{Qwen3-Coder-30B} \cite{yang2025qwen3}, and \texttt{QwQ-32B} \cite{qwq-32b-preview}. These models span representative families such as Llama~\cite{grattafiori2024llama}, DeepSeek~\cite{liu2024deepseek}, and Qwen~\cite{bai2023qwentechnicalreport}. This selection provides a cross-sectional view of the current LLM landscape across model scale, architecture, training objective, and deployment type. 
Although DeepSeek-V3.2 is used in benchmark construction, downstream evaluation prompts contain only the final task instance and do not expose the reference solver, construction prompts, or validation traces. We also report results for diverse model families to avoid conclusions depending on the construction model.

\textbf{Hyperparameters.}
All LLMs are evaluated with greedy decoding (\texttt{temperature}$=0$) and one sample per prompt, so the reported scores reflect each model's deterministic behavior. Unless otherwise specified, the maximum output length is set to $8{,}192$ tokens. For large-scale stress tests with $|V|=10{,}000$, we increase the output budget according to each model's context window while ensuring that the combined prompt and output length remains within the supported limit. Reasoning models are evaluated under their default reasoning configurations. We use zero-shot chain-of-thought prompting~\cite{wei2022chain} throughout.

\textbf{LLM Usage and Access.}
LLM usage refers to the role of LLMs in our benchmark construction and evaluation here, rather than their use in paper writing. LLMs serve two roles in our benchmark. First, they are used as data generators in the five-stage construction pipeline, including drafting graph generators, reference solvers, task descriptions, question formats, graph-loading scripts, and quality-validation checks. The generated artifacts are further validated by LLMs before being included in the benchmark.
Second, LLMs are evaluated as downstream reasoners under both text- and code-based settings. For each task instance, the evaluated model receives only the constructed prompt and produces either executable code or a textual answer, without access to the reference solver or construction pipeline. All LLMs are accessed through official or accessible API providers, including OpenAI \cite{openai2024o4mini}, DeepSeek \cite{deepseekai2024deepseekcoderv2breakingbarrierclosedsource}, and OpenRouter \cite{openrouter2026}.

\textbf{Baseline Training and Inference Environment.}
Two categories of baselines require local GPU computation: fine-tuned LLMs and supervised GNNs. For the fine-tuned LLM baseline, we reproduce GCoder \cite{zhang2024gcoderimprovinglargelanguage} by applying LoRA-based supervised fine-tuning to an instruction-tuned backbone on the GraphWild corpus. The training configuration follows the original GCoder setting, including low-rank adaptation, two training epochs, cosine learning-rate decay, and gradient checkpointing. Details are provided in \S~\ref{sec:appendix_finetune}.
For task-specific GNN baselines, we follow GraphArena~\cite{tang2025grapharena} and train one model for each task and architecture, covering GCN, GAT, GIN, and GraphSAGE. Each GNN uses two message-passing layers, hidden dimension $16$, AdamW optimization, early stopping on a validation split, and task-appropriate losses for regression and Boolean prediction. Details are provided in \S~\ref{sec:appendix_gnn}.
GCoder fine-tuning and GNN training are conducted on a NVIDIA A6000 GPU. API-served LLMs are evaluated through external providers and therefore incur no local GPU cost.

\begin{table*}[t]
\centering
\caption{Exact Match (EM) by graph size, combo size, and descriptions. The Gap column reports Explicit $-$ Implicit. Panels A1 and A2 split the coding-mode results by graph-loading scenario: Panel A1 reports \emph{file} loading; Panel A2 reports \emph{inline} loading. Panel B reports textual-mode results (always inline). For \textbf{Overall}, \textbf{Explicit}, and \textbf{Implicit}, \colorbox{rk1}{\textbf{bold red}} / \colorbox{rk2}{\underline{underlined yellow}} / \colorbox{rk3}{light green} mark the best / second-best / third-best models within each panel. Background shading on Graph size (peach) and Combo size (sky blue) indicates problem difficulty (light $\to$ deep = easier $\to$ harder). }
\label{tab:merged-size-combo-description}
\tiny
\setlength{\tabcolsep}{2pt}
\renewcommand{\arraystretch}{1.1}
\begin{tabularx}{\linewidth}{l Y >{\columncolor{gs1}}Y >{\columncolor{gs2}}Y >{\columncolor{gs3}}Y >{\columncolor{gs4}}Y >{\columncolor{cs1}}Y >{\columncolor{cs2}}Y >{\columncolor{cs3}}Y >{\columncolor{cs4}}Y YY >{\columncolor{gapColumn}}Y}
\toprule
& \textbf{Overall} 
& \multicolumn{4}{c}{\textbf{Graph size}}
& \multicolumn{4}{c}{\textbf{Combo size}}
& \multicolumn{3}{c}{\textbf{Descriptions}} \\
\cmidrule(lr){3-6}\cmidrule(lr){7-10}\cmidrule(lr){11-13}
\textbf{Model} & \textbf{EM}
& \textbf{10} & \textbf{100} & \textbf{1k} & \textbf{10k}
& \textbf{1} & \textbf{2} & \textbf{3} & \textbf{4}
& \textbf{Explicit} & \textbf{Implicit} & \textbf{Gap} \\
\midrule
\rowcolor{panelBg}
\multicolumn{13}{l}{\textit{Panel A1: Coding-mode strict EM --- file graph loading}} \\
\midrule
DeepSeek-V3.2
& \third{$74.8$} & $87.3$ & $82.3$ & $79.1$ & $50.2$ & $82.3$ & $80.6$ & $68.8$ & $61.3$ & \third{$76.0$} & \second{$73.5$} & $2.5$ \\
DeepSeek-R1
& \second{$75.3$} & $86.9$ & $84.2$ & $79.7$ & $50.5$ & $84.6$ & $80.0$ & $72.9$ & $56.5$ & $75.7$ & \first{75.0} & $0.7$ \\
DS-R1-Distill-32B
& $70.2$ & $81.7$ & $76.6$ & $73.3$ & $49.5$ & $77.6$ & $76.8$ & $61.7$ & $49.4$ & $72.8$  & $67.6$ & $5.2$ \\
DS-R1-Distill-70B
& $63.3$ & $72.4$ & $69.6$ & $67.0$ & $44.0$ & $70.0$ & $66.6$ & $59.0$ & $50.0$ & $69.5$ & $56.9$ & $12.6$ \\
Gemma-4-31B
& $74.3$ & $85.6$ & $82.1$ & $78.1$ & $51.5$ & $80.0$ & $78.7$ & $71.6$ & $62.5$ & $75.2$ & $73.4$ & $1.8$ \\
Llama-3.1-8B
& $35.9$ & $39.8$ & $39.3$ & $38.5$ & $26.3$ & $39.4$ & $45.0$ & $28.4$ & $24.4$ & $35.7$ & $36.1$ & $-0.4$ \\
Llama-3.3-70B
& $57.5$ & $65.7$ & $62.7$ & $60.2$ & $41.4$ & $66.3$ & $64.7$ & $54.3$ & $36.9$ & $59.8$ & $55.2$ & $4.6$ \\
Llama-4-Maverick
& $58.5$ & $68.1$ & $65.1$ & $61.8$ & $39.2$ & $71.1$ & $62.2$ & $52.7$ & $41.2$ & $60.9$ & $56.1$ & $4.8$ \\
Llama-4-Scout
& $50.4$ & $59.2$ & $54.9$ & $51.8$ & $35.6$ & $57.9$ & $58.3$ & $45.3$ & $32.4$ & $51.9$ & $48.9$ & $3.0$ \\
Nemotron-3-Super
& \first{75.5} & $86.6$ & $83.7$ & $80.4$ & $51.5$ & $82.9$ & $79.7$ & $72.9$ & $60.6$ & \first{77.6} & \third{$73.5$} & $4.1$ \\
o4-mini
& $74.1$ & $85.9$ & $81.9$ & $79.0$ & $49.5$ & $80.0$ & $79.4$ & $72.0$ & $59.8$ & \second{$76.1$} & $72.0$ & $4.1$ \\
Qwen2.5-72B
& $61.8$ & $71.3$ & $67.3$ & $65.8$ & $42.6$ & $70.9$ & $69.2$ & $54.6$ & $44.8$ & $62.4$ & $61.1$ & $1.3$ \\
Qwen2.5-7B
& $32.8$ & $37.4$ & $35.4$ & $34.4$ & $24.0$ & $32.7$ & $43.5$ & $25.6$ & $23.2$ & $33.8$ & $31.8$ & $2.0$ \\
Qwen3-32B
& $73.4$ & $84.3$ & $80.4$ & $78.8$ & $50.3$ & $76.3$ & $75.0$ & $71.4$ & $66.7$ & $74.8$ & $72.0$ & $2.8$ \\
Qwen3-Coder-30B
& $68.2$ & $78.7$ & $74.4$ & $71.9$ & $47.6$ & $78.0$ & $73.1$ & $63.6$ & $51.9$ & $70.1$ & $66.2$ & $3.9$ \\
\cmidrule(l){2-13}
\textit{Average}
& $63.1$ & $72.7$ & $69.3$ & $66.7$ & $43.6$ & $70.0$ & $68.9$ & $58.3$ & $48.1$ & $64.8$ & $61.3$ & $3.5$ \\
\midrule
\rowcolor{panelBg}
\multicolumn{13}{l}{\textit{Panel A2: Coding-mode strict EM --- inline graph loading}} \\
\midrule
DeepSeek-V3.2
& \third{$46.7$} & $89.5$ & $65.2$ & $21.0$ & $11.0$ & $55.0$ & $46.9$ & $44.1$ & $37.8$ & $48.6$ & \second{$44.8$} & $3.8$ \\
DeepSeek-R1
& \second{$48.0$} & $87.7$ & $56.8$ & $26.2$ & $21.5$ & $51.2$ & $45.9$ & $51.3$ & $44.3$ & \second{$51.7$} & $44.3$ & $7.4$ \\
DS-R1-Distill-32B
& $33.9$ & $67.4$ & $48.6$ & $16.6$ & $3.0$ & $42.2$ & $31.1$ & $31.6$ & $27.2$ & $33.8$ & $34.0$ & $-0.2$ \\
DS-R1-Distill-70B
& $37.1$ & $68.6$ & $48.4$ & $25.0$ & $6.4$ & $44.7$ & $35.8$ & $38.2$ & $27.4$ & $37.3$ & $36.9$ & $0.4$ \\
Gemma-4-31B
& $36.0$ & $83.5$ & $42.1$ & $11.5$ & $7.0$ & $43.6$ & $37.2$ & $33.8$ & $26.2$ & $36.5$ & $35.6$ & $0.9$ \\
Llama-3.1-8B
& $22.9$ & $43.9$ & $33.3$ & $9.5$ & $5.0$ & $28.7$ & $24.2$ & $22.8$ & $13.4$ & $23.6$ & $22.2$ & $1.4$ \\
Llama-3.3-70B
& $34.9$ & $72.9$ & $46.5$ & $14.8$ & $5.5$ & $42.9$ & $35.5$ & $36.6$ & $21.6$ & $35.1$ & $34.7$ & $0.4$ \\
Llama-4-Maverick
& $43.4$ & $75.7$ & $52.5$ & $24.2$ & $21.2$ & $50.0$ & $46.5$ & $42.1$ & $31.1$ & $43.9$ & $43.0$ & $0.9$ \\
Llama-4-Scout
& $37.4$ & $61.8$ & $46.5$ & $22.9$ & $18.3$ & $43.3$ & $43.3$ & $37.2$ & $19.6$ & $38.3$ & $36.4$ & $1.9$ \\
Nemotron-3-Super
& $46.5$ & $84.3$ & $55.3$ & $24.9$ & $21.4$ & $55.7$ & $44.4$ & $45.2$ & $37.7$ & \third{$47.9$} & \third{$45.0$} & $2.9$ \\
o4-mini
& \first{57.3} & $89.8$ & $69.1$ & $43.1$ & $27.2$ & $64.7$ & $58.1$ & $57.5$ & $46.0$ &  \first{58.6} & \first{56.0} & $2.6$ \\
Qwen2.5-72B
& $38.3$ & $73.8$ & $55.6$ & $20.2$ & $3.7$ & $45.2$ & $41.3$ & $38.8$ & $24.1$ & $38.9$ & $37.8$ & $1.1$ \\
Qwen2.5-7B
& $26.6$ & $42.6$ & $36.9$ & $22.2$ & $4.5$ & $29.1$ & $32.1$ & $24.4$ & $16.5$ & $26.6$ & $26.5$ & $0.1$ \\
Qwen3-32B
& $34.9$ & $73.5$ & $42.4$ & $17.6$ & $6.1$ & $39.1$ & $34.5$ & $34.4$ & $29.0$ & $35.3$ & $34.5$ & $0.8$ \\
Qwen3-Coder-30B
& $41.1$ & $80.7$ & $56.8$ & $16.3$ & $10.8$ & $46.8$ & $42.3$ & $39.7$ & $33.1$ & $45.4$ & $36.9$ & $8.5$ \\
\cmidrule(l){2-13}
\textit{Average}
& $39.0$ & $73.0$ & $50.4$ & $21.1$ & $11.5$ & $45.5$ & $39.9$ & $38.5$ & $29.0$ & $40.1$ & $37.9$ & $2.2$ \\
\midrule
\rowcolor{panelBg}
\multicolumn{13}{l}{\textit{Panel B: Textual-mode strict EM}} \\
\midrule
DeepSeek-V3.2
& \third{$44.5$} & $83.0$ & $53.9$ & $29.6$ & $11.8$ & $52.6$ & $47.3$ & $40.3$ & $33.8$ & \third{$45.9$} & \third{$43.2$} & $2.7$ \\
DeepSeek-R1
& $35.1$ & $77.6$ & $35.4$ & $18.2$ & $9.2$ & $42.4$ & $38.3$ & $31.6$ & $23.8$ & $35.2$ & $35.0$ & $0.2$ \\
DS-R1-Distill-32B
& $25.2$ & $60.2$ & $29.5$ & $9.1$ & $2.0$ & $30.6$ & $28.0$ & $20.3$ & $17.5$ & $25.8$ & $24.6$ & $1.2$ \\
DS-R1-Distill-70B
& $30.8$ & $68.3$ & $34.4$ & $16.5$ & $4.0$ & $35.1$ & $34.4$ & $29.1$ & $21.0$ & $31.1$ & $30.5$ & $0.6$ \\
Gemma-4-31B
& \second{$47.4$} & $84.8$ & $58.4$ & $29.9$ & $16.5$ & $54.4$ & $49.6$ & $43.8$ & $38.1$ & \second{$47.8$} & \second{$47.0$} & $0.8$ \\
Llama-3.1-8B
& $6.2$ & $13.0$ & $8.5$ & $2.5$ & $0.7$ & $6.7$ & $6.3$ & $6.2$ & $5.2$ & $8.2$ & $4.1$ & $4.1$ \\
Llama-3.3-70B
& $15.0$ & $28.2$ & $19.2$ & $10.0$ & $2.5$ & $16.7$ & $15.8$ & $15.0$ & $11.3$ & $15.4$ & $14.6$ & $0.8$ \\
Llama-4-Maverick
& $29.7$ & $60.3$ & $28.4$ & $17.2$ & $12.7$ & $33.0$ & $32.3$ & $30.3$ & $20.4$ & $29.6$ & $29.7$ & $-0.1$ \\
Llama-4-Scout
& $22.3$ & $41.4$ & $23.4$ & $13.7$ & $10.7$ & $27.1$ & $25.6$ & $20.0$ & $13.1$ & $22.6$ & $22.0$ & $0.6$ \\
Nemotron-3-Super
& $29.2$ & $70.1$ & $28.7$ & $10.0$ & $8.2$ & $35.3$ & $32.5$ & $26.3$ & $18.9$ & $30.2$ & $28.2$ & $2.0$ \\
o4-mini
& \first{48.9} & $83.5$ & $57.6$ & $34.4$ & $20.2$ & $57.8$ & $49.8$ & $46.9$ & $37.8$ & \first{50.4} & \first{47.5} & $2.9$ \\
Qwen2.5-72B
& $12.7$ & $31.7$ & $14.2$ & $4.5$ & $0.5$ & $13.5$ & $12.7$ & $12.5$ & $11.9$ & $13.9$ & $11.5$ & $2.4$ \\
Qwen2.5-7B
& $16.2$ & $33.9$ & $19.0$ & $10.5$ & $1.5$ & $18.8$ & $17.7$ & $12.8$ & $12.7$ & $17.8$ & $14.6$ & $3.2$ \\
Qwen3-32B
& $32.5$ & $68.1$ & $37.9$ & $20.2$ & $3.7$ & $36.0$ & $36.0$ & $29.4$ & $25.3$ & $32.6$ & $32.4$ & $0.2$ \\
Qwen3-Coder-30B
& $31.5$ & $59.4$ & $33.4$ & $21.9$ & $11.5$ & $36.2$ & $35.6$ & $30.0$ & $20.4$ & $32.7$ & $30.4$ & $2.3$ \\
QwQ-32B
& $34.9$ & $74.5$ & $44.3$ & $17.5$ & $3.4$ & $39.2$ & $37.0$ & $34.1$ & $25.4$ & $35.8$ & $34.0$ & $1.8$ \\
\cmidrule(l){2-13}
\textit{Average}
& $28.9$ & $58.6$ & $32.9$ & $16.6$ & $7.4$ & $33.5$ & $31.2$ & $26.8$ & $21.0$ & $29.7$ & $28.1$ & $1.6$ \\
\bottomrule
\end{tabularx}
\end{table*}

\subsection{Graph Reasoning across Graph Sizes}
\label{sec:appendix_graph_sizes}

\begin{table*}[t]
\centering
\caption{Exact Match (EM, \%) and Partial Credit (PC, \%) by graph size. The Retention columns report the largest-setting score divided by the smallest-setting score, i.e., $n{=}10{,}000$ divided by $n{=}10$. Panel~A1 reports code-based reasoning with \textit{file} graph loading; Panel~A2 reports code-based reasoning with \textit{inline} graph loading. Panel~B reports text-based reasoning, which is also inline. Bold entries within each panel mark the column maximum.}
\label{tab:size-mode-combined}
\scriptsize
\setlength{\tabcolsep}{2pt}
\begin{tabularx}{\linewidth}{l YY YY YY YY YY}
\toprule
                         & \multicolumn{2}{c}{10} & \multicolumn{2}{c}{100} & \multicolumn{2}{c}{1k} & \multicolumn{2}{c}{10k} & \multicolumn{2}{c}{Retention} \\
\cmidrule(lr){2-3}\cmidrule(lr){4-5}\cmidrule(lr){6-7}\cmidrule(lr){8-9}\cmidrule(lr){10-11}
Model                    & EM & PC & EM & PC & EM & PC & EM & PC & EM & PC \\
\midrule
\multicolumn{11}{l}{\textit{Panel A1: Coding-mode Performance --- file graph loading}} \\
\midrule
DeepSeek-V3.2            & $\mathbf{87.3}$ & $\mathbf{88.2}$ & $82.3$ & $83.1$ & $79.1$ & $80.3$ & $50.2$ & $50.4$ & $58\%$ & $57\%$ \\
DeepSeek-R1              & $86.9$ & $87.5$ & $\mathbf{84.2}$ & $\mathbf{84.1}$ & $79.7$ & $80.4$ & $50.5$ & $50.2$ & $58\%$ & $57\%$ \\
DeepSeek-R1-Distill-32B  & $81.7$ & $81.6$ & $76.6$ & $76.3$ & $73.3$ & $73.3$ & $49.5$ & $49.1$ & $61\%$ & $60\%$ \\
DeepSeek-R1-Distill-70B  & $72.4$ & $73.4$ & $69.6$ & $70.4$ & $67.0$ & $68.2$ & $44.0$ & $44.6$ & $61\%$ & $61\%$ \\
Gemma-4-31B              & $85.6$ & $86.3$ & $82.1$ & $82.2$ & $78.1$ & $78.5$ & $\mathbf{51.5}$ & $\mathbf{51.4}$ & $60\%$ & $60\%$ \\
Llama-3.1-8B             & $39.8$ & $42.0$ & $39.3$ & $41.8$ & $38.5$ & $40.6$ & $26.3$ & $27.9$ & $\mathbf{66\%}$ & $\mathbf{66\%}$ \\
Llama-3.3-70B            & $65.7$ & $67.5$ & $62.7$ & $64.2$ & $60.2$ & $61.4$ & $41.4$ & $42.1$ & $63\%$ & $62\%$ \\
Llama-4-Maverick         & $68.1$ & $69.4$ & $65.1$ & $65.9$ & $61.8$ & $63.2$ & $39.2$ & $40.1$ & $58\%$ & $58\%$ \\
Llama-4-Scout            & $59.2$ & $60.5$ & $54.9$ & $55.9$ & $51.8$ & $52.9$ & $35.6$ & $36.5$ & $60\%$ & $60\%$ \\
Nemotron-3-Super         & $86.6$ & $86.9$ & $83.7$ & $83.7$ & $\mathbf{80.4}$ & $\mathbf{80.7}$ & $\mathbf{51.5}$ & $51.0$ & $59\%$ & $59\%$ \\
o4-mini                  & $85.9$ & $86.2$ & $81.9$ & $82.1$ & $79.0$ & $79.4$ & $49.5$ & $49.3$ & $58\%$ & $57\%$ \\
Qwen2.5-72B              & $71.3$ & $73.2$ & $67.3$ & $68.7$ & $65.8$ & $67.6$ & $42.6$ & $42.8$ & $60\%$ & $58\%$ \\
Qwen2.5-7B               & $37.4$ & $40.2$ & $35.4$ & $38.3$ & $34.4$ & $37.8$ & $24.0$ & $25.9$ & $64\%$ & $64\%$ \\
Qwen3-32B                & $84.3$ & $84.2$ & $80.4$ & $80.4$ & $78.8$ & $78.9$ & $50.3$ & $49.8$ & $60\%$ & $59\%$ \\
Qwen3-Coder-30B          & $78.7$ & $80.2$ & $74.4$ & $75.1$ & $71.9$ & $73.0$ & $47.6$ & $47.7$ & $60\%$ & $59\%$ \\
\cmidrule(l){2-11}
\textit{Average}         & $72.7$ & $73.8$ & $69.3$ & $70.2$ & $66.7$ & $67.7$ & $43.6$ & $43.9$ & $60\%$ & $59\%$ \\
\midrule
\multicolumn{11}{l}{\textit{Panel A2: Coding-mode Performance --- inline graph loading}} \\
\midrule
DeepSeek-V3.2            & $89.5$ & $\mathbf{90.2}$ & $65.2$ & $65.3$ & $21.0$ & $21.8$ & $11.0$ & $11.2$ & $12\%$ & $12\%$ \\
DeepSeek-R1              & $87.7$ & $87.8$ & $56.8$ & $56.9$ & $26.2$ & $25.7$ & $21.5$ & $20.9$ & $25\%$ & $24\%$ \\
DeepSeek-R1-Distill-32B  & $67.4$ & $67.4$ & $48.6$ & $48.8$ & $16.6$ & $16.6$ & $3.0$ & $3.0$ & $4\%$ & $4\%$ \\
DeepSeek-R1-Distill-70B  & $68.6$ & $69.3$ & $48.4$ & $48.6$ & $25.0$ & $25.6$ & $6.4$ & $6.4$ & $9\%$ & $9\%$ \\
Gemma-4-31B              & $83.5$ & $84.2$ & $42.1$ & $42.4$ & $11.5$ & $11.5$ & $7.0$ & $7.0$ & $8\%$ & $8\%$ \\
Llama-3.1-8B             & $43.9$ & $47.4$ & $33.3$ & $34.5$ & $9.5$ & $11.1$ & $5.0$ & $5.5$ & $11\%$ & $12\%$ \\
Llama-3.3-70B            & $72.9$ & $74.0$ & $46.5$ & $47.6$ & $14.8$ & $15.6$ & $5.5$ & $5.7$ & $8\%$ & $8\%$ \\
Llama-4-Maverick         & $75.7$ & $75.9$ & $52.5$ & $52.8$ & $24.2$ & $24.7$ & $21.2$ & $21.0$ & $28\%$ & $28\%$ \\
Llama-4-Scout            & $61.8$ & $63.5$ & $46.5$ & $47.8$ & $22.9$ & $22.8$ & $18.3$ & $18.6$ & $\mathbf{30\%}$ & $29\%$ \\
Nemotron-3-Super         & $84.3$ & $84.7$ & $55.3$ & $55.1$ & $24.9$ & $24.8$ & $21.4$ & $21.5$ & $25\%$ & $25\%$ \\
o4-mini                  & $\mathbf{89.8}$ & $90.1$ & $\mathbf{69.1}$ & $\mathbf{69.1}$ & $\mathbf{43.1}$ & $\mathbf{42.8}$ & $\mathbf{27.2}$ & $\mathbf{26.7}$ & $\mathbf{30\%}$ & $\mathbf{30\%}$ \\
Qwen2.5-72B              & $73.8$ & $75.0$ & $55.6$ & $56.0$ & $20.2$ & $20.2$ & $3.7$ & $3.7$ & $5\%$ & $5\%$ \\
Qwen2.5-7B               & $42.6$ & $45.5$ & $36.9$ & $39.2$ & $22.2$ & $23.1$ & $4.5$ & $4.8$ & $11\%$ & $11\%$ \\
Qwen3-32B                & $73.5$ & $74.0$ & $42.4$ & $42.6$ & $17.6$ & $17.7$ & $6.1$ & $6.1$ & $8\%$ & $8\%$ \\
Qwen3-Coder-30B          & $80.7$ & $81.9$ & $56.8$ & $57.9$ & $16.3$ & $16.8$ & $10.8$ & $11.1$ & $13\%$ & $14\%$ \\
\cmidrule(l){2-11}
\textit{Average}         & $73.0$ & $74.1$ & $50.4$ & $51.0$ & $21.1$ & $21.4$ & $11.5$ & $11.5$ & $16\%$ & $16\%$ \\
\midrule
\multicolumn{11}{l}{\textit{Panel B: Textual-mode Performance}} \\
\midrule
DeepSeek-V3.2            & $83.0$ & $85.4$ & $53.9$ & $58.9$ & $29.6$ & $31.4$ & $11.8$ & $12.3$ & $14\%$ & $14\%$ \\
DeepSeek-R1              & $77.6$ & $77.7$ & $35.4$ & $36.3$ & $18.2$ & $18.6$ & $9.2$ & $9.6$ & $12\%$ & $12\%$ \\
DeepSeek-R1-Distill-32B  & $60.2$ & $62.3$ & $29.5$ & $29.9$ & $9.1$ & $9.7$ & $2.0$ & $2.0$ & $3\%$ & $3\%$ \\
DeepSeek-R1-Distill-70B  & $68.3$ & $70.7$ & $34.4$ & $35.9$ & $16.5$ & $18.2$ & $4.0$ & $4.3$ & $6\%$ & $6\%$ \\
Gemma-4-31B              & $\mathbf{84.8}$ & $\mathbf{87.2}$ & $\mathbf{58.4}$ & $\mathbf{62.3}$ & $29.9$ & $31.4$ & $16.5$ & $17.5$ & $19\%$ & $20\%$ \\
Llama-3.1-8B             & $13.0$ & $15.1$ & $8.5$ & $10.3$ & $2.5$ & $3.2$ & $0.7$ & $1.1$ & $5\%$ & $7\%$ \\
Llama-3.3-70B            & $28.2$ & $33.6$ & $19.2$ & $21.7$ & $10.0$ & $10.5$ & $2.5$ & $3.7$ & $9\%$ & $11\%$ \\
Llama-4-Maverick         & $60.3$ & $64.8$ & $28.4$ & $31.1$ & $17.2$ & $18.6$ & $12.7$ & $13.9$ & $21\%$ & $21\%$ \\
Llama-4-Scout            & $41.4$ & $49.8$ & $23.4$ & $27.0$ & $13.7$ & $14.6$ & $10.7$ & $11.4$ & $\mathbf{26\%}$ & $23\%$ \\
Nemotron-3-Super         & $70.1$ & $71.0$ & $28.7$ & $28.6$ & $10.0$ & $10.3$ & $8.2$ & $8.4$ & $12\%$ & $12\%$ \\
o4-mini                  & $83.5$ & $85.7$ & $57.6$ & $60.9$ & $\mathbf{34.4}$ & $\mathbf{36.4}$ & $\mathbf{20.2}$ & $\mathbf{20.6}$ & $24\%$ & $\mathbf{24\%}$ \\
Qwen2.5-72B              & $31.7$ & $35.2$ & $14.2$ & $15.3$ & $4.5$ & $5.3$ & $0.5$ & $0.6$ & $2\%$ & $2\%$ \\
Qwen2.5-7B               & $33.9$ & $40.3$ & $19.0$ & $22.2$ & $10.5$ & $13.1$ & $1.5$ & $2.2$ & $4\%$ & $5\%$ \\
Qwen3-32B                & $68.1$ & $70.1$ & $37.9$ & $40.1$ & $20.2$ & $21.5$ & $3.7$ & $4.2$ & $5\%$ & $6\%$ \\
Qwen3-Coder-30B          & $59.4$ & $66.4$ & $33.4$ & $38.0$ & $21.9$ & $24.5$ & $11.5$ & $12.5$ & $19\%$ & $19\%$ \\
\cmidrule(l){2-11}
\textit{Average}         & $58.6$ & $62.1$ & $32.9$ & $35.3$ & $16.6$ & $17.9$ & $7.4$ & $8.0$ & $13\%$ & $13\%$ \\
\bottomrule
\end{tabularx}
\end{table*}

\begin{table*}[t]
\centering
\caption{Coding-mode \textbf{file graph loading}: per-size strict and lenient Exact Match / Partial Credit (\%) and per-size error count. EM-S/PC-S are strict; EM-L/PC-L are lenient (excludes context-length, runtime, and missing-output cases from the denominator); Err is the per-size count of erroneous outputs. The header reports the total cases per size at full coverage.}
\label{tab:coding-file-strict-lenient}
\scriptsize
\setlength{\tabcolsep}{2.5pt}
\begin{tabularx}{\linewidth}{l YYYYY YYYYY YYYYY YYYYY}
\toprule
& \multicolumn{5}{c}{$n{=}10$ (total$=$404)} & \multicolumn{5}{c}{$n{=}100$ (total$=$404)} & \multicolumn{5}{c}{$n{=}1{,}000$ (total$=$404)} & \multicolumn{5}{c}{$n{=}10{,}000$ (total$=$404)} \\
\cmidrule(lr){2-6}\cmidrule(lr){7-11}\cmidrule(lr){12-16}\cmidrule(lr){17-21}
Model & EM-S & PC-S & EM-L & PC-L & Err & EM-S & PC-S & EM-L & PC-L & Err & EM-S & PC-S & EM-L & PC-L & Err & EM-S & PC-S & EM-L & PC-L & Err \\
\midrule
DeepSeek-V3.2 & 87.3 & 88.2 & 87.3 & 88.2 & 0 & 82.3 & 83.1 & 86.2 & 87.0 & 18 & 79.1 & 80.3 & 85.9 & 87.2 & 32 & 50.2 & 50.4 & 86.0 & 86.1 & 167 \\
DeepSeek-R1 & 86.9 & 87.5 & 86.9 & 87.5 & 0 & 84.2 & 84.1 & 88.2 & 88.1 & 17 & 79.7 & 80.4 & 86.9 & 87.7 & 31 & 50.5 & 50.2 & 87.1 & 86.5 & 157 \\
DeepSeek-R1-Distill-32B & 81.7 & 81.6 & 81.7 & 81.6 & 0 & 76.6 & 76.3 & 79.8 & 79.5 & 11 & 73.3 & 73.3 & 79.7 & 79.7 & 22 & 49.5 & 49.1 & 81.3 & 80.7 & 107 \\
DeepSeek-R1-Distill-70B & 72.4 & 73.4 & 72.4 & 73.4 & 0 & 69.6 & 70.4 & 72.5 & 73.3 & 14 & 67.0 & 68.2 & 72.6 & 73.9 & 27 & 44.0 & 44.6 & 74.2 & 75.1 & 143 \\
Gemma-4-31B & 85.6 & 86.3 & 85.6 & 86.3 & 0 & 82.1 & 82.2 & 85.9 & 86.1 & 18 & 78.1 & 78.5 & 84.9 & 85.3 & 32 & 51.5 & 51.4 & 88.1 & 87.9 & 167 \\
Llama-3.1-8B & 39.8 & 42.0 & 39.8 & 42.0 & 0 & 39.3 & 41.8 & 41.1 & 43.8 & 18 & 38.5 & 40.6 & 41.8 & 44.2 & 32 & 26.3 & 27.9 & 44.7 & 47.5 & 165 \\
Llama-3.3-70B & 65.7 & 67.5 & 65.7 & 67.5 & 0 & 62.7 & 64.2 & 65.4 & 67.0 & 17 & 60.2 & 61.4 & 65.2 & 66.5 & 31 & 41.4 & 42.1 & 70.2 & 71.4 & 164 \\
Llama-4-Maverick & 68.1 & 69.4 & 68.1 & 69.4 & 0 & 65.1 & 65.9 & 68.1 & 69.0 & 18 & 61.8 & 63.2 & 67.2 & 68.6 & 32 & 39.2 & 40.1 & 67.4 & 69.0 & 168 \\
Llama-4-Scout & 59.2 & 60.5 & 59.2 & 60.5 & 0 & 54.9 & 55.9 & 57.4 & 58.5 & 17 & 51.8 & 52.9 & 56.1 & 57.3 & 30 & 35.6 & 36.5 & 61.5 & 63.0 & 164 \\
Nemotron-3-Super & 86.6 & 86.9 & 86.6 & 86.9 & 0 & 83.7 & 83.7 & 87.5 & 87.5 & 16 & 80.4 & 80.7 & 87.5 & 87.8 & 30 & 51.5 & 51.0 & 87.9 & 87.0 & 152 \\
o4-mini & 85.9 & 86.2 & 85.9 & 86.2 & 0 & 81.9 & 82.1 & 85.8 & 86.0 & 18 & 79.0 & 79.4 & 85.8 & 86.3 & 32 & 49.5 & 49.3 & 84.7 & 84.3 & 168 \\
Qwen2.5-72B & 71.3 & 73.2 & 71.3 & 73.2 & 0 & 67.3 & 68.7 & 70.5 & 71.9 & 18 & 65.8 & 67.6 & 71.5 & 73.4 & 32 & 42.6 & 42.8 & 72.9 & 73.3 & 168 \\
Qwen2.5-7B & 37.4 & 40.2 & 37.4 & 40.2 & 0 & 35.4 & 38.3 & 37.0 & 40.1 & 18 & 34.4 & 37.8 & 37.4 & 41.0 & 32 & 24.0 & 25.9 & 41.1 & 44.3 & 168 \\
Qwen3-32B & 84.3 & 84.2 & 84.3 & 84.2 & 0 & 80.4 & 80.4 & 83.7 & 83.7 & 12 & 78.8 & 78.9 & 85.2 & 85.3 & 23 & 50.3 & 49.8 & 84.6 & 83.8 & 124 \\
Qwen3-Coder-30B & 78.7 & 80.2 & 78.7 & 80.2 & 0 & 74.4 & 75.1 & 78.0 & 78.6 & 18 & 71.9 & 73.0 & 78.2 & 79.4 & 32 & 47.6 & 47.7 & 81.9 & 82.1 & 167 \\
\bottomrule
\end{tabularx}
\end{table*}

\begin{table*}[t]
\centering
\caption{Coding-mode \textbf{inline graph loading}: per-size strict and lenient EM/PC (\%) and per-size error count. Columns follow Table~\ref{tab:coding-file-strict-lenient}.}
\label{tab:coding-inline-strict-lenient}
\scriptsize
\setlength{\tabcolsep}{2.5pt}
\begin{tabularx}{\linewidth}{l YYYYY YYYYY YYYYY YYYYY}
\toprule
& \multicolumn{5}{c}{$n{=}10$ (total$=$404)} & \multicolumn{5}{c}{$n{=}100$ (total$=$404)} & \multicolumn{5}{c}{$n{=}1{,}000$ (total$=$404)} & \multicolumn{5}{c}{$n{=}10{,}000$ (total$=$404)} \\
\cmidrule(lr){2-6}\cmidrule(lr){7-11}\cmidrule(lr){12-16}\cmidrule(lr){17-21}
Model & EM-S & PC-S & EM-L & PC-L & Err & EM-S & PC-S & EM-L & PC-L & Err & EM-S & PC-S & EM-L & PC-L & Err & EM-S & PC-S & EM-L & PC-L & Err \\
\midrule
DeepSeek-V3.2 & 89.5 & 90.2 & 89.5 & 90.2 & 0 & 65.2 & 65.3 & 69.6 & 69.6 & 25 & 21.0 & 21.8 & 35.6 & 37.0 & 164 & 11.0 & 11.2 & 52.4 & 53.3 & 316 \\
DeepSeek-R1 & 87.7 & 87.8 & 89.8 & 89.9 & 9 & 56.8 & 56.9 & 66.0 & 66.1 & 53 & 26.2 & 25.7 & 48.1 & 47.1 & 174 & 21.5 & 20.9 & 71.9 & 70.2 & 268 \\
DeepSeek-R1-Distill-32B & 67.4 & 67.4 & 81.7 & 81.7 & 58 & 48.6 & 48.8 & 68.2 & 68.5 & 95 & 16.6 & 16.6 & 66.3 & 66.3 & 248 & 3.0 & 3.0 & 76.9 & 76.9 & 318 \\
DeepSeek-R1-Distill-70B & 68.6 & 69.3 & 76.8 & 77.6 & 40 & 48.4 & 48.6 & 59.7 & 59.9 & 71 & 25.0 & 25.6 & 50.3 & 51.5 & 189 & 6.4 & 6.4 & 64.9 & 65.2 & 339 \\
Gemma-4-31B & 83.5 & 84.2 & 83.5 & 84.2 & 0 & 42.1 & 42.4 & 44.9 & 45.2 & 25 & 11.5 & 11.5 & 19.5 & 19.5 & 163 & 7.0 & 7.0 & 21.7 & 21.7 & 270 \\
Llama-3.1-8B & 43.9 & 47.4 & 43.9 & 47.4 & 0 & 33.3 & 34.5 & 37.9 & 39.2 & 48 & 9.5 & 11.1 & 18.4 & 21.4 & 192 & 5.0 & 5.5 & 22.2 & 24.4 & 309 \\
Llama-3.3-70B & 72.9 & 74.0 & 72.9 & 74.0 & 0 & 46.5 & 47.6 & 49.7 & 51.0 & 26 & 14.8 & 15.6 & 25.0 & 26.4 & 162 & 5.5 & 5.7 & 24.4 & 25.0 & 308 \\
Llama-4-Maverick & 75.7 & 75.9 & 75.7 & 75.9 & 0 & 52.5 & 52.8 & 55.0 & 55.3 & 18 & 24.2 & 24.7 & 33.3 & 33.9 & 109 & 21.2 & 21.0 & 36.6 & 36.2 & 168 \\
Llama-4-Scout & 61.8 & 63.5 & 61.8 & 63.5 & 0 & 46.5 & 47.8 & 49.1 & 50.4 & 21 & 22.9 & 22.8 & 38.2 & 38.1 & 160 & 18.3 & 18.6 & 37.2 & 37.7 & 202 \\
Nemotron-3-Super & 84.3 & 84.7 & 86.9 & 87.3 & 11 & 55.3 & 55.1 & 69.9 & 69.6 & 77 & 24.9 & 24.8 & 54.1 & 53.8 & 199 & 21.4 & 21.5 & 74.5 & 74.8 & 263 \\
o4-mini & 89.8 & 90.1 & 89.8 & 90.1 & 0 & 69.1 & 69.1 & 82.0 & 82.0 & 63 & 43.1 & 42.8 & 86.1 & 85.4 & 200 & 27.2 & 26.7 & 80.7 & 79.3 & 266 \\
Qwen2.5-72B & 73.8 & 75.0 & 73.8 & 75.0 & 0 & 55.6 & 56.0 & 61.6 & 62.0 & 39 & 20.2 & 20.2 & 44.0 & 44.1 & 217 & 3.7 & 3.7 & 65.2 & 65.2 & 378 \\
Qwen2.5-7B & 42.6 & 45.5 & 42.6 & 45.5 & 0 & 36.9 & 39.2 & 40.0 & 42.4 & 31 & 22.2 & 23.1 & 41.8 & 43.5 & 188 & 4.5 & 4.8 & 51.4 & 54.7 & 366 \\
Qwen3-32B & 73.5 & 74.0 & 84.2 & 84.7 & 44 & 42.4 & 42.6 & 63.1 & 63.5 & 114 & 17.6 & 17.7 & 47.3 & 47.7 & 218 & 6.1 & 6.1 & 87.5 & 87.5 & 323 \\
Qwen3-Coder-30B & 80.7 & 81.9 & 80.7 & 81.9 & 0 & 56.8 & 57.9 & 60.6 & 61.8 & 25 & 16.3 & 16.8 & 27.8 & 28.6 & 164 & 10.8 & 11.1 & 31.6 & 32.6 & 262 \\
\bottomrule
\end{tabularx}
\end{table*}

\begin{table*}[t]
\centering
\caption{Textual-mode: per-size strict and lenient EM/PC (\%) and per-size error count. Columns follow Table~\ref{tab:coding-file-strict-lenient}.}
\label{tab:textual-strict-lenient}
\scriptsize
\setlength{\tabcolsep}{2.5pt}
\begin{tabularx}{\linewidth}{l YYYYY YYYYY YYYYY YYYYY}
\toprule
& \multicolumn{5}{c}{$n{=}10$ (total$=$404)} & \multicolumn{5}{c}{$n{=}100$ (total$=$404)} & \multicolumn{5}{c}{$n{=}1{,}000$ (total$=$404)} & \multicolumn{5}{c}{$n{=}10{,}000$ (total$=$404)} \\
\cmidrule(lr){2-6}\cmidrule(lr){7-11}\cmidrule(lr){12-16}\cmidrule(lr){17-21}
Model & EM-S & PC-S & EM-L & PC-L & Err & EM-S & PC-S & EM-L & PC-L & Err & EM-S & PC-S & EM-L & PC-L & Err & EM-S & PC-S & EM-L & PC-L & Err \\
\midrule
DeepSeek-V3.2 & 83.0 & 85.4 & 83.0 & 85.4 & 0 & 53.9 & 58.9 & 57.3 & 62.6 & 24 & 29.6 & 31.4 & 50.4 & 53.5 & 165 & 11.8 & 12.3 & 56.0 & 58.5 & 315 \\
DeepSeek-R1 & 77.6 & 77.7 & 77.6 & 77.7 & 0 & 35.4 & 36.3 & 37.7 & 38.6 & 24 & 18.2 & 18.6 & 30.9 & 31.6 & 165 & 9.2 & 9.6 & 44.0 & 45.8 & 317 \\
DeepSeek-R1-Distill-32B & 60.2 & 62.3 & 60.7 & 62.8 & 3 & 29.5 & 29.9 & 33.5 & 34.1 & 48 & 9.1 & 9.7 & 25.0 & 26.7 & 253 & 2.0 & 2.0 & 34.8 & 35.1 & 374 \\
DeepSeek-R1-Distill-70B & 68.3 & 70.7 & 68.3 & 70.7 & 0 & 34.4 & 35.9 & 36.6 & 38.2 & 24 & 16.5 & 18.2 & 28.0 & 31.0 & 165 & 4.0 & 4.3 & 17.8 & 19.3 & 311 \\
Gemma-4-31B & 84.8 & 87.2 & 84.8 & 87.2 & 0 & 58.4 & 62.3 & 61.3 & 65.4 & 19 & 29.9 & 31.4 & 49.2 & 51.5 & 157 & 16.5 & 17.5 & 33.5 & 35.7 & 204 \\
Llama-3.1-8B & 13.0 & 15.1 & 13.0 & 15.1 & 0 & 8.5 & 10.3 & 9.0 & 11.0 & 24 & 2.5 & 3.2 & 4.2 & 5.5 & 165 & 0.7 & 1.1 & 3.3 & 5.0 & 311 \\
Llama-3.3-70B & 28.2 & 33.6 & 28.2 & 33.6 & 0 & 19.2 & 21.7 & 20.4 & 23.0 & 24 & 10.0 & 10.5 & 16.9 & 17.8 & 165 & 2.5 & 3.7 & 11.1 & 16.4 & 311 \\
Llama-4-Maverick & 60.3 & 64.8 & 60.3 & 64.8 & 0 & 28.4 & 31.1 & 29.9 & 32.7 & 20 & 17.2 & 18.6 & 21.7 & 23.4 & 83 & 12.7 & 13.9 & 22.2 & 24.2 & 171 \\
Llama-4-Scout & 41.4 & 49.8 & 41.4 & 49.8 & 0 & 23.4 & 27.0 & 24.5 & 28.3 & 18 & 13.7 & 14.6 & 17.9 & 19.0 & 93 & 10.7 & 11.4 & 18.5 & 19.7 & 169 \\
Nemotron-3-Super & 70.1 & 71.0 & 70.1 & 71.0 & 0 & 28.7 & 28.6 & 30.5 & 30.4 & 24 & 10.0 & 10.3 & 16.9 & 17.5 & 165 & 8.2 & 8.4 & 24.3 & 24.8 & 265 \\
o4-mini & 83.5 & 85.7 & 83.5 & 85.7 & 0 & 57.6 & 60.9 & 61.1 & 64.6 & 23 & 34.4 & 36.4 & 58.0 & 61.4 & 163 & 20.2 & 20.6 & 58.7 & 60.0 & 263 \\
Qwen2.5-72B & 31.7 & 35.2 & 31.7 & 35.2 & 0 & 14.2 & 15.3 & 15.7 & 16.9 & 38 & 4.5 & 5.3 & 9.5 & 11.1 & 211 & 0.5 & 0.6 & 8.7 & 11.2 & 378 \\
Qwen2.5-7B & 33.9 & 40.3 & 33.9 & 40.3 & 0 & 19.0 & 22.2 & 20.6 & 24.1 & 32 & 10.5 & 13.1 & 19.6 & 24.6 & 187 & 1.5 & 2.2 & 17.1 & 25.1 & 366 \\
Qwen3-32B & 68.1 & 70.1 & 68.1 & 70.1 & 0 & 37.9 & 40.1 & 40.3 & 42.7 & 24 & 20.2 & 21.5 & 34.3 & 36.5 & 165 & 3.7 & 4.2 & 31.9 & 35.6 & 354 \\
Qwen3-Coder-30B & 59.4 & 66.4 & 59.4 & 66.4 & 0 & 33.4 & 38.0 & 35.5 & 40.5 & 24 & 21.9 & 24.5 & 37.3 & 41.7 & 165 & 11.5 & 12.5 & 33.8 & 36.9 & 265 \\
\bottomrule
\end{tabularx}
\end{table*}

\begin{table*}[t]
\centering
\caption{Error analysis at $n{=}10{,}000$ for DeepSeek-V3.2.
}
\label{tab:error-analysis-n10k}
\footnotesize
\setlength{\tabcolsep}{4pt}
\begin{tabularx}{\linewidth}{l r YYY YYY}
\toprule
                       & \multirow{2}{*}{Total}
                       & \multicolumn{3}{c}{Outcome split (\%)}
                       & \multicolumn{3}{c}{Error breakdown (\%)} \\
\cmidrule(lr){3-5}\cmidrule(lr){6-8}
Mode                   &       & Correct & Incorrect & Error & Timeout & Ctx-len & Other \\
\midrule
Coding (file)          & $404$ & $50.2$  & $6.7$     & $43.0$ & $42.8$  & ---     & $0.2$ \\
Coding (inline)        & $404$ & $11.0$  & $1.5$     & $87.5$ & ---     & $78.8$  & $8.8$ \\
Textual                & $404$ & $11.8$  & $9.3$     & $78.9$ & ---     & $78.9$  & --- \\
\bottomrule
\end{tabularx}
\end{table*}

{\dataset} covers graph inputs of different scales and further extends to graphs with $10{,}000$ nodes, enabling a stress test of LLMs' long-context graph reasoning capability. This section studies the effect of graph size on reasoning performance, aiming to understand how LLM graph reasoning changes when inputs scale from small graphs to large and even ultra-large graph settings.

\textbf{Experimental setup.}
Each composite task in {\dataset} is instantiated at four node sizes, $n\in\{10,100,1{,}000,10{,}000\}$, yielding a $202$-task $\times$ $4$-size matrix for each (model, reasoning mode) combination. We evaluate two reasoning modes: code-based and text-based reasoning. For code-based reasoning, we further consider both file-based and inline graph loading. Each reported cell averages over the remaining benchmark dimensions.
We report two metrics. \emph{Exact Match} (EM) measures the fraction of instances whose prediction exactly matches the ground-truth answer, while \emph{Partial Credit} (PC) gives graded credit for near-correct outputs, such as partially correct lists with missing or misordered elements. Both metrics are computed under two denominators: a \emph{strict} view, where every benchmark instance is counted and invalid outputs such as syntax errors, context-window rejections, runtime/timeout failures, and missing predictions are treated as wrong; and a \emph{lenient} view, where invalid outputs are excluded and the metric is computed only over parseable predictions. We use strict EM/PC as the headline metrics throughout the paper, while lenient EM/PC help separate answer quality from infrastructure and context-window failures.
Table~\ref{tab:size-mode-combined} reports strict EM and PC by graph size across reasoning modes. Tables~\ref{tab:coding-file-strict-lenient}, \ref{tab:coding-inline-strict-lenient}, and~\ref{tab:textual-strict-lenient} provide per-mode strict and lenient EM/PC together with per-size error counts. Table~\ref{tab:error-analysis-n10k} further decomposes the $n{=}10{,}000$ errors of DeepSeek-V3.2 into context-length, timeout, and other failure causes.

\textbf{Finding 1: Performance decreases monotonically as graph size increases.}
Across all (model, mode) combinations in Table~\ref{tab:size-mode-combined}, EM decreases monotonically as graph size grows from $n{=}10$ to $n{=}10{,}000$; PC shows the same trend, so we focus on EM. Averaged across models, code-based file reasoning decreases from $72.7$ to $43.6$ EM, retaining $60\%$ of its small-graph performance. In contrast, code-based inline reasoning drops from $73.0$ to $11.5$ EM, with $16\%$ retention, and text-based reasoning drops from $58.6$ to $7.4$ EM, with $13\%$ retention. These results indicate that graph size is a stable difficulty dimension, but its impact is strongly mediated by the graph-loading format. As graphs grow, the input context becomes longer and contains more structural information to parse, maintain, and reason over, which increases reasoning difficulty and leads to lower performance.

\textbf{Finding 2: Inline graph loading nearly collapses on ultra-large graphs.}
The two inline settings (Panels A2 and B in Table~\ref{tab:size-mode-combined}) lose more than $80\%$ of their EM scores at $n=10{,}000$ compared with $n=10$. The average EM of code-based inline reasoning decreases from $73.0$ at $n=10$ to $11.5$ at $n=10{,}000$, retaining only $16\%$ of its original performance. Similarly, the average EM of text-based inline reasoning decreases from $58.6$ to $7.4$, retaining only $13\%$. Several models almost completely collapse: the inline EM of Qwen2.5-72B drops from $73.8$ to $3.7$, that of Gemma-4-31B drops from $83.5$ to $7.0$, and that of DeepSeek-R1-Distill-32B drops from $67.4$ to $3.0$. This collapse is mainly caused by serializing the entire node and edge lists into the prompt, which forces the model to process the full graph directly. When $n$ reaches the range of $10^3$ to $10^4$, the serialized graph can exceed the context windows of many models.

\textbf{Finding 3: File-based code reasoning is substantially more robust to graph-size scaling than inline code-based reasoning.}
File-based code reasoning retains approximately $60\%$ of its small-graph EM at $n{=}10{,}000$ (average EM: $72.7 \to 43.6$), whereas inline code-based reasoning retains only about $16\%$ ($73.0 \to 11.5$). Although the two settings achieve nearly identical EM at $n{=}10$ ($72.7$ for file-based reasoning vs.\ $73.0$ for inline reasoning), the performance gap increases to $18.9$ percentage points at $n{=}100$, $45.6$ at $n{=}1{,}000$, and $32.1$ at $n{=}10{,}000$. This contrast suggests that file-based graph loading effectively decouples input size from prompt length: the model only receives the path to an edge-list file and generates code to read and process the graph. In contrast, inline graph loading serializes the entire graph into the prompt, forcing the model to reason over a long textual representation and thereby converting graph reasoning into a long-context processing bottleneck.

\textbf{Finding 4: Code-based reasoning consistently outperforms text-based reasoning across graph sizes, even under inline graph loading.}
In Table~\ref{tab:size-mode-combined}, the average per-model EM gap between code-based reasoning and text-based reasoning, where code-based reasoning averages over the file-based and inline settings, is $14.3$ percentage points at $n{=}10$, increases to $27.0$ at $n{=}100$, remains comparable at $27.3$ at $n{=}1{,}000$, and is still $20.2$ at $n{=}10{,}000$. The increase from $n{=}10$ to $n{=}100$ reflects the rapid degradation of text-based reasoning once graph inputs exceed the scale that can be reliably handled through short chain-of-thought reasoning. The slight narrowing at $n{=}10{,}000$ suggests that code-based reasoning also begins to encounter limitations from context length, input parsing, and program execution errors, particularly in the inline setting. Across all $60$ paired model-size comparisons, the per-model averaged code EM exceeds the corresponding text-based EM in every case. These results indicate that translating graph reasoning into executable programs and delegating computation to an interpreter is more reliable than directly simulating graph algorithms in natural language, especially for medium- and large-scale graphs.

\textbf{Finding 5: At $n{=}10{,}000$, errors become the dominant factor, primarily driven by context-window rejection in inline modes and timeouts in the file-based mode.}
Tables~\ref{tab:coding-file-strict-lenient}, \ref{tab:coding-inline-strict-lenient}, \ref{tab:textual-strict-lenient}, and~\ref{tab:error-analysis-n10k} report the scale-specific error counts for the DeepSeek-V3.2 plain (i.e., zero-shot CoT) baseline at $n{=}10{,}000$, together with a breakdown by error source. In the coding-file setting, the error rate reaches $43.0\%$ across all $404$ evaluation instances, with nearly all errors attributed to timeouts. Since prompts in the file-based setting are short, context length is not the main bottleneck; instead, the LLM-generated programs must finish execution on graphs with $10{,}000$ nodes within the $60$-second wall-clock budget imposed by the evaluation framework, and most failures are caused by programs that hang during execution. In the coding-inline setting, the error rate increases sharply to $87.5\%$ across all $400$ evaluation instances, of which $78.8\%$ are caused by API-level context-window rejection: at $n{=}10{,}000$, inline prompts often exceed the $131$K-token context window supported by DeepSeek-V3.2. The textual setting exhibits the same pattern: $78.9\%$ of all $404$ evaluation instances result in errors, all of which are attributable to context-length limitations. This asymmetry between the coding-file setting, where failures are dominated by timeouts, and the two inline settings, where failures are dominated by context-length limits, explains the underlying mechanism behind Findings~2 and~3: under inline graph loading, tasks encounter the long-context processing bottleneck before reaching the graph-reasoning bottleneck.

\begin{takeawaybox}
In {\dataset}, graph reasoning performance decreases monotonically with graph size across all evaluated models. However, the magnitude of this degradation is governed more by the graph loading than by the model itself. Text-based reasoning and inline code-based reasoning both collapse under ultra-long graph inputs: most models lose $80$--$95\%$ of their EM when scaling from $n{=}10$ to $n{=}10{,}000$, and inline-code mode error rates approach $90\%$ at $n{=}10{,}000$, almost entirely due to context-window rejections. In contrast, file-code-based reasoning is substantially more scalable, retaining approximately $60\%$ of its small-graph EM at the largest graph size because it decouples prompt length from graph size. The failures at $n{=}10{,}000$ are primarily caused by timeouts during program execution rather than context overflows. These results establish graph size as a critical complexity dimension for evaluating LLM-based graph reasoning, particularly for exposing the fragility of inline graph encodings under long-context inputs.
\end{takeawaybox}

\subsection{Long-Context Stress Test}
\label{sec:appendix_stress}

\begin{table*}[t]
\centering
\caption{Relative graph sizes for the stress-test subset. Each task is instantiated at four budgets that target a fraction of the group's model context window: \emph{tenth} ($0.1\times$), \emph{quarter} ($0.25\times$), \emph{half} ($0.5\times$), and \emph{full} ($0.98\times$). The \emph{Average} row reports the per-column mean over the selected tasks.}
\label{tab:stress-sizes}
\footnotesize
\setlength{\tabcolsep}{3pt}
\begin{tabularx}{0.7\linewidth}{l rrrr}
\toprule
Task & tenth & quarter & half & full \\
\midrule
\multicolumn{5}{l}{\textit{Panel A: Group A --- Qwen2.5-72B / Qwen2.5-7B ($32{,}768$-token window)}}\\
\midrule
path\_existence                           &  $215$ &   $537$ & $1{,}066$ & $1{,}966$ \\
graph\_connectivity                       &  $199$ &   $618$ & $1{,}367$ & $2{,}367$ \\
triangle\_detection                       &   $53$ &    $90$ &    $126$ &    $184$ \\
max\_triangle\_sum                         &   $36$ &    $63$ &     $92$ &    $131$ \\
distance\_k                               &  $147$ &   $387$ &    $798$ & $1{,}476$ \\
distance\_threshold                       &   $32$ &    $53$ &     $76$ &    $107$ \\
local\_clustering\_to\_island\_count          &   $52$ &    $89$ &    $130$ &    $186$ \\
k\_core\_of\_largest\_connected\_component   &   $54$ &   $106$ &    $186$ &    $242$ \\
path\_existence\_and\_bipartite\_check       &   $54$ &    $91$ &    $131$ &    $187$ \\
cycle\_and\_star\_center\_decision           &  $295$ &   $830$ & $1{,}541$ & $2{,}879$ \\
redundant\_edge\_in\_acyclic\_course\_graph   &  $297$ &   $832$ & $1{,}545$ & $2{,}882$ \\
distance\_k\_then\_walls\_and\_gates          &   $51$ &    $89$ &    $129$ &    $184$ \\
\cmidrule(l){2-5}
\textit{Average}                          &  $124$ &   $315$ &    $599$ & $1{,}066$ \\
\midrule
\multicolumn{5}{l}{\textit{Panel B: Group B --- DeepSeek-R1 / DeepSeek-V3.2 ($131{,}072$-token window)}}\\
\midrule
path\_existence                           &   $906$ & $1{,}966$ & $3{,}901$ &  $7{,}071$ \\
graph\_connectivity                       &   $914$ & $2{,}192$ & $4{,}161$ &  $8{,}790$ \\
triangle\_detection                       &   $115$ &    $184$ &    $262$ &     $374$ \\
max\_triangle\_sum                         &    $81$ &    $131$ &    $186$ &     $266$ \\
distance\_k                               &   $677$ & $1{,}575$ & $2{,}719$ &  $5{,}632$ \\
distance\_threshold                       &    $68$ &    $108$ &    $153$ &     $217$ \\
local\_clustering\_to\_island\_count          &   $116$ &    $186$ &    $264$ &     $376$ \\
k\_core\_of\_largest\_connected\_component   &   $165$ &    $337$ &    $346$ &     $589$ \\
path\_existence\_and\_bipartite\_check       &   $116$ &    $187$ &    $265$ &     $376$ \\
cycle\_and\_star\_center\_decision           & $1{,}274$ & $2{,}880$ & $5{,}556$ & $10{,}908$ \\
redundant\_edge\_in\_acyclic\_course\_graph   & $1{,}276$ & $2{,}879$ & $5{,}558$ & $10{,}907$ \\
distance\_k\_then\_walls\_and\_gates          &   $115$ &    $187$ &    $265$ &     $375$ \\
\cmidrule(l){2-5}
\textit{Average}                          &   $485$ & $1{,}068$ & $1{,}970$ &  $3{,}823$ \\
\midrule
\multicolumn{5}{l}{\textit{Panel C: Group C --- Gemma-4-31B / Nemotron-3-Super ($262{,}144$-token window)}}\\
\midrule
path\_existence                           & $1{,}479$ & $3{,}901$ &  $7{,}071$ & $14{,}250$ \\
graph\_connectivity                       & $1{,}673$ & $5{,}360$ &  $9{,}150$ & $15{,}336$ \\
triangle\_detection                       &    $163$ &    $262$ &     $375$ &     $530$ \\
max\_triangle\_sum                         &    $117$ &    $186$ &     $266$ &     $377$ \\
distance\_k                               & $1{,}370$ & $2{,}994$ &  $5{,}925$ & $10{,}238$ \\
distance\_threshold                       &     $96$ &    $153$ &     $217$ &     $307$ \\
local\_clustering\_to\_island\_count          &    $166$ &    $264$ &     $376$ &     $532$ \\
k\_core\_of\_largest\_connected\_component   &    $326$ &    $360$ &     $406$ &     $551$ \\
path\_existence\_and\_bipartite\_check       &    $167$ &    $265$ &     $376$ &     $532$ \\
cycle\_and\_star\_center\_decision           & $2{,}344$ & $5{,}556$ & $10{,}908$ & $21{,}612$ \\
redundant\_edge\_in\_acyclic\_course\_graph   & $2{,}346$ & $5{,}561$ & $10{,}906$ & $21{,}611$ \\
distance\_k\_then\_walls\_and\_gates          &    $164$ &    $265$ &     $374$ &     $532$ \\
\cmidrule(l){2-5}
\textit{Average}                          &    $868$ & $2{,}094$ &  $3{,}863$ &  $7{,}201$ \\
\bottomrule
\end{tabularx}
\end{table*}

\begin{table*}[t]
\centering
\caption{Stress-test strict Exact Match (EM, \%) and Partial Credit (PC, \%) for the six target models. The leftmost \emph{$n{=}10$} block reports the same six models on the same $12$-task subset under the canonical absolute scan ($n{=}10$ graphs); we include it as a model-agnostic non-stress baseline. The remaining four blocks report the four relative context-usage levels ($0.1\times / 0.25\times / 0.5\times / 0.98\times$ of each model's context window). The \emph{Average} row reports the per-column mean over the six models. Bold entries mark the column maximum within each panel.}
\label{tab:stress-em-pc}
\scriptsize
\setlength{\tabcolsep}{2pt}
\begin{tabularx}{\linewidth}{l YY YY YY YY YY}
\toprule
                  & \multicolumn{2}{c}{$n{=}10$}
                  & \multicolumn{2}{c}{tenth ($0.1\times$)}
                  & \multicolumn{2}{c}{quarter ($0.25\times$)}
                  & \multicolumn{2}{c}{half ($0.5\times$)}
                  & \multicolumn{2}{c}{full ($1.0\times$)} \\
\cmidrule(lr){2-3}\cmidrule(lr){4-5}\cmidrule(lr){6-7}\cmidrule(lr){8-9}\cmidrule(lr){10-11}
Model              & EM & PC & EM & PC & EM & PC & EM & PC & EM & PC \\
\midrule
\multicolumn{11}{l}{\textit{Panel A: Coding-mode --- file graph loading}}\\
\midrule
DeepSeek-R1        & $85.7$ & $\mathbf{90.5}$ & $75.0$ & $77.1$ & $75.0$ & $77.1$ & $\mathbf{83.3}$ & $\mathbf{85.4}$ & $75.0$ & $77.1$ \\
DeepSeek-V3.2      & $\mathbf{87.5}$ & $87.5$ & $62.5$ & $62.5$ & $58.3$ & $58.3$ & $58.3$ & $58.3$ & $54.2$ & $54.2$ \\
Gemma-4-31B        & $79.2$ & $79.2$ & $83.3$ & $85.4$ & $75.0$ & $77.1$ & $75.0$ & $77.1$ & $75.0$ & $78.1$ \\
Nemotron-3-Super   & $78.3$ & $80.4$ & $\mathbf{87.5}$ & $\mathbf{87.5}$ & $\mathbf{79.2}$ & $\mathbf{79.2}$ & $79.2$ & $79.2$ & $\mathbf{79.2}$ & $\mathbf{79.2}$ \\
Qwen2.5-72B        & $70.8$ & $71.8$ & $75.0$ & $75.0$ & $70.8$ & $72.9$ & $66.7$ & $66.7$ & $66.7$ & $66.7$ \\
Qwen2.5-7B         & $37.5$ & $42.8$ & $37.5$ & $41.8$ & $37.5$ & $40.3$ & $37.5$ & $37.5$ & $37.5$ & $42.8$ \\
\cmidrule(l){2-11}
\textit{Average}   & $73.2$ & $75.4$ & $70.1$ & $71.6$ & $66.0$ & $67.5$ & $66.7$ & $67.4$ & $64.6$ & $66.4$ \\
\midrule
\multicolumn{11}{l}{\textit{Panel B: Coding-mode --- inline graph loading}}\\
\midrule
DeepSeek-R1        & $79.2$ & $83.3$ & $75.0$ & $77.8$ & $\mathbf{54.2}$ & $56.3$ & $\mathbf{54.2}$ & $\mathbf{56.3}$ & $\mathbf{50.0}$ & $\mathbf{50.0}$ \\
DeepSeek-V3.2      & $83.3$ & $85.4$ & $58.3$ & $58.3$ & $33.3$ & $33.3$ & $8.3$ & $8.3$ & $4.2$ & $4.2$ \\
Gemma-4-31B        & $79.2$ & $79.2$ & $25.0$ & $25.0$ & $0.0$ & $0.0$ & $0.0$ & $0.0$ & $0.0$ & $0.0$ \\
Nemotron-3-Super   & $\mathbf{87.0}$ & $\mathbf{87.0}$ & $41.7$ & $41.7$ & $41.7$ & $41.7$ & $29.2$ & $29.2$ & $12.5$ & $12.5$ \\
Qwen2.5-72B        & $75.0$ & $77.1$ & $\mathbf{83.3}$ & $\mathbf{83.3}$ & $50.0$ & $\mathbf{60.6}$ & $8.3$ & $8.3$ & $4.2$ & $4.2$ \\
Qwen2.5-7B         & $50.0$ & $52.5$ & $33.3$ & $39.1$ & $20.8$ & $22.9$ & $4.2$ & $4.2$ & $0.0$ & $0.0$ \\
\cmidrule(l){2-11}
\textit{Average}   & $75.6$ & $77.4$ & $52.8$ & $54.2$ & $33.3$ & $35.8$ & $17.4$ & $17.7$ & $11.8$ & $11.8$ \\
\midrule
\multicolumn{11}{l}{\textit{Panel C: Textual-mode}}\\
\midrule
DeepSeek-R1        & $\mathbf{62.5}$ & $62.5$ & $\mathbf{50.0}$ & $\mathbf{53.6}$ & $\mathbf{41.7}$ & $\mathbf{41.7}$ & $\mathbf{29.2}$ & $29.2$ & $20.8$ & $20.8$ \\
DeepSeek-V3.2      & $\mathbf{62.5}$ & $\mathbf{75.9}$ & $33.3$ & $42.4$ & $29.2$ & $31.2$ & $\mathbf{29.2}$ & $\mathbf{30.0}$ & $\mathbf{25.0}$ & $\mathbf{27.5}$ \\
Gemma-4-31B        & $\mathbf{62.5}$ & $72.2$ & $41.7$ & $47.2$ & $37.5$ & $39.0$ & $16.7$ & $25.4$ & $4.2$ & $4.2$ \\
Nemotron-3-Super   & $54.2$ & $56.3$ & $16.7$ & $16.7$ & $12.5$ & $12.5$ & $4.2$ & $4.2$ & $0.0$ & $0.0$ \\
Qwen2.5-72B        & $20.8$ & $23.7$ & $12.5$ & $12.5$ & $8.3$ & $8.3$ & $4.2$ & $4.2$ & $0.0$ & $0.0$ \\
Qwen2.5-7B         & $20.8$ & $25.6$ & $25.0$ & $25.0$ & $16.7$ & $20.8$ & $0.0$ & $0.0$ & $0.0$ & $0.0$ \\
\cmidrule(l){2-11}
\textit{Average}   & $47.2$ & $52.7$ & $29.9$ & $32.9$ & $24.3$ & $25.6$ & $13.9$ & $15.5$ & $8.3$ & $8.8$ \\
\bottomrule
\end{tabularx}
\end{table*}

Tables~\ref{tab:size-mode-combined}--\ref{tab:textual-strict-lenient} characterize how performance scales with absolute graph size on a fixed grid $\{10, 100, 1{,}000, 10{,}000\}$ shared by all models. This setting reveals clear scale dependence. However, a graph that is small in absolute size may still occupy most of a $32$K-token context window, whereas the same graph occupies only a small fraction of a model with a $1$M-token context window. To isolate this factor, we further conduct model-specific stress tests in which graph size is calibrated relative to each model's context length.

\textbf{Experimental setup.}
We evaluate six models and organize them into three context-length-matched pairs:
Group~A includes Qwen2.5-72B and Qwen2.5-7B, both with a $32{,}768$-token context window;
Group~B includes DeepSeek-R1 and DeepSeek-V3.2, both with a $131{,}072$-token context window;
and Group~C includes Gemma-4-31B and Nemotron-3-Super, both with a $262{,}144$-token context window.
For each group, we generate a shared set of stress-test graph sizes. Specifically, for each task and each target budget, we use binary search to identify the largest node count such that the final inline-coding prompt, including the task description, question, and serialized edge list, fits within a target fraction of the group's context window. We consider four target budgets: \emph{full} ($0.98\times$), \emph{half} ($0.5\times$), \emph{quarter} ($0.25\times$), and \emph{tenth} ($0.1\times$). The \emph{full} setting uses a $2\%$ safety margin and represents the highest feasible stress level, while \emph{tenth} represents the weakest stress level. 
Since a model's context budget is shared between input and output tokens, the \emph{full} setting leaves little room for generation and may result in incomplete responses. For the remaining settings, we set the generation budget to $\min(\text{context length}-\text{input length}, \text{output length cap})$, where the output length cap is the maximum output length specified by the API provider, to reserve sufficient space for response generation.
These relative graph sizes are measured under the inline setting, since file-based graph loading does not substantially affect the task-description length.

All six models are evaluated on the same subset of $12$ composite tasks, with $3$ tasks sampled for each \emph{combo size}~$\in\{1,2,3,4\}$, enabling paired comparisons across models (Table~\ref{tab:stress-sizes}). We use the graph-generation code in our pipeline to instantiate graphs under each target budget and use the corresponding labeling code to obtain ground-truth answers.

Each (model, fraction) cell in Table~\ref{tab:stress-em-pc} aggregates over $12$ tasks $\times$ $2$ task descriptions (explicit and implicit), resulting in $24$ instances. We evaluate three reasoning-loading modes: \emph{coding-file}, where the prompt provides the path to a stress-test edge-list file and the generated program reads the file, corresponding to low input and low output; \emph{coding-inline}, where the edge list is directly placed in the prompt and the generated program must reconstruct the graph from text, corresponding to high input and high output; and \emph{textual}, where the model directly solves the same inlined graph in natural language without explicitly reconstructing the whole graph, corresponding to high input and medium-length output. All results are reported under the strict evaluation. We also include the graph-size-$10$ setting as a non-stress reference for each model's baseline capability.

\textbf{Finding 1: In the low-input, low-output coding-file mode, models retain most of their non-stress capability.}
Table~\ref{tab:stress-em-pc}, Panel A reports the stress-test results for coding-file reasoning. On average, the evaluated models achieve $64.6$ EM and $66.4$ PC under the \emph{full} stress setting, compared with $73.2$ EM and $75.4$ PC in the non-stress $n{=}10$ setting. This corresponds to retention rates of $88.3\%$ and $88.1\%$, respectively. Such high retention is expected: although larger graphs increase computational workload, they do not substantially increase the LLM's own prompt-processing burden, since the graph is stored externally and graph operations are delegated to the generated program and execution environment. The remaining performance loss is mainly caused by solver timeouts on large graphs. We use a default timeout of $60$ seconds; a longer execution budget would likely further improve retention. Thus, coding-file stress testing primarily stresses computational resources rather than the LLM's context-processing ability. 

\textbf{Finding 2: In the long-input, long-output coding-inline mode, most models collapse under stress.}
Table~\ref{tab:stress-em-pc}, Panel B reports the stress-test results for coding-inline reasoning. We characterize this mode as high-input and high-output because the model would process a long serialized graph in the prompt and generate code that reconstructs the graph, even when the downstream algorithm does not require explicitly using every edge. This creates a joint long-input and long-output bottleneck. On average, performance drops from $75.6$ EM and $77.4$ PC at $n{=}10$ to $52.8$ EM and $54.2$ PC under the \emph{tenth} setting, and further to $11.8$ EM and $11.8$ PC under the \emph{full} stress setting. The corresponding \emph{full}-stress retention rates are only $15.6\%$ and $15.2\%$, showing that coding-inline reasoning degrades substantially under stress.

We further observe that some models use the available generation budget more effectively than others. Certain models avoid explicit graph reconstruction and output only the \texttt{solve} function, thereby retaining some performance under \emph{full} stress; DeepSeek-R1 is the clearest example, achieving $50.0$ EM and $50.0$ PC. In contrast, other models repeatedly attempt to reconstruct the entire graph in the generated code, and their outputs are truncated under the limited generation budget, leading to invalid responses; Gemma-4-31B and Qwen2.5-7B are representative examples of this failure mode, both dropping to $0.0$ EM and $0.0$ PC under \emph{full} stress.

The degree of degradation varies substantially across models. DeepSeek-R1 is the most robust: under the \emph{full} setting, it drops from $79.2$ EM to $50.0$ EM and from $83.3$ PC to $50.0$ PC, retaining $63.1\%$ and $60.0\%$ of its non-stress capability. Nemotron-3-Super shows limited but nonzero robustness, decreasing from $87.0$ EM and $87.0$ PC to $12.5$ EM and $12.5$ PC. In contrast, DeepSeek-V3.2 and Qwen2.5-72B retain only $4.2$ EM and $4.2$ PC under the highest stress level, despite much stronger performance under weaker settings. Gemma-4-31B and Qwen2.5-7B collapse completely under \emph{full} stress. Interestingly, Qwen2.5-72B performs best under the \emph{tenth} setting, achieving $83.3$ EM and $83.3$ PC, but degrades sharply as input-output pressure increases. This degradation is mainly caused by output truncation and its reliance on explicit graph extraction. Overall, coding-inline stress testing reveals substantial model-level variation: DeepSeek-R1 is the most robust under extreme stress, Nemotron-3-Super retains limited capability, while DeepSeek-V3.2, Qwen2.5-72B, Gemma-4-31B, and Qwen2.5-7B are fragile under the highest input-output load.

\textbf{Finding 3: In the high-input, moderate-output textual mode, models also struggle to preserve reasoning capability.}
Table~\ref{tab:stress-em-pc}, Panel C reports the stress-test results for textual reasoning. We characterize this mode as high-input and moderate-output because the model must process the full serialized graph in the prompt and reason over it directly, but does not need to generate code that explicitly reconstructs the whole graph. Even so, this setting remains highly stressful. On average, performance drops from $47.2$ EM and $52.7$ PC at $n{=}10$ to $29.9$ EM and $32.9$ PC under the \emph{tenth} setting, and further to $8.3$ EM and $8.8$ PC under the \emph{full} stress setting. The corresponding \emph{full}-stress retention rates are $17.6\%$ and $16.7\%$, respectively.

Model-level trends again differ. DeepSeek-R1 drops from $62.5$ EM to $20.8$ EM under the \emph{full} setting, retaining $33.3\%$ of its non-stress EM, whereas DeepSeek-V3.2 retains $40.0\%$, decreasing from $62.5$ EM to $25.0$ EM and showing the strongest robustness under the \emph{full} textual stress setting. Gemma-4-31B achieves strong retention under the moderate \emph{quarter} setting, reaching $37.5$ EM and retaining $60.0\%$, and modest retention $6.7\%$ under full stress. In contrast, {Nemotron-3-Super}, {Qwen2.5-72B}, and Qwen2.5-7B collapse to $0.0$ EM and $0.0$ PC under the \emph{full} setting. These results show that long-input textual graph reasoning remains difficult even when the output does not need to reconstruct the graph, and that robustness depends strongly on each model's long-context processing ability and answer-format behavior.

\textit{Why Qwen2.5-72B often underperforms {Qwen2.5-7B}?}
We also observe that {Qwen2.5-72B} often underperforms {Qwen2.5-7B} in this stress-test setting. A plausible explanation comes from qualitative inspection of outputs: {Qwen2.5-72B} tends to generate longer, more conversational, and more hedged responses, often including multi-step exposition such as ``we will use BFS for traversal,'' whereas {Qwen2.5-7B} is more concise and reaches a structured final answer more directly. Since our strict EM metric is extracted from \texttt{<answer>...</answer>}, concise and format-stable outputs are advantageous. Therefore, the smaller model's better performance in this setting does not necessarily indicate stronger graph reasoning, but more likely reflects better alignment with the target output format under long-context stress.

\textit{Why do some models degrade sharply after a certain stress point?}
Several models show abrupt performance drops once the stress level exceeds a threshold. In the coding-inline setting, Gemma-4-31B drops from $25.0$ EM under \emph{tenth} stress to $0.0$ under \emph{quarter} stress, while DeepSeek-V3.2 drops from $33.3$ EM under \emph{quarter} stress to $8.3$ under \emph{half} stress. This is mainly caused by output truncation: these models tend to strictly follow the graph-extraction step and reproduce the serialized graph in generated code. As graph size increases, the output exceeds the token budget, producing incomplete code and invalid predictions.
The same models degrade less severely in textual mode because they do not need to reconstruct the graph explicitly, leaving more output budget for reasoning and final answers. 
Nemotron-3-Super shows the opposite pattern. In coding-inline mode, it retains nonzero performance under \emph{full} stress, decreasing from $87.0$ EM at $n{=}10$ to $12.5$ EM, suggesting that it can sometimes avoid explicit graph reconstruction. However, in textual mode, it collapses from $54.2$ EM to $0.0$, because it spends excessive tokens on reasoning traces instead of concise final answers.
Overall, robustness under extreme stress depends not only on long-context processing, but also on stable output-budget management. Many models still lack reliable strategies for avoiding unnecessary token consumption, shortening intermediate reasoning, and prioritizing valid final-answer generation.

\begin{takeawaybox}
\small 
When graph size is scaled relative to each model's context window for model-specific stress testing, robustness is mainly determined by the input--output structure of the reasoning mode. \emph{Coding-file reasoning} keeps both input and output short and therefore retains high non-stress EM/PC even under \emph{full stress}, with most remaining failures caused by solver timeouts. In contrast, \emph{coding-inline reasoning} requires long-input processing and long-output graph reconstruction, leading to the weakest retention. Textual reasoning degrades sharply because models must reason directly over long serialized graphs. Different models adopt different output-token strategies under stress: some models use the generation budget more effectively, but this is unstable, suggesting the need for reliable token-allocation strategies under long-context stress.
\end{takeawaybox}

\subsection{Graph Reasoning across Combo Sizes}
\label{sec:combo-sizes}

\begin{table*}[t]
\centering
\caption{Exact Match (EM) and Partial Credit (PC) by combo size, with the across-combo Overall in the right pair of columns. Panel A reports coding-mode results. Panel B reports textual-mode results. Panel C reports the per-model gap, Coding $-$ Textual, for the $15$ paired models in percentage points. Bold entries in Panels A and B mark the column maximum for each metric.}
\label{tab:combo-mode}
\scriptsize
\setlength{\tabcolsep}{2pt}
\begin{tabularx}{\linewidth}{l YY YY YY YY YY}
\toprule
                         & \multicolumn{2}{c}{1} & \multicolumn{2}{c}{2} & \multicolumn{2}{c}{3} & \multicolumn{2}{c}{4} & \multicolumn{2}{c}{Overall} \\
\cmidrule(lr){2-3}\cmidrule(lr){4-5}\cmidrule(lr){6-7}\cmidrule(lr){8-9}\cmidrule(lr){10-11}
Model                    & EM & PC & EM & PC & EM & PC & EM & PC & EM & PC \\
\midrule
\multicolumn{11}{l}{\textit{Panel A: Coding-mode performance (averaged over file and inline graph loading)}} \\
\midrule
DeepSeek-V3.2            & $68.7$ & $69.2$ & $63.8$ & $64.8$ & $56.5$ & $57.5$ & $49.5$ & $49.2$ & $60.8$ & $61.3$ \\
DeepSeek-R1              & $68.2$ & $68.2$ & $62.7$ & $63.4$ & $62.0$ & $62.6$ & $50.2$ & $48.5$ & $61.5$ & $61.6$ \\
DeepSeek-R1-Distill-32B  & $58.7$ & $58.9$ & $51.9$ & $52.1$ & $44.9$ & $44.9$ & $36.9$ & $35.6$ & $50.3$ & $50.3$ \\
DeepSeek-R1-Distill-70B  & $57.3$ & $57.6$ & $50.9$ & $51.4$ & $48.0$ & $50.5$ & $37.8$ & $37.2$ & $49.8$ & $50.4$ \\
Gemma-4-31B              & $61.9$ & $61.9$ & $57.9$ & $58.7$ & $52.8$ & $54.5$ & $44.5$ & $42.8$ & $55.2$ & $55.5$ \\
Llama-3.1-8B             & $34.6$ & $36.9$ & $34.1$ & $34.6$ & $25.6$ & $28.7$ & $18.9$ & $20.9$ & $29.4$ & $31.4$ \\
Llama-3.3-70B            & $54.6$ & $54.8$ & $50.1$ & $51.2$ & $45.5$ & $47.9$ & $29.2$ & $30.1$ & $46.2$ & $47.3$ \\
Llama-4-Maverick         & $60.6$ & $61.1$ & $54.3$ & $55.0$ & $47.5$ & $48.5$ & $36.1$ & $36.5$ & $51.0$ & $51.6$ \\
Llama-4-Scout            & $50.7$ & $51.5$ & $50.0$ & $51.0$ & $41.2$ & $42.7$ & $26.0$ & $26.9$ & $43.8$ & $44.7$ \\
Nemotron-3-Super         & $69.1$ & $69.1$ & $61.9$ & $62.6$ & $59.5$ & $60.1$ & $49.1$ & $47.4$ & $61.0$ & $61.0$ \\
o4-mini                  & $\mathbf{72.4}$ & $\mathbf{72.4}$ & $\mathbf{68.8}$ & $\mathbf{69.3}$ & $\mathbf{64.8}$ & $\mathbf{65.9}$ & $\mathbf{52.9}$ & $\mathbf{51.1}$ & $\mathbf{65.7}$ & $\mathbf{65.7}$ \\
Qwen2.5-72B              & $58.1$ & $58.6$ & $55.3$ & $56.2$ & $46.8$ & $50.0$ & $34.5$ & $33.4$ & $50.1$ & $51.0$ \\
Qwen2.5-7B               & $37.8$ & $39.6$ & $30.9$ & $31.3$ & $25.0$ & $31.8$ & $19.8$ & $20.3$ & $29.7$ & $31.9$ \\
Qwen3-32B                & $57.3$ & $57.4$ & $54.1$ & $54.9$ & $50.6$ & $50.9$ & $45.5$ & $43.9$ & $52.9$ & $53.0$ \\
Qwen3-Coder-30B          & $62.4$ & $62.6$ & $57.6$ & $58.8$ & $51.7$ & $53.9$ & $42.5$ & $42.2$ & $54.7$ & $55.5$ \\
\cmidrule(l){2-11}
\textit{Average}         & $58.2$ & $58.7$ & $53.6$ & $54.4$ & $48.2$ & $50.0$ & $38.2$ & $37.7$ & $50.8$ & $51.5$ \\
\midrule
\multicolumn{11}{l}{\textit{Panel B: Textual-mode performance}} \\
\midrule
DeepSeek-V3.2            & $52.6$ & $54.0$ & $47.3$ & $49.1$ & $40.3$ & $47.3$ & $33.8$ & $34.2$ & $44.5$ & $47.0$ \\
DeepSeek-R1              & $42.4$ & $42.4$ & $38.3$ & $38.4$ & $31.6$ & $34.0$ & $23.8$ & $23.4$ & $35.1$ & $35.5$ \\
DeepSeek-R1-Distill-32B  & $30.6$ & $31.3$ & $28.0$ & $28.1$ & $20.3$ & $22.8$ & $17.5$ & $17.8$ & $25.2$ & $26.0$ \\
DeepSeek-R1-Distill-70B  & $35.1$ & $35.3$ & $34.4$ & $35.7$ & $29.1$ & $33.3$ & $21.0$ & $21.9$ & $30.8$ & $32.3$ \\
Gemma-4-31B              & $54.4$ & $55.5$ & $49.6$ & $51.5$ & $43.8$ & $49.5$ & $\mathbf{38.1}$ & $\mathbf{38.8}$ & $47.4$ & $49.6$ \\
Llama-3.1-8B             & $6.7$ & $7.2$ & $6.3$ & $7.4$ & $6.2$ & $8.8$ & $5.2$ & $6.5$ & $6.2$ & $7.4$ \\
Llama-3.3-70B            & $16.7$ & $19.9$ & $15.8$ & $18.1$ & $15.0$ & $18.5$ & $11.3$ & $11.7$ & $15.0$ & $17.4$ \\
Llama-4-Maverick         & $33.0$ & $35.3$ & $32.3$ & $34.8$ & $30.3$ & $33.8$ & $20.4$ & $21.8$ & $29.7$ & $32.1$ \\
Llama-4-Scout            & $27.1$ & $30.3$ & $25.6$ & $28.4$ & $20.0$ & $26.0$ & $13.1$ & $15.1$ & $22.3$ & $25.7$ \\
Nemotron-3-Super         & $35.3$ & $35.9$ & $32.5$ & $32.4$ & $26.3$ & $27.7$ & $18.9$ & $18.5$ & $29.2$ & $29.6$ \\
o4-mini                  & $\mathbf{57.8}$ & $\mathbf{58.5}$ & $\mathbf{49.8}$ & $\mathbf{52.3}$ & $\mathbf{46.9}$ & $\mathbf{51.7}$ & $37.8$ & $37.8$ & $\mathbf{48.9}$ & $\mathbf{50.9}$ \\
Qwen2.5-72B              & $13.5$ & $15.0$ & $12.7$ & $13.7$ & $12.5$ & $15.2$ & $11.9$ & $12.4$ & $12.7$ & $14.1$ \\
Qwen2.5-7B               & $18.8$ & $20.3$ & $17.7$ & $19.7$ & $12.8$ & $19.9$ & $12.7$ & $17.5$ & $16.2$ & $19.5$ \\
Qwen3-32B                & $36.0$ & $37.0$ & $36.0$ & $36.7$ & $29.4$ & $34.3$ & $25.3$ & $25.3$ & $32.5$ & $34.0$ \\
Qwen3-Coder-30B          & $36.2$ & $39.0$ & $35.6$ & $38.4$ & $30.0$ & $38.3$ & $20.4$ & $22.8$ & $31.5$ & $35.4$ \\
QwQ-32B                  & $39.2$ & $40.1$ & $37.0$ & $37.9$ & $34.1$ & $39.2$ & $25.4$ & $27.0$ & $34.9$ & $36.8$ \\
\cmidrule(l){2-11}
\textit{Average}         & $33.5$ & $34.8$ & $31.2$ & $32.7$ & $26.8$ & $31.3$ & $21.0$ & $22.0$ & $28.9$ & $30.8$ \\
\midrule
\multicolumn{11}{l}{\textit{Panel C: Gap (Coding $-$ Textual) for the 15 paired models}} \\
\midrule
DeepSeek-V3.2            & $+16.1$ & $+15.2$ & $+16.5$ & $+15.7$ & $+16.2$ & $+10.2$ & $+15.7$ & $+15.0$ & $+16.3$ & $+14.3$ \\
DeepSeek-R1              & $+25.8$ & $+25.8$ & $+24.4$ & $+25.0$ & $+30.4$ & $+28.6$ & $+26.4$ & $+25.1$ & $+26.4$ & $+26.1$ \\
DeepSeek-R1-Distill-32B  & $+28.1$ & $+27.6$ & $+23.9$ & $+24.0$ & $+24.6$ & $+22.1$ & $+19.4$ & $+17.8$ & $+25.1$ & $+24.3$ \\
DeepSeek-R1-Distill-70B  & $+22.2$ & $+22.3$ & $+16.5$ & $+15.7$ & $+18.9$ & $+17.2$ & $+16.8$ & $+15.3$ & $+19.0$ & $+18.1$ \\
Gemma-4-31B              & $+7.5$ & $+6.4$ & $+8.3$ & $+7.2$ & $+9.0$ & $+5.0$ & $+6.4$ & $+4.0$ & $+7.8$ & $+5.9$ \\
Llama-3.1-8B             & $+27.9$ & $+29.7$ & $+27.8$ & $+27.2$ & $+19.4$ & $+19.9$ & $+13.7$ & $+14.4$ & $+23.2$ & $+24.0$ \\
Llama-3.3-70B            & $+37.9$ & $+34.9$ & $+34.3$ & $+33.1$ & $+30.5$ & $+29.4$ & $+17.9$ & $+18.4$ & $+31.2$ & $+29.9$ \\
Llama-4-Maverick         & $+27.6$ & $+25.8$ & $+22.0$ & $+20.2$ & $+17.2$ & $+14.7$ & $+15.7$ & $+14.7$ & $+21.3$ & $+19.5$ \\
Llama-4-Scout            & $+23.6$ & $+21.2$ & $+24.4$ & $+22.6$ & $+21.2$ & $+16.7$ & $+12.9$ & $+11.8$ & $+21.5$ & $+19.0$ \\
Nemotron-3-Super         & $+33.8$ & $+33.2$ & $+29.4$ & $+30.2$ & $+33.2$ & $+32.4$ & $+30.2$ & $+28.9$ & $+31.8$ & $+31.4$ \\
o4-mini                  & $+14.6$ & $+13.9$ & $+19.0$ & $+17.0$ & $+17.9$ & $+14.2$ & $+15.1$ & $+13.3$ & $+16.8$ & $+14.8$ \\
Qwen2.5-72B              & $+44.6$ & $+43.6$ & $+42.6$ & $+42.5$ & $+34.3$ & $+34.8$ & $+22.6$ & $+21.0$ & $+37.4$ & $+36.9$ \\
Qwen2.5-7B               & $+19.0$ & $+19.3$ & $+13.2$ & $+11.6$ & $+12.2$ & $+11.9$ & $+7.1$ & $+2.8$ & $+13.5$ & $+12.4$ \\
Qwen3-32B                & $+21.3$ & $+20.4$ & $+18.1$ & $+18.2$ & $+21.2$ & $+16.6$ & $+20.2$ & $+18.6$ & $+20.4$ & $+19.0$ \\
Qwen3-Coder-30B          & $+26.2$ & $+23.6$ & $+22.0$ & $+20.4$ & $+21.7$ & $+15.6$ & $+22.1$ & $+19.4$ & $+23.2$ & $+20.1$ \\
\cmidrule(l){2-11}
\textit{Average}         & $+25.1$ & $+24.2$ & $+22.8$ & $+22.0$ & $+21.9$ & $+19.3$ & $+17.5$ & $+16.0$ & $+22.3$ & $+21.0$ \\
\bottomrule
\end{tabularx}
\end{table*}

Our benchmark introduces composition depth: each task can be composed of one or more seed tasks. This subsection studies how the combo size affects graph reasoning performance, intending to understand how LLMs behave when a task expands from a single-step problem to a multi-step composite problem.

\textbf{Experimental setup.}
We consider tasks composed of one to four seed tasks, and use the combo size to denote the compositional depth. This partitions all tasks into four buckets, $\{1,2,3,4\}$. When combo size=1, the task corresponds to a single seed task, such as \emph{cycle detection}; when combo size=4, the task combines four independent reasoning steps into one composite task. We evaluate each (model, mode) pair using Exact Match (EM) and Partial Credit (PC), where EM measures whether the prediction exactly matches the ground truth, and PC gives graded credit for near-correct outputs. We analyze the effect of combo size under the plain prompt (i.e., zero-shot CoT) setting for both code-based and text-based reasoning. Plain prompt means zero-shot chain-of-thought prompting.

\textbf{Finding 1: Model performance consistently decreases as combo size increases.}
Table~\ref{tab:combo-mode} reports the performance of multiple LLMs across combo sizes under both reasoning modes. On average, the strict EM of the $15$ models under coding reasoning drops from $58.2$ at combo size=1 to $38.2$ at combo size=4, a decrease of $20.0$ percentage points. For the $16$ models under textual reasoning, strict EM drops from $33.5$ to $21.0$, a decrease of $12.5$ percentage points. PC follows the same trend. These results indicate that compositional depth is a stable difficulty axis: as a task requires integrating more seed tasks, models must handle longer dependency chains and more complex intermediate states, leading to lower reasoning accuracy.

\textbf{Finding 2: Code-based reasoning outperforms text-based reasoning at every combo size.}
As shown in Table~\ref{tab:combo-mode}, code-based reasoning consistently outperforms text-based reasoning across all combo sizes. The average EM gap decreases from $+25.1$ percentage points at combo size=1 to $+17.5$ percentage points at combo size=4, but the advantage remains throughout. This suggests that translating the reasoning process into executable code and delegating execution to the interpreter is more reliable than requiring the model to chain multiple reasoning steps in natural language. As combo size increases, both modes degrade, which narrows the gap, but code-based reasoning maintains higher absolute performance.


\begin{takeawaybox}
As tasks expand from one seed task to four seed tasks, all evaluated models degrade under both reasoning modes, showing that combo size is a stable and effective complexity dimension. 
Code-based reasoning outperforms text-based reasoning at every combo size, but both modes degrade as compositional depth increases. 
\texttt{o4-mini} is the only model that maintains strict EM above $50$ at combo size=4 under code-based reasoning. Under text-based reasoning, \texttt{o4-mini} leads at combo size=1--3, while \texttt{Gemma-4-31B} slightly surpasses it at combo size=4.
\end{takeawaybox}

\subsection{Explicit vs. Implicit Task Descriptions}
\label{sec:appendix_implicit_description}

\begin{table*}[t]
\centering
\scriptsize
\setlength{\tabcolsep}{2pt}
\caption{Strict Exact Match (EM) and Partial Credit (PC) by description style. Values are reported as percentages without the \% symbol. Panel A reports coding-mode results, where Explicit and Implicit columns each average the file and inline graph-loading scenarios. Panel B reports textual-mode results. The Gap columns report Explicit $-$ Implicit in percentage points. Bold entries mark the column maximum for each metric.}
\label{tab:explicit-implicit}
\begin{tabularx}{\linewidth}{l YY YY YY @{\hspace{0.8em}} YY YY YY}
\toprule
& \multicolumn{6}{c}{\textit{Panel A: Coding-mode performance}}
& \multicolumn{6}{c}{\textit{Panel B: Textual-mode performance}} \\
\cmidrule(lr){2-7}\cmidrule(lr){8-13}
& \multicolumn{2}{c}{Explicit} & \multicolumn{2}{c}{Implicit} & \multicolumn{2}{c}{Gap}
& \multicolumn{2}{c}{Explicit} & \multicolumn{2}{c}{Implicit} & \multicolumn{2}{c}{Gap} \\
\cmidrule(lr){2-3}\cmidrule(lr){4-5}\cmidrule(lr){6-7}
\cmidrule(lr){8-9}\cmidrule(lr){10-11}\cmidrule(lr){12-13}
Model & EM & PC & EM & PC & EM & PC
& EM & PC & EM & PC & EM & PC \\
\midrule
DeepSeek-V3.2
& $62.3$ & $62.9$ & $59.2$ & $59.7$ & $+3.1$ & $+3.2$
& $45.9$ & $47.9$ & $43.2$ & $46.1$ & $+2.7$ & $+1.8$ \\

DeepSeek-R1
& $63.7$ & $63.7$ & $59.7$ & $59.7$ & $+4.0$ & $+4.0$
& $35.2$ & $35.7$ & $35.0$ & $35.4$ & $+0.2$ & $+0.3$ \\

DeepSeek-R1-Distill-32B
& $53.4$ & $53.1$ & $50.7$ & $50.9$ & $+2.7$ & $+2.2$
& $25.8$ & $26.6$ & $24.6$ & $25.3$ & $+1.2$ & $+1.3$ \\

DeepSeek-R1-Distill-70B
& $53.4$ & $53.8$ & $46.9$ & $47.7$ & $+6.5$ & $+6.1$
& $31.1$ & $32.4$ & $30.5$ & $32.2$ & $+0.6$ & $+0.2$ \\

Gemma-4-31B
& $55.8$ & $55.9$ & $54.5$ & $54.9$ & $+1.3$ & $+1.0$
& $47.8$ & $49.9$ & $47.0$ & $49.3$ & $+0.8$ & $+0.6$ \\

Llama-3.1-8B
& $29.8$ & $32.0$ & $28.9$ & $30.7$ & $+0.9$ & $+1.3$
& $8.2$ & $9.8$ & $4.1$ & $5.0$ & $+4.1$ & $+4.8$ \\

Llama-3.3-70B
& $47.4$ & $48.5$ & $45.0$ & $46.0$ & $+2.4$ & $+2.5$
& $15.4$ & $17.7$ & $14.6$ & $17.0$ & $+0.8$ & $+0.7$ \\

Llama-4-Maverick
& $52.4$ & $52.8$ & $49.6$ & $50.4$ & $+2.8$ & $+2.4$
& $29.6$ & $32.6$ & $29.7$ & $31.6$ & $-0.1$ & $+1.0$ \\

Llama-4-Scout
& $45.1$ & $46.1$ & $42.6$ & $43.6$ & $+2.5$ & $+2.5$
& $22.6$ & $26.2$ & $22.0$ & $25.1$ & $+0.6$ & $+1.1$ \\

Nemotron-3-Super
& $62.7$ & $62.7$ & $59.3$ & $59.4$ & $+3.4$ & $+3.3$
& $30.2$ & $30.4$ & $28.2$ & $28.7$ & $+2.0$ & $+1.7$ \\

o4-mini
& $\mathbf{67.3}$ & $\mathbf{67.5}$ & $\mathbf{64.0}$ & $\mathbf{63.9}$ & $+3.3$ & $+3.6$
& $\mathbf{50.4}$ & $\mathbf{51.9}$ & $\mathbf{47.5}$ & $\mathbf{49.9}$ & $+2.9$ & $+2.0$ \\

Qwen2.5-72B
& $50.6$ & $51.4$ & $49.4$ & $50.4$ & $+1.2$ & $+1.0$
& $13.9$ & $15.3$ & $11.5$ & $12.9$ & $+2.4$ & $+2.4$ \\

Qwen2.5-7B
& $30.2$ & $32.9$ & $29.1$ & $30.8$ & $+1.1$ & $+2.1$
& $17.8$ & $20.8$ & $14.6$ & $18.2$ & $+3.2$ & $+2.6$ \\

Qwen3-32B
& $55.0$ & $54.7$ & $53.2$ & $53.7$ & $+1.8$ & $+1.0$
& $32.6$ & $34.2$ & $32.4$ & $33.7$ & $+0.2$ & $+0.5$ \\

Qwen3-Coder-30B
& $57.8$ & $58.1$ & $51.6$ & $52.8$ & $\mathbf{+6.2}$ & $\mathbf{+5.3}$
& $32.7$ & $36.7$ & $30.4$ & $34.1$ & $+2.3$ & $+2.6$ \\
\midrule
\textit{Average}
& $52.5$ & $53.1$ & $49.6$ & $50.3$ & $+2.9$ & $+2.8$
& $29.7$ & $31.6$ & $28.1$ & $30.0$ & $+1.6$ & $+1.6$ \\
\bottomrule
\end{tabularx}
\end{table*}


Existing graph reasoning benchmarks often state the graph nature of the task explicitly and directly specify the target graph-algorithmic objective (e.g., ``Given a graph G, find the shortest path from source node s to target node t''). In realistic applications, however, users may describe the same underlying problem through domain-specific scenarios, such as social networks, course prerequisites, or transit routes, without using graph-theoretic terminology. The model must therefore infer both the latent graph structure and the corresponding algorithmic objective from natural language. To study this factor, we introduce description-level complexity by rendering each underlying graph problem in two forms: an explicit graph-theoretic description and an implicit scenario-based description. This subsection examines how this description style affects LLM graph reasoning.

\textbf{Experimental setup.}
Each task instance is paired with two natural-language descriptions. The \emph{explicit} description states the task using formal graph-theoretic terminology, while the \emph{implicit} description reframes the same task as a real-world domain scenario, such as a social network, course system, or transit network. Apart from the description text, the input graph, seed-task composition, and ground-truth evaluator are unchanged. We evaluate each (model, mode) pair using Exact Match (EM) and Partial Credit (PC), averaging over the remaining benchmark dimensions so that Explicit and Implicit results are directly comparable.

\textbf{Finding 1: Implicit descriptions introduce a mild but limited challenge in both code-based and text-based reasoning.}
Table~\ref{tab:explicit-implicit} summarizes performance by description style under both reasoning modes. The average Explicit$-$Implicit EM gap is $+2.9$ percentage points for code-based reasoning and $+1.6$ percentage points for text-based reasoning, with PC showing the same trend. This suggests that implicit scenario-based descriptions do increase the difficulty of mapping natural language to the underlying graph task, but the effect is modest compared with the degradation caused by graph size and compositional depth. A potential explanation is that current LLMs already possess strong natural-language understanding capabilities, so recognizing the graph formulation from an implicit description is not the dominant bottleneck in this benchmark.

\textbf{Finding 2: Code-based reasoning is more sensitive to task description style.}
Code-based reasoning is roughly twice as sensitive to description style as text-based reasoning (EM gap: $+2.9$ vs. $+1.6$). This pattern also holds at the model level: $12/15$ paired models show a larger explicit advantage under code-based reasoning than under text-based reasoning. The potential reason is that explicit prompts allow the model to translate formal graph-theoretic definitions more directly into standard algorithmic implementations. In contrast, implicit prompts require an additional semantic mapping step from a domain scenario to the correct graph primitive. Code-based reasoning is more brittle to errors in this step: if the model maps a phrase such as ``shortest delivery route'' to the wrong primitive, e.g., \texttt{nx.shortest\_path} instead of \texttt{nx.minimum\_spanning\_tree}, the generated program is likely to fail as a whole. Text-based chain-of-thought reasoning, by contrast, can adjust its interpretation more gradually during the reasoning process, making it relatively less sensitive to description style.

\begin{takeawaybox}
Formal explicit descriptions yield slightly higher performance than real-world implicit descriptions across both reasoning modes and almost all evaluated models. The effect is consistent but small: for current strong models, description style usually changes EM by only $1$--$7$ points, far less than graph size or compositional depth. Code-based reasoning is more affected by description style because an incorrect mapping from scenario language to graph primitives can directly lead to an incorrect program.
\end{takeawaybox}

\subsection{File-Based vs. Inline Graph Loading in Code-Based Reasoning}
\label{sec:appendix_file}

\begin{table*}[t]
\centering
\caption{Coding-mode strict Exact Match (EM) and Partial Credit (PC) under file-based and inline graph loading. \emph{Inline (\texttt{solve})} reports the correctness of the model's final answer by evaluating the ``solve'' function; \emph{Inline (fully)} additionally requires the model's program to have parsed the inlined graph correctly (a cell counts as correct only if both graph extraction and the final answer are correct). \emph{Graph Extraction (GE)} is the inline-only graph-extraction accuracy (the fraction of inline cells where the program's parsed graph exactly matches the ground truth; file mode is trivially correct and excluded). \emph{Gap} denotes File $-$ Inline (\texttt{solve}) in percentage points. The \emph{Average} row reports the per-column mean over the evaluated models. Bold entries mark the maximum for each metric.}
\label{tab:inline-file}
\footnotesize
\setlength{\tabcolsep}{2pt}
\begin{tabularx}{\linewidth}{l YY YY YY Y YY}
\toprule
                         & \multicolumn{2}{c}{File}
                         & \multicolumn{2}{c}{Inline (\texttt{solve})}
                         & \multicolumn{3}{c}{Inline (fully)}
                         & \multicolumn{2}{c}{Gap (File $-$ Inline)} \\
\cmidrule(lr){2-3}\cmidrule(lr){4-5}\cmidrule(lr){6-8}\cmidrule(lr){9-10}
Model                    & EM & PC & EM & PC & EM & PC & GE & EM & PC \\
\midrule
DeepSeek-V3.2            & $74.8$ & $75.5$ & $46.7$ & $47.1$ & $\mathbf{38.5}$ & $\mathbf{39.4}$ & $\mathbf{88.4}$ & $+28.1$ & $+28.4$ \\
DeepSeek-R1              & $75.4$ & $75.6$ & $48.0$ & $47.8$ & $23.4$ & $23.8$ & $61.5$ & $+27.4$ & $+27.8$ \\
DeepSeek-R1-Distill-32B  & $70.2$ & $70.1$ & $33.9$ & $34.0$ & $7.9$ & $8.0$ & $23.0$ & $+36.3$ & $+36.1$ \\
DeepSeek-R1-Distill-70B  & $63.2$ & $64.1$ & $37.1$ & $37.5$ & $17.5$ & $18.0$ & $37.0$ & $+26.1$ & $+26.6$ \\
Gemma-4-31B              & $74.3$ & $74.6$ & $36.0$ & $36.2$ & $37.8$ & $38.9$ & $77.0$ & $+38.3$ & $\mathbf{+38.4}$ \\
Llama-3.1-8B             & $35.9$ & $38.1$ & $22.9$ & $24.6$ & $17.4$ & $19.0$ & $55.1$ & $+13.0$ & $+13.5$ \\
Llama-3.3-70B            & $57.5$ & $58.8$ & $34.9$ & $35.7$ & $29.9$ & $31.3$ & $74.8$ & $+22.6$ & $+23.0$ \\
Llama-4-Maverick         & $58.5$ & $59.7$ & $43.4$ & $43.6$ & $30.2$ & $31.2$ & $62.6$ & $+15.0$ & $+16.1$ \\
Llama-4-Scout            & $50.4$ & $51.5$ & $37.3$ & $38.2$ & $23.0$ & $24.4$ & $47.6$ & $+13.1$ & $+13.4$ \\
Nemotron-3-Super         & $\mathbf{75.6}$ & $\mathbf{75.6}$ & $46.5$ & $46.5$ & $21.0$ & $21.6$ & $44.2$ & $+29.1$ & $+29.1$ \\
o4-mini                  & $74.0$ & $74.2$ & $\mathbf{57.3}$ & $\mathbf{57.2}$ & $32.0$ & $32.9$ & $53.1$ & $+16.7$ & $+17.1$ \\
Qwen2.5-72B              & $61.7$ & $63.1$ & $38.4$ & $38.7$ & $29.1$ & $30.4$ & $54.9$ & $+23.4$ & $+24.4$ \\
Qwen2.5-7B               & $32.8$ & $35.6$ & $26.6$ & $28.1$ & $13.8$ & $15.1$ & $14.7$ & $+6.2$ & $+7.5$ \\
Qwen3-32B                & $73.4$ & $73.4$ & $34.9$ & $35.1$ & $18.6$ & $18.9$ & $59.0$ & $\mathbf{+38.5}$ & $+38.3$ \\
Qwen3-Coder-30B          & $68.1$ & $69.0$ & $41.1$ & $41.9$ & $27.6$ & $28.5$ & $64.8$ & $+27.0$ & $+27.1$ \\
\cmidrule(l){2-10}
\textit{Average}         & $63.1$ & $63.9$ & $39.0$ & $39.5$ & $24.5$ & $25.4$ & $54.5$ & $+24.0$ & $+24.4$ \\
\bottomrule
\end{tabularx}
\end{table*}

Graph data can be supplied to LLMs in two ways: either embedded directly in the prompt (\emph{inline graph loading}) or provided through a file path that the generated program must read (\emph{file-based loading}). Existing LLM graph reasoning benchmarks often adopt the inline setting, where the full graph is serialized into the prompt. While this setting is natural for text-based reasoning, it becomes less suitable for large graphs under code-based reasoning: the model must process a large amount of structured input within its context window, which increases token cost and may amplify long-context degradation. In contrast, local-file graph loading better reflects practical code-generation scenarios, where users provide a task description, a file path, and the data format, and expect the model to generate executable code that loads the graph and solves the task. In such cases, placing the entire graph in the prompt introduces an avoidable parsing burden.
Motivated by this distinction, we compare inline graph loading with file-based graph loading under code-based reasoning. We do not evaluate file-based loading for text-based reasoning because text-based reasoning requires direct access to the graph structure in the prompt and cannot operate from a file path alone.

\textbf{Experimental setup.}
In the \emph{file} setting, the prompt provides the path to a graph file, and the generated program reads and parses the graph from this file. In the \emph{inline} setting, the graph data are serialized directly in the prompt, and the generated program must recover the graph from the textual representation. The two settings use the same underlying graphs and tasks, differing only in the graph-loading mechanism. By default, we extract the generated \texttt{solve} function and pass it to the evaluation scripts, which assess only the functional correctness of \texttt{solve}. We refer to this setting as \textit{inline (\texttt{solve})}. We also consider an alternative setting that executes the full generated code and evaluates both answer correctness and graph-extraction correctness, which is referred to as \textit{inline (full)}. Graph-extraction accuracy is defined as the proportion of cases in which the recovered graph is identical to the ground-truth graph. We report Exact Match (EM) and Partial Credit (PC), averaged over the remaining dimensions, to enable a direct comparison between the file and inline settings. Table~\ref{tab:inline-file} reports the per-model results and the corresponding graph-loading performance gaps.

\textbf{Finding 1: File-based loading is consistently easier than inline graph loading.}
All models under coding reasoning perform better with file-based loading than with inline graph loading. Averaged over $15$ models, strict EM increases from $39.0$ under inline graph loading to $63.1$ under file-based graph loading, yielding a $+24.0$ percentage-point gap, or about a $1.6\times$ relative improvement. PC shows the same pattern, with an average gap of $+24.4$ percentage points. The largest EM gaps occur for \texttt{qwen3-32b} ($+38.5$), \texttt{gemma-4-31b} ($+38.3$), \texttt{deepseek-r1-distill-32b} ($+36.3$), and \texttt{nemotron-3-super} ($+29.1$), while the smallest gaps occur for smaller or weaker models such as \texttt{qwen2.5-7b} ($+6.2$) and \texttt{llama-3.1-8b} ($+13.0$).

These results suggest that file-based loading removes a major input-parsing burden from code-based reasoning. Stronger models can better exploit this cleaner interface by focusing on task modeling and algorithm implementation. Weaker models, however, may remain limited by the algorithmic reasoning step itself, so reducing the input-parsing burden yields smaller absolute gains.

\textbf{Finding 2: Full inline execution is much harder than evaluating the extracted \texttt{solve} function, because inline graph loading introduces an additional graph-extraction bottleneck.}
Table~\ref{tab:inline-file} shows that the average graph-extraction accuracy (GE) is only $54.5\%$, ranging from $88.4\%$ for DeepSeek-V3.2 to $14.7\%$ for Qwen2.5-7B. This indicates that models often fail to faithfully reconstruct the original graph from the serialized prompt, even when they can generate a correct solution procedure. Consequently, the full inline setting is substantially more stringent than inline (\texttt{solve}): the average EM decreases from $39.0\%$ to $24.5\%$. This drop is consistent with the GE bottleneck: $\text{EM}_{\text{inline (full)}} \approx \text{EM}_{\text{inline}(\texttt{solve})} \times \text{GE}$, since a prediction is counted as correct only when both graph extraction and downstream reasoning are correct.

These results show that inline graph loading adds a nontrivial preprocessing step before graph reasoning begins. The model must parse long, highly structured node and edge lists, reconstruct the exact graph object, and then apply the target algorithm. Errors such as missing edges, duplicated edges, incorrect node identifiers, or faulty parsing logic can invalidate the final answer even when the algorithmic reasoning is correct. Thus, inline graph loading evaluates not only graph reasoning, but also the model's ability to convert serialized graph descriptions into executable graph representations, which helps explain the lower reliability of full inline execution.

\begin{takeawaybox}
Across all evaluated models under coding reasoning, switching from inline graph loading to file-based graph loading improves strict EM by an average of $+24.0$ percentage points, and every model benefits. This shows that the advantage of code-based graph reasoning is closely tied to a realistic file-based input interface. Inline graph loading, by contrast, introduces an additional parsing burden and can substantially weaken the benefit of executable code generation.
\end{takeawaybox}

\subsection{OA-Derived vs. Classical Graph Problems.}
\label{sec:appendix_leetcode}

Classical graph algorithms are common in textbooks and tutorials, so strong performance on them may partly reflect prior exposure to LLM pre-training. OA graph problems, by contrast, often combine graph primitives with open-ended descriptions and task-specific constraints. We therefore compare the two sources to assess whether OA tasks provide a more challenging testbed.

\textbf{Experimental setup.}
The seed tasks in {\dataset} come from two sources: LeetCode, used as a representative OA source, such as \emph{Cheapest Flights Within K Stops}, and the classical graph-algorithm canon, such as \emph{shortest path}; more examples are in Figure~\ref{fig:seed_tasks}. For brevity, we refer to them as OA and classical (CL) tasks, respectively. We restrict this analysis to the combo size=1 slice since task-source attribution is unambiguous only for single-seed tasks. To reduce confounding from intrinsic difficulty, we balance major problem-complexity types across the two sources where possible. We report Exact Match (EM) and Partial Credit (PC) under both code-based and text-based reasoning, and further stratify results by graph size. We define the source gap as CL minus OA, so positive values indicate that OA tasks are harder.

\begin{table*}[t]
\centering
\caption{Per-model strict EM and PC on OA vs. CL tasks. 
The \emph{Average} row reports the per-column mean over the models.
}
\label{tab:leetcode-overall}
\scriptsize
\setlength{\tabcolsep}{2pt}
\begin{tabularx}{\linewidth}{l YY YY YY YY YY YY}
\toprule
& \multicolumn{6}{c}{Coding mode} & \multicolumn{6}{c}{Textual mode} \\
\cmidrule(lr){2-7}\cmidrule(lr){8-13}
& \multicolumn{2}{c}{OA} & \multicolumn{2}{c}{CL} & \multicolumn{2}{c}{CL $-$ OA} 
& \multicolumn{2}{c}{OA} & \multicolumn{2}{c}{CL} & \multicolumn{2}{c}{CL $-$ OA} \\
\cmidrule(lr){2-3}\cmidrule(lr){4-5}\cmidrule(lr){6-7}
\cmidrule(lr){8-9}\cmidrule(lr){10-11}\cmidrule(lr){12-13}
Model & EM & PC & EM & PC & EM & PC & EM & PC & EM & PC & EM & PC \\
\midrule
DeepSeek-V3.2
& $59.4$ & $59.5$ & $74.1$ & $74.8$ & $+14.7$ & $+15.3$
& $47.5$ & $48.0$ & $55.6$ & $57.6$ & $+8.1$ & $+9.6$ \\
DeepSeek-R1
& $60.4$ & $60.4$ & $71.9$ & $72.0$ & $+11.5$ & $+11.6$
& $28.7$ & $28.7$ & $50.4$ & $50.4$ & $+21.7$ & $+21.7$ \\
DeepSeek-R1-Distill-32B
& $48.0$ & $48.1$ & $63.9$ & $64.1$ & $+15.9$ & $+16.0$
& $20.0$ & $20.0$ & $32.6$ & $32.7$ & $+12.6$ & $+12.7$ \\
DeepSeek-R1-Distill-70B
& $49.3$ & $49.7$ & $61.5$ & $61.7$ & $+12.2$ & $+12.0$
& $24.4$ & $24.4$ & $41.3$ & $41.7$ & $+16.9$ & $+17.3$ \\
Gemma-4-31B
& $49.7$ & $49.7$ & $68.9$ & $68.9$ & $+19.2$ & $+19.2$
& $38.8$ & $40.7$ & $63.4$ & $64.1$ & $+24.6$ & $+23.4$ \\
Llama-3.1-8B
& $29.1$ & $29.4$ & $36.9$ & $37.5$ & $+7.8$ & $+8.1$
& $3.1$ & $4.2$ & $8.7$ & $8.9$ & $+5.6$ & $+4.7$ \\
Llama-3.3-70B
& $47.2$ & $47.6$ & $58.8$ & $59.0$ & $+11.6$ & $+11.4$
& $10.0$ & $12.4$ & $20.7$ & $24.2$ & $+10.7$ & $+11.8$ \\
Llama-4-Maverick
& $50.3$ & $50.4$ & $66.5$ & $67.2$ & $+16.2$ & $+16.8$
& $21.9$ & $23.6$ & $39.5$ & $42.2$ & $+17.6$ & $+18.6$ \\
Llama-4-Scout
& $33.8$ & $35.0$ & $60.5$ & $60.9$ & $+26.7$ & $+25.9$
& $11.9$ & $15.1$ & $35.9$ & $39.1$ & $+24.0$ & $+24.0$ \\
Nemotron-3-Super
& $57.7$ & $57.7$ & $75.0$ & $75.0$ & $+17.3$ & $+17.3$
& $23.1$ & $23.3$ & $42.4$ & $43.2$ & $+19.3$ & $+19.9$ \\
o4-mini
& $58.1$ & $58.1$ & $80.6$ & $80.6$ & $+22.5$ & $+22.5$
& $46.9$ & $46.9$ & $64.1$ & $65.3$ & $+17.2$ & $+18.4$ \\
Qwen2.5-72B
& $47.8$ & $48.3$ & $64.0$ & $64.6$ & $+16.2$ & $+16.3$
& $3.7$ & $4.3$ & $19.2$ & $21.3$ & $+15.5$ & $+17.0$ \\
Qwen2.5-7B
& $21.9$ & $22.4$ & $36.2$ & $36.5$ & $+14.3$ & $+14.1$
& $6.2$ & $7.8$ & $26.1$ & $27.6$ & $+19.9$ & $+19.8$ \\
Qwen3-32B
& $46.4$ & $46.4$ & $63.0$ & $63.1$ & $+16.6$ & $+16.7$
& $21.9$ & $22.9$ & $44.2$ & $45.2$ & $+22.3$ & $+22.3$ \\
Qwen3-Coder-30B
& $51.3$ & $51.4$ & $68.5$ & $68.9$ & $+17.2$ & $+17.5$
& $21.9$ & $24.1$ & $43.5$ & $46.7$ & $+21.6$ & $+22.6$ \\
\cmidrule(l){2-13}
\textit{Average}
& $47.4$ & $47.6$ & $63.4$ & $63.7$ & $+16.0$ & $+16.0$
& $22.0$ & $23.1$ & $39.2$ & $40.7$ & $+17.2$ & $+17.6$ \\
\bottomrule
\end{tabularx}
\end{table*}

\textbf{Finding 1: OA tasks exhibit consistently lower performance than classical graph tasks.}
Table~\ref{tab:leetcode-overall} reports the overall results and source gaps. Under code-based reasoning, the average EM is $47.4$ on OA tasks and $63.4$ on classical tasks, yielding a $16.0$ percentage-point EM gap; the PC gap is also $16.0$ percentage points. Under text-based reasoning, the corresponding average EM values are $22.0$ and $39.2$, yielding an EM gap of $17.2$ percentage points and a PC gap of $17.6$ percentage points. For every model and reasoning mode, performance on classical tasks is higher than that on OA tasks, indicating that OA tasks are consistently more difficult in our benchmark.



\begin{table*}[t]
\centering
\caption{Per-size OA and CL performance for the two best-performing models on classical at graph size ${=}10$ in each reasoning mode. 
}
\label{tab:leetcode-top2-by-size}
\footnotesize
\setlength{\tabcolsep}{3pt}
\begin{tabularx}{\linewidth}{l l l YY YY YY YY}
\toprule
& & & \multicolumn{2}{c}{$n{=}10$} 
& \multicolumn{2}{c}{$n{=}100$} 
& \multicolumn{2}{c}{$n{=}1{,}000$} 
& \multicolumn{2}{c}{$n{=}10{,}000$} \\
\cmidrule(lr){4-5}\cmidrule(lr){6-7}\cmidrule(lr){8-9}\cmidrule(lr){10-11}
Mode & Model & Source & EM & PC & EM & PC & EM & PC & EM & PC \\
\midrule
\multirow{4}{*}{Coding}
        & \multirow{2}{*}{DeepSeek-V3.2}  & OA  
        & $85.0$ & $85.2$ & $73.8$ & $73.9$ & $51.2$ & $51.6$ & $31.2$ & $31.4$ \\
        &                                 & CL 
        & $\mathbf{97.8}$ & $\mathbf{98.3}$ & $88.5$ & $89.1$ & $64.9$ & $65.6$ & $47.3$ & $48.6$ \\
        & \multirow{2}{*}{o4-mini}        & OA  
        & $80.0$ & $80.0$ & $65.0$ & $65.0$ & $50.0$ & $50.0$ & $37.5$ & $37.5$ \\
        &                                 & CL 
        & $\mathbf{97.8}$ & $97.8$ & $93.5$ & $93.5$ & $75.5$ & $75.5$ & $55.4$ & $55.4$ \\
\midrule
\multirow{4}{*}{Textual}
        & \multirow{2}{*}{DeepSeek-R1}    & OA  
        & $72.5$ & $72.5$ & $27.5$ & $27.5$ & $10.0$ & $10.0$ & $5.0$ & $5.0$ \\
        &                                 & CL 
        & $\mathbf{98.6}$ & $\mathbf{98.6}$ & $49.3$ & $49.3$ & $36.2$ & $36.2$ & $17.4$ & $17.4$ \\
        & \multirow{2}{*}{Gemma-4-31B}    & OA  
        & $82.5$ & $86.0$ & $52.5$ & $52.5$ & $15.0$ & $17.9$ & $5.0$ & $6.6$ \\
        &                                 & CL 
        & $\mathbf{98.6}$ & $\mathbf{98.6}$ & $72.5$ & $75.3$ & $49.3$ & $49.3$ & $33.3$ & $33.3$ \\
\bottomrule
\end{tabularx}
\end{table*}

\textbf{Finding 2: Strong models approach saturation on classical graph tasks with small graphs, but not on OA tasks.}
Table~\ref{tab:leetcode-top2-by-size} reports the two best-performing models in each reasoning mode on classical tasks at graph size $10$. The four representative models all achieve at least $97.8$ EM on classical tasks, with two models under textual reasoning reaching $98.6$, indicating that performance approaches saturation on classical graph tasks with small graphs. However, on OA tasks at the same graph size, performance remains substantially lower: DeepSeek-V3.2 reaches $85.0$, o4-mini reaches $80.0$, Gemma-4-31B reaches $82.5$, and DeepSeek-R1 reaches $72.5$. The corresponding within-model gaps are $12.8$, $17.8$, $16.1$, and $26.1$ percentage points, respectively. These results show that current strong models are close to saturating small-scale classical graph tasks, but still make substantial errors on OA graph problems of the same size. Therefore, OA tasks are not trivial even in small-graph settings. This gap at the easiest size suggests that benchmarks relying only on classical graph problems may be insufficiently challenging for evaluating LLMs' graph reasoning. Including OA tasks helps avoid saturation on standard graph algorithms and provides a stronger test of problem understanding and graph-task modeling.

\textbf{Finding 3: OA-sourced problems with ultra-large graphs are highly challenging for LLM text-based reasoning.}
Table~\ref{tab:leetcode-top2-by-size} shows that text-based reasoning degrades sharply as graph size increases, especially on OA tasks. The two selected models under textual reasoning achieve $72.5$ and $82.5$ EM at graph size $10$, but both drop to only $5.0$ EM when the graph size increases to $10{,}000$. In contrast, the two models under coding reasoning still retain $31.2$ and $37.5$ EM on OA tasks at the same graph size. This suggests that code-based reasoning is more robust to graph-scale increases, while pure natural-language reasoning struggles with large serialized graph inputs in OA-sourced graph tasks. Since OA tasks are already harder at small graph sizes, combining OA problem structures with ultra-large graph inputs creates an even more challenging evaluation setting.

\begin{takeawaybox}
Every evaluated model performs worse on OA tasks than on classical graph tasks. Even at graph size $10$, where strong models approach saturation on classical tasks, performance remains substantially lower on OA tasks. This suggests that benchmarks relying only on classical graph problems are insufficient for fully evaluating LLM graph reasoning. OA graph problems therefore provide a less saturated and more challenging evaluation setting that better tests problem understanding and graph-task modeling.
\end{takeawaybox}

\subsection{Evaluating Domain-Knowledge Retrieval Augmentation on {\dataset}}
\label{sec:appendix_rq_rag}

\begin{figure}[!htbp]
    \centering
    \includegraphics[width=0.9\linewidth]{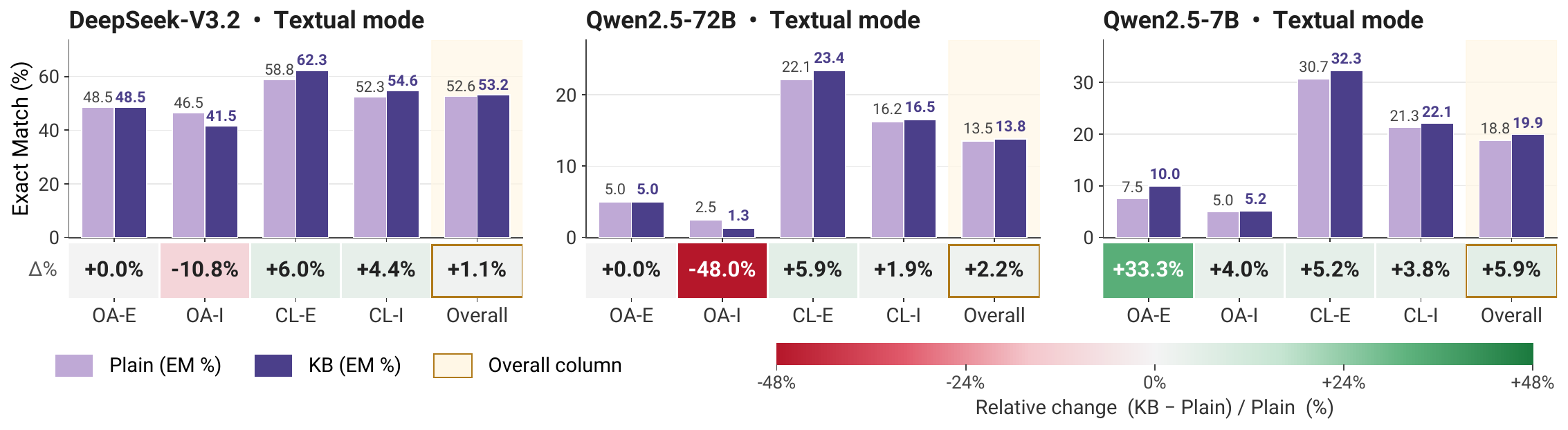}
    \caption{
\textbf{KB vs. Plain: Exact Match (EM) under text mode across task types.} Plain means no retrieval and uses zero-shot CoT prompting.
Results at combo size$=1$ by reasoning mode, model, and task source - task description (E = Explicit, I = Implicit). 
Grouped bars report Plain vs. KB EM, and the bottom strip reports the relative change (red = KB hurts, green = KB helps). The ``Overall'' column aggregates across task types.
}
    \label{fig:rag-leetcode-merged2}
\end{figure}

\begin{table*}[t]
\centering
\caption{Overall strict Exact Match (\%) results for retrieval-augmented reasoning. For coding mode, the knowledge base (KB) is \texttt{kb-graphteam}, retrieved with a dense retriever; results are reported under \texttt{file} and \texttt{inline} graph loading. For textual mode, the KB is the top-3 retrieved algorithmic knowledge entries. The \textit{rel. change} rows report $(\text{KB}-\text{Plain})/\text{Plain}$ in percentage.}
\label{tab:rag-overall}
\scriptsize
\setlength{\tabcolsep}{3pt}
\begin{tabularx}{0.86\linewidth}{l YY YY Y YYY}
\toprule
& \multicolumn{5}{c}{Coding} 
& \multicolumn{3}{c}{Textual} \\
\cmidrule(lr){2-6}\cmidrule(lr){7-9}
Model / Setting 
& \multicolumn{2}{c}{\texttt{file}} 
& \multicolumn{2}{c}{\texttt{inline}} 
& Overall
& Explicit & Implicit & Overall \\
\cmidrule(lr){2-3}\cmidrule(lr){4-5}
& Explicit & Implicit & Explicit & Implicit & & & & \\
\midrule
DeepSeek-V3.2 / Plain 
& $74.8$ & $70.2$ 
& $50.0$ & $50.9$ 
& $63.4$ 
& $45.9$ & $43.2$ & $44.5$ \\
DeepSeek-V3.2 / KB   
& $77.7$ & $60.8$ 
& $44.1$ & $46.9$ 
& $59.5$ 
& $46.1$ & $44.1$ & $45.1$ \\
\textit{rel. change}        
& $+3.9\%$ & $-13.4\%$ 
& $-11.8\%$ & $-7.9\%$ 
& $-6.2\%$ 
& $+0.4\%$ & $+2.1\%$ & $+1.3\%$ \\
\midrule
Qwen2.5-72B / Plain 
& $62.4$ & $61.1$ 
& $37.8$ & $38.9$ 
& $50.1$ 
& $13.9$ & $11.5$ & $12.7$ \\
Qwen2.5-72B / KB   
& $63.5$ & $59.7$ 
& $35.0$ & $32.8$ 
& $46.2$ 
& $17.4$ & $13.4$ & $15.4$ \\
\textit{rel. change}       
& $+1.8\%$ & $-2.3\%$ 
& $-7.4\%$ & $-15.7\%$ 
& $-7.8\%$ 
& $+25.2\%$ & $+16.5\%$ & $+21.3\%$ \\
\midrule
Qwen2.5-7B / Plain 
& $31.8$ & $29.8$ 
& $26.6$ & $26.5$ 
& $29.7$ 
& $17.8$ & $14.6$ & $16.2$ \\
Qwen2.5-7B / KB   
& $33.2$ & $32.3$ 
& $25.8$ & $22.5$ 
& $28.3$ 
& $19.7$ & $19.4$ & $19.6$ \\
\textit{rel. change}       
& $+4.4\%$ & $+8.4\%$ 
& $-3.0\%$ & $-15.1\%$ 
& $-4.7\%$ 
& $+10.7\%$ & $+32.9\%$ & $+21.0\%$ \\
\bottomrule
\end{tabularx}
\end{table*}

\begin{table*}[t]
\centering
\caption{Coding-mode strict Exact Match (EM) and Partial Credit (PC) at combo size=1, grouped by task source and description style. OA and CL denote LeetCode and classical tasks, respectively. The \textit{rel. change} rows report $(\text{KB}-\text{Plain})/\text{Plain}$ in percentage.}
\label{tab:rag-coding-leetcode}
\scriptsize
\setlength{\tabcolsep}{2pt}
\begin{tabularx}{\linewidth}{l YY YY YY YY YY}
\toprule
 & \multicolumn{2}{c}{OA Explicit} 
 & \multicolumn{2}{c}{OA Implicit} 
 & \multicolumn{2}{c}{CL Explicit} 
 & \multicolumn{2}{c}{CL Implicit} 
 & \multicolumn{2}{c}{Overall} \\
\cmidrule(lr){2-3}\cmidrule(lr){4-5}\cmidrule(lr){6-7}\cmidrule(lr){8-9}\cmidrule(lr){10-11}
Model / Setting & EM & PC & EM & PC & EM & PC & EM & PC & EM & PC \\
\midrule
DeepSeek-V3.2 / Plain
& $60.5$ & $60.8$ & $58.9$ & $59.0$ & $75.3$ & $76.1$ & $72.4$ & $73.1$ & $68.5$ & $69.0$ \\
DeepSeek-V3.2 / KB
& $57.2$ & $57.3$ & $56.4$ & $56.8$ & $79.5$ & $80.9$ & $70.2$ & $71.8$ & $67.4$ & $68.4$ \\
\textit{rel. change}
& $-5.5\%$ & $-5.8\%$ & $-4.2\%$ & $-3.7\%$ & $+5.6\%$ & $+6.3\%$ & $-3.0\%$ & $-1.8\%$ & $-1.6\%$ & $-0.9\%$ \\
\midrule
Qwen2.5-72B / Plain
& $48.0$ & $48.3$ & $47.6$ & $48.2$ & $69.1$ & $69.1$ & $59.0$ & $60.1$ & $58.1$ & $58.6$ \\
Qwen2.5-72B / KB
& $42.4$ & $43.2$ & $41.4$ & $42.2$ & $71.3$ & $71.7$ & $56.4$ & $57.0$ & $55.6$ & $56.2$ \\
\textit{rel. change}
& $-11.7\%$ & $-10.6\%$ & $-13.0\%$ & $-12.4\%$ & $+3.2\%$ & $+3.8\%$ & $-4.4\%$ & $-5.2\%$ & $-4.3\%$ & $-4.1\%$ \\
\midrule
Qwen2.5-7B / Plain
& $21.8$ & $22.8$ & $22.0$ & $22.0$ & $39.4$ & $39.6$ & $32.9$ & $33.3$ & $30.9$ & $31.3$ \\
Qwen2.5-7B / KB
& $26.0$ & $26.2$ & $17.4$ & $17.4$ & $45.0$ & $45.7$ & $27.9$ & $28.9$ & $31.3$ & $31.8$ \\
\textit{rel. change}
& $+19.3\%$ & $+14.9\%$ & $-20.9\%$ & $-20.9\%$ & $+14.2\%$ & $+15.4\%$ & $-15.2\%$ & $-13.2\%$ & $+1.3\%$ & $+1.6\%$ \\
\bottomrule
\end{tabularx}
\end{table*}

\begin{table*}[t]
\centering
\caption{Textual-mode strict Exact Match (EM) and Partial Credit (PC) at combo size=1, grouped by task source and description style. OA and CL denote LeetCode and classical tasks, respectively. The \textit{rel. change} rows report $(\text{KB}-\text{Plain})/\text{Plain}$ in percentage.}
\label{tab:rag-textual-leetcode}
\footnotesize
\setlength{\tabcolsep}{2pt}
\begin{tabularx}{\linewidth}{l YY YY YY YY YY}
\toprule
 & \multicolumn{2}{c}{OA Explicit} 
 & \multicolumn{2}{c}{OA Implicit} 
 & \multicolumn{2}{c}{CL Explicit} 
 & \multicolumn{2}{c}{CL Implicit} 
 & \multicolumn{2}{c}{Overall} \\
\cmidrule(lr){2-3}\cmidrule(lr){4-5}\cmidrule(lr){6-7}\cmidrule(lr){8-9}\cmidrule(lr){10-11}
Model / Setting & EM & PC & EM & PC & EM & PC & EM & PC & EM & PC \\
\midrule
DeepSeek-V3.2 / Plain    
& $48.5$ & $49.0$ & $46.5$ & $47.0$ & $58.8$ & $60.7$ & $52.3$ & $54.3$ & $52.6$ & $54.0$ \\
DeepSeek-V3.2 / KB 
& $48.5$ & $49.5$ & $41.5$ & $43.3$ & $62.3$ & $63.5$ & $54.6$ & $56.3$ & $53.2$ & $54.6$ \\
\textit{rel. change}
& $+0.0\%$ & $+1.0\%$ & $-10.8\%$ & $-7.9\%$ & $+6.0\%$ & $+4.6\%$ & $+4.4\%$ & $+3.7\%$ & $+1.1\%$ & $+1.1\%$ \\
\midrule
Qwen2.5-72B / Plain      
& $5.0$ & $5.5$ & $2.5$ & $3.0$ & $22.1$ & $24.1$ & $16.2$ & $18.3$ & $13.5$ & $15.0$ \\
Qwen2.5-72B / KB   
& $5.0$ & $6.4$ & $1.3$ & $1.3$ & $23.4$ & $25.4$ & $16.5$ & $18.6$ & $13.8$ & $15.3$ \\
\textit{rel. change}
& $+0.0\%$ & $+16.4\%$ & $-48.0\%$ & $-56.7\%$ & $+5.9\%$ & $+5.4\%$ & $+1.9\%$ & $+1.6\%$ & $+2.2\%$ & $+2.0\%$ \\
\midrule
Qwen2.5-7B / Plain       
& $7.5$ & $9.1$ & $5.0$ & $6.6$ & $30.7$ & $32.4$ & $21.3$ & $22.7$ & $18.8$ & $20.4$ \\
Qwen2.5-7B / KB    
& $10.0$ & $11.0$ & $5.2$ & $6.6$ & $32.3$ & $33.5$ & $22.1$ & $24.3$ & $19.9$ & $21.3$ \\
\textit{rel. change}
& $+33.3\%$ & $+20.9\%$ & $+4.0\%$ & $+0.0\%$ & $+5.2\%$ & $+3.4\%$ & $+3.8\%$ & $+7.0\%$ & $+5.9\%$ & $+4.4\%$ \\
\bottomrule
\end{tabularx}
\end{table*}

This subsection evaluates the effect of preliminary domain-knowledge retrieval augmentation (RAG) on {\dataset}. Our goal is to assess whether existing retrieval-augmented methods can consistently deliver benefits in more complex graph-reasoning settings.

\textbf{Experimental setup.}
We pair each retrieval-augmented run with its corresponding Plain baseline under the same base model, in order to isolate the effect of external knowledge. In the retrieval-augmented setting, the model first queries an external knowledge base and then incorporates the top-ranked retrieved context into the prompt. Retrieval is performed by a dense retriever using Sentence-Transformers MiniLM-L6-v2 embeddings \cite{reimers2019sentence} and cosine similarity for semantic matching. For the coding reasoning mode, we use the NetworkX documentation curated by GraphTeam \cite{li2025graphteamfacilitatinglargelanguage} as the knowledge base\footnote{\url{https://github.com/BUPT-GAMMA/GraphTeam/blob/main/multi-agents-4-graph-analysis/GraphTeam/data/filtered_networkx_reference_edition.json}}, denoted as \texttt{kb-graphteam}. For the textual reasoning mode, we manually construct reasoning procedures for classical graph algorithms and retrieve the top-3 relevant algorithmic knowledge entries as references, denoted as \texttt{alg-top3}.

\textbf{Finding 1: Retrieval augmentation is unstable in coding reasoning and can even reduce overall performance.}
Table~\ref{tab:rag-overall} reports the overall performance before and after retrieval augmentation in both coding and textual reasoning modes. For coding reasoning, retrieval augmentation decreases the overall EM for all three base models: the relative changes for DeepSeek-V3.2, Qwen2.5-72B, and Qwen2.5-7B are $-6.2\%$, $-7.8\%$, and $-4.7\%$, respectively. This indicates that directly introducing external knowledge does not necessarily improve code-based graph reasoning.

To better understand this trend, Table~\ref{tab:rag-coding-leetcode} further breaks down the results by task source and description style: OA-Explicit (OA-E), OA-Implicit (OA-I), classical-Explicit (CL-E), and classical-Implicit (CL-I). Retrieval improves all three coding models on the CL-E cell, with EM relative changes of $+5.6\%$, $+3.2\%$, and $+14.2\%$, and PC follows the same direction. This suggests that external knowledge is helpful when tasks are closely aligned with standard graph algorithms and are explicitly described.

However, the benefit becomes much weaker, or even negative, in other settings. In the OA-E setting, the three models show mixed changes. In the OA-I setting, all three models degrade, with relative EM changes of $-4.2\%$, $-13.0\%$, and $-20.9\%$. In the CL-I setting, the changes are also consistently negative, with $-3.0\%$, $-4.4\%$, and $-15.2\%$. These results suggest that once tasks become closer to OA-style problems, or require the model to infer the underlying graph problem from implicit natural language, retrieved external knowledge becomes harder to use reliably.

\textit{Why does coding reasoning benefit less?}
In coding reasoning, retrieved graph-algorithm knowledge typically enters the prompt as relatively concrete implementation guidance that the model must integrate into its generated Python program. When the task closely matches a textbook-style graph algorithm, such as in the CL-E setting, this knowledge can be naturally converted into code and thus brings stable gains. In contrast, OA-style problems often combine standard graph primitives with task-specific constraints, such as state augmentation, custom recurrence relations, or special-case handling. In such cases, the model may copy the retrieved classical algorithm without sufficiently adapting it to the current task, producing code that resembles a standard algorithm but does not match the actual input distribution. In addition, implicit real-world descriptions require the model to first map narrative semantics to graph-algorithm primitives, which further increases retrieval difficulty. If the retrieved content anchors the model to an incorrect primitive, external knowledge can become distracting rather than helpful. Therefore, in the OA-I setting, these two challenges compound, making retrieval augmentation most likely to produce negative effects. This result suggests that future graph reasoning retrieval methods should better handle open-ended task sources, implicit task descriptions, and the adaptation of retrieved knowledge to task-specific constraints.

\textbf{Finding 2: Textual reasoning benefits more from retrieval augmentation, but the gain remains scenario-dependent.}
Table~\ref{tab:rag-overall} shows that retrieval augmentation improves the overall EM of all three models in textual reasoning, with relative changes of $+1.3\%$, $+21.3\%$, and $+21.0\%$ for DeepSeek-V3.2, Qwen2.5-72B, and Qwen2.5-7B, respectively. Table~\ref{tab:rag-textual-leetcode} provides a more fine-grained analysis by task source and description style. In the CL-E cell, all three models improve, with EM relative changes of $+6.0\%$, $+5.9\%$, and $+5.2\%$, consistent with the trend observed in coding reasoning.

For the other cells, retrieval gains in textual reasoning are generally milder but still scenario-dependent. In the OA-E cell, all three models show non-negative changes, with $+0.0\%$, $+0.0\%$, and $+33.3\%$. The OA-I cell is the least stable, with changes of $-10.8\%$, $-48.0\%$, and $+4.0\%$. In the CL-I cell, all three models show positive changes, with $+4.4\%$, $+1.9\%$, and $+3.8\%$. Overall, textual reasoning benefits from retrieval more often than coding reasoning and is less likely to suffer large performance drops. Nevertheless, retrieval remains unstable in the OA-I setting, where task source and implicit description jointly increase difficulty.

\textit{Why is textual reasoning more stable?}
Textual reasoning uses retrieved knowledge as a conceptual context rather than as a code template that must be directly integrated into a program. Thus, retrieved passages can provide relevant terminology or algorithmic steps without directly causing binary code-execution failures. By contrast, an incorrect implementation in coding reasoning can make the entire code output wrong. As a result, textual reasoning is relatively more robust to retrieval noise. However, when the retrieved knowledge does not match the true task structure, textual reasoning also struggles to use external information effectively. This is especially evident in the OA-I setting, where uncertainty from both task source and description form leads to clear fluctuation in retrieval gains.

\begin{takeawaybox}
Across Tables~\ref{tab:rag-overall}, \ref{tab:rag-coding-leetcode}, and \ref{tab:rag-textual-leetcode}, retrieval augmentation is not a universally effective improvement, but rather a scenario-dependent auxiliary mechanism. It is most stable and useful for classical tasks with explicit descriptions, because these tasks are closer to standard graph-algorithm knowledge. When tasks are OA-derived or implicitly described, the retrieval benefit becomes weaker and can even harm performance. Compared with coding reasoning, textual reasoning is a more stable setting for retrieval augmentation, but its gains are still affected by task source and description style. Therefore, RAG methods for complex graph reasoning should go beyond simple semantic retrieval and further consider task modeling, constraint identification, and the alignment between retrieved knowledge and the structure of the task.
\end{takeawaybox}

\subsection{Fine-Tuned LLMs vs. Base Models}
\label{sec:appendix_finetune}

\begin{table*}[t]
\centering
\scriptsize
\caption{Comparison between \textsc{GraphWiz} and its base model \textsc{LLaMA-2-7B} in textual reasoning. We report strict Exact Match (EM) and Partial Credit (PC). All values are percentages, with the percent sign omitted. E and I denote explicit and implicit textual scenarios, respectively.}
\label{tab:graphwiz-merged}
\setlength{\tabcolsep}{2pt}
\renewcommand{\arraystretch}{1.08}
\begin{tabularx}{\linewidth}{l *{14}{>{\centering\arraybackslash}X}}
\toprule
 & \multicolumn{2}{c}{\textbf{Overall}}
 & \multicolumn{2}{c}{\textbf{Scenario}}
 & \multicolumn{4}{c}{\textbf{Graph Size}}
 & \multicolumn{4}{c}{\textbf{Combo Size}}
 & \multicolumn{2}{c}{\textbf{Task Source}} \\
\cmidrule(lr){2-3}
\cmidrule(lr){4-5}
\cmidrule(lr){6-9}
\cmidrule(lr){10-13}
\cmidrule(lr){14-15}
Model & EM & PC & E & I & $10$ & $100$ & $1$K & $10$K & 1 & 2 & 3 & 4 & OA & CL \\
\midrule
\textsc{GraphWiz}
& $\mathbf{4.2}$ & $\mathbf{4.4}$
& $\mathbf{4.4}$ & $\mathbf{4.1}$
& $\mathbf{15.0}$ & $\mathbf{2.0}$ & $0.0$ & $0.0$
& $\mathbf{4.8}$ & $\mathbf{4.2}$ & $\mathbf{4.1}$ & $\mathbf{3.6}$
& $\mathbf{0.6}$ & $\mathbf{7.2}$ \\

\textsc{LLaMA-2-7B}
& $0.5$ & $0.6$
& $0.7$ & $0.3$
& $2.0$ & $0.0$ & $0.0$ & $0.0$
& $1.3$ & $0.5$ & $0.2$ & $0.0$
& $0.0$ & $0.0$ \\
\midrule
$\Delta$ 
& $+3.7$ & $+3.8$
& $+3.7$ & $+3.8$
& $+13.0$ & $+2.0$ & $0.0$ & $0.0$
& $+3.5$ & $+3.7$ & $+3.9$ & $+3.6$
& $+0.6$ & $+7.2$ \\
\bottomrule
\end{tabularx}
\end{table*}

\begin{table*}[t]
\centering
\scriptsize
\caption{
Comparison between \textsc{GCoder} and its base model \textsc{LLaMA-3.1-8B} in the coding setting.
We report strict Exact Match (EM) and Partial Credit (PC), together with breakdowns by reasoning scenario, graph size, and inline graph-extraction quality.
All values are percentages, with the percent sign omitted.
EF, EI, IF, and II denote explicit-file, {explicit-inline}, {implicit-file}, and {implicit-inline}, respectively.
}
\label{tab:gcoder-merged}
\setlength{\tabcolsep}{2pt}
\renewcommand{\arraystretch}{1.08}
\begin{tabularx}{\linewidth}{l *{13}{>{\centering\arraybackslash}X}}
\toprule
& \multicolumn{2}{c}{\textbf{Overall}}
& \multicolumn{4}{c}{\textbf{Scenario}}
& \multicolumn{4}{c}{\textbf{Graph Size}}
& \multicolumn{3}{c}{\textbf{Graph Extraction}} \\
\cmidrule(lr){2-3}
\cmidrule(lr){4-7}
\cmidrule(lr){8-11}
\cmidrule(lr){12-14}
Model
& EM & PC
& EF & EI & IF & II
& $10$ & $100$ & $1$K & $10$K
& Full & Nodes & Edges \\
\midrule
\textsc{GCoder}
& $16.1$ & $17.0$
& $24.1$ & $9.7$ & $20.5$ & $10.0$
& $24.0$ & $19.4$ & $12.5$ & $8.5$
& $9.6$ & $11.3$ & $12.8$ \\

\textsc{LLaMA-3.1-8B}
& $\mathbf{29.4}$ & $\mathbf{31.4}$
& $\mathbf{35.7}$ & $\mathbf{23.6}$ & $\mathbf{36.1}$ & $\mathbf{22.2}$
& $\mathbf{41.8}$ & $\mathbf{36.3}$ & $\mathbf{24.0}$ & $\mathbf{15.6}$
& $\mathbf{55.1}$ & $\mathbf{61.4}$ & $\mathbf{71.7}$ \\
\midrule
$\Delta$
& $+13.3$ & $+14.4$
& $+11.6$ & $+13.9$ & $+15.6$ & $+12.2$
& $+17.8$ & $+16.9$ & $+11.5$ & $+7.1$
& $+45.5$ & $+50.1$ & $+58.9$ \\
\bottomrule
\end{tabularx}
\end{table*}


Prior studies have explored graph-domain instruction-tuning as a way to improve LLMs' graph reasoning ability, including instruction tuning for textual graph reasoning, as in GraphWiz~\cite{chen2024graphwizinstructionfollowinglanguagemodel}, and code-oriented fine-tuning for graph coding, as in GCoder~\cite{zhang2024gcoderimprovinglargelanguage}. This subsection evaluates whether such graph-domain fine-tuned models can generalize to our {\dataset} benchmark. 

\textbf{Experimental setup.}
In the textual setting, we evaluate GraphWiz~\cite{chen2024graphwizinstructionfollowinglanguagemodel} and its base model, LLaMA-2-7B. GraphWiz is built on LLaMA-2-7B and instruction-tuned on GraphInstruct, which contains $17{,}158$ graph question--answer examples across nine canonical graph tasks, such as cycle detection, connectivity, shortest path, and topological sorting. Its training template follows an Alpaca-style instruction format and uses chain-of-thought rationales ending with a \texttt{\#\#\#~<answer>} marker. We download the public checkpoint \href{https://huggingface.co/GraphWiz/LLaMA2-7B}{\texttt{GraphWiz/LLaMA2-7B}} and compare it with the unmodified {LLaMA-2-7B} under the same textual reasoning pipeline. Both models use identical prompts and \texttt{max\_tokens}$=2048$. Because GraphWiz's output format differs from the \texttt{<answer>...</answer>} tags expected by our evaluator, we normalize the outputs before evaluation: for GraphWiz, we convert the text after the final \texttt{\#\#\#} separator into an \texttt{<answer>} tag and normalize \texttt{Yes}/\texttt{No}, answer prefixes, and list literals; for the base model, we preserve explicitly generated \texttt{<answer>} tags and otherwise treat the prediction as empty. This post-processing reduces the effect of format mismatch and better isolates reasoning ability.

In the code setting, we evaluate GCoder~\cite{zhang2024gcoderimprovinglargelanguage} and its base model, {LLaMA-3.1-8B}. GCoder is based on LLaMA-3-8B and post-trained on the GraphWild graph-code corpus~\footnote{https://github.com/Graph-Reasoner/GCoder/blob/master/Dataset/training\_dataset/GWild.json}, together with supervised fine-tuning using compiler feedback. Since the authors have not released the trained checkpoint, we reproduce GCoder on our A6000 GPU server and integrate the resulting model into our code reasoning pipeline. We then compare it with {LLaMA-3.1-8B} on the full {\dataset} task suite. For both experiments, we report strict EM and PC results. Table~\ref{tab:graphwiz-merged} summarizes the textual reasoning results for GraphWiz, and Table~\ref{tab:gcoder-merged} summarizes the code reasoning results for GCoder.

\textbf{Experimental analysis.}
Table~\ref{tab:graphwiz-merged} shows that GraphWiz yields a clear relative improvement in textual reasoning: its strict EM is $4.2$, compared with only $0.5$ for {LLaMA-2-7B}, corresponding to a $+3.7$ absolute gain. This advantage is consistent across both explicit and implicit textual scenarios, suggesting that instruction tuning improves the model's ability to produce evaluable answers under graph question-answering formats. However, the absolute performance remains very low and drops rapidly as graph size increases: GraphWiz achieves an EM of $15.0$ at $n{=}10$, decreases to $2.0$ at $n{=}100$, and reaches $0.0$ at both $n{=}1{,}000$ and $n{=}10{,}000$. Moreover, GraphWiz achieves an EM of $7.2$ on classical graph tasks, substantially higher than its $0.6$ EM on OA tasks, indicating that its gains mainly arise on tasks closer to the GraphInstruct distribution.

In contrast, Table~\ref{tab:gcoder-merged} shows that GCoder consistently underperforms its base model in code-based reasoning. GCoder obtains an overall strict EM of $16.1$, whereas {LLaMA-3.1-8B} reaches $29.4$, corresponding to a $13.3$ percentage-point absolute drop; PC similarly decreases from $31.4$ to $17.0$. This degradation appears across all scenarios. The gap is particularly pronounced in graph extraction: GCoder's full-graph extraction accuracy is only $9.6$, while the base model reaches $55.1$; node and edge extraction accuracy also drops from $61.4$ and $71.7$ to $11.3$ and $12.8$, respectively. These results suggest that graph-domain code fine-tuning does not improve GCoder's code reasoning ability on our benchmark and instead weakens its ability to handle graph reasoning. 

\textbf{Finding 1: Graph-domain instruction-tuning has limited generalization to out-of-distribution tasks.}
Both GraphWiz and GCoder exhibit strong dependence on their training distributions. GraphWiz improves over {LLaMA-2-7B} in textual reasoning, but its gains are mainly observed on small graphs and classical graph-theory tasks. As the graph size increases from $n{=}10$ to $n{=}100$ and beyond, its performance rapidly approaches zero. By task source, GraphWiz performs much better on classical tasks than on OA tasks. This pattern is consistent with the construction of GraphInstruct, which primarily covers canonical graph tasks such as cycle detection, connectivity, shortest path, and topological sorting, rather than OA-style or compositional graph reasoning tasks.

GCoder shows a stronger form of negative transfer. Although GraphWild contains examples similar to single-task graph coding problems, it does not cover the richer compositional tasks, graph-loading modes, and large-scale graph settings in {\dataset}. GCoder's performance curve across different combo sizes is relatively flat but consistently low, suggesting that the model may have learned a fixed set of graph-code templates rather than a general graph reasoning capability that adapts to task complexity. Thus, graph-domain instruction-tuning is effective only when the training corpus covers the scale, task source, and compositional structure of the target benchmark; otherwise, the fine-tuned model can fail under out-of-distribution settings.

\textbf{Finding 2: Fine-tuning gains or regressions largely reflect format adherence, library usage, and parsing behavior rather than reasoning depth alone.}
GraphWiz's improvement partly comes from more reliable answer formatting. Since GraphWiz is trained to produce a fixed \texttt{\#\#\#~<answer>} output pattern, after output normalization, it is more likely than the base model to produce parseable answers. In contrast, {LLaMA-2-7B} often fails to generate the answer tags required by the evaluator. Therefore, GraphWiz's relative improvement does not necessarily indicate substantially stronger graph-algorithmic reasoning; part of the gain comes from better format adherence.

GCoder's regression is mainly associated with code-level failure modes. By inspecting cached responses and error labels, we find that GCoder frequently hallucinates nonexistent NetworkX APIs. In the \texttt{coding\_explicit\_file} scenario, among roughly $424$ runtime errors, $281$ ($66.3\%$) are \texttt{AttributeError}s, including incorrect uses of nonexistent or inappropriate functions such as \texttt{nx.local\_clustering}, \texttt{nx.tree\_radius}, \texttt{nx.minimum\_height\_trees}, and \texttt{nx.is\_triangle\_free}. This suggests that GCoder may have learned function-name patterns from its training corpus without preserving reliable knowledge of the actual NetworkX API. In addition, GCoder tends to output long inline graph data inside \texttt{\_\_main\_\_}, even when the prompt already provides a \texttt{file\_path}, which causes truncation and syntax errors under \texttt{max\_tokens}$=8192$. Other errors include repeated small spelling or bracket mistakes, such as missing quotation marks, mismatched parentheses, and incorrect list delimiters. Taken together, GCoder's failures are not solely due to graph reasoning difficulty; they also arise from degradation in library knowledge, output structure, and parsing behavior after graph-code fine-tuning.

\begin{takeawaybox}
Fine-tuning on narrow graph-task corpora does not necessarily improve LLM graph reasoning. GraphWiz provides measurable gains in the textual setting, but these gains are mainly limited to small graphs and tasks close to the GraphInstruct distribution, and part of the improvement comes from better format adherence. GCoder, in contrast, substantially underperforms its base model in the code setting, mainly due to hallucinated NetworkX API calls, truncated inline graph data, and brittle code formatting. Overall, unless the fine-tuning corpus covers the graph sizes, task sources, compositional structures, and practical library APIs required by the target benchmark, graph-domain instruction-tuning may fail to generalize and can even weaken abilities already present in the base model.
\end{takeawaybox}

\subsection{Task-Specific GNNs vs. LLM-Based Graph Reasoning}
\label{sec:appendix_gnn}

GNNs are neural architectures designed to learn representations over graph-structured data and are widely used for graph learning tasks, where models learn mappings from node/edge features and graph structures to label spaces. Although their primary purpose is not to explicitly execute graph algorithms, prior work has explored their use for algorithmic graph learning and graph reasoning evaluation. Following GraphArena~\cite{tang2025grapharena}, we select four representative GNN backbones: GCN~\cite{kipf2017semisupervised}, GAT~\cite{velikov2018graph}, GraphSAGE~\cite{hamilton2017inductive}, and GIN~\cite{xu2018how}, and compare them with representative LLMs under both text-based and code-based reasoning.

\textbf{Experimental setup.}
Since GNNs typically require a fixed scalar output space, we first filter the full set of $202$ tasks to obtain a subset suitable for GNN evaluation. A task is retained only if its ground-truth label is a single scalar, i.e., a boolean or non-negative integer scalar, across all node sizes. This yields $70$ GNN-eligible tasks. The remaining tasks have labels such as lists, dictionaries, paths, matrices, sets, or other non-scalar structures, which cannot be directly modeled by a single graph-level regression head. For each retained task, we execute its graph generation code $800$ times using our benchmark construction framework, with graph sizes uniformly sampled from $[10,50]$, and run the corresponding \texttt{solve} function to obtain labels under a $5$-second timeout.

Each graph is converted into a graph representation with self-loops. The GNN model follows a two-layer architecture: \texttt{conv1} and \texttt{conv2} are selected from \{GIN, GAT, SAGE, GCN\}, with hidden dimension $16$ and LeakyReLU activation, followed by sum-pooling and a linear layer that outputs a single scalar. For tasks with source/target inputs, we concatenate the sum-pooled graph representation with the source-node and target-node embeddings before prediction. We train one model for each $(\text{task}, \text{GNN})$ pair, using a $90/10$ train/validation split, batch size $32$, Adam optimizer, learning rate $\mathrm{lr}=10^{-2}$, MSE loss, up to $50$ epochs, and patience-$10$ early stopping based on validation loss. In total, we train $4 \times 70 = 280$ GNN checkpoints.

During evaluation, each checkpoint is tested on the same evaluation instances used for LLMs, with node sizes $\{10,100,1{,}000,10{,}000\}$. Continuous predictions are rounded and compared with ground-truth labels using exact match. To ensure a fair comparison, the LLM results are also aggregated over the same $70$ GNN-eligible tasks. We compare the four GNN backbones with three representative LLMs under code-based and text-based reasoning: \texttt{DeepSeek-V3.2}, \texttt{DeepSeek-R1}, and \texttt{Qwen2.5-72B}. We define the retention ratio as the EM at the largest setting divided by the EM at the smallest setting. Table~\ref{tab:gnn-vs-llms} reports the results, from which we draw the following findings.

\begin{table*}[t]
\centering
\caption{Strict Exact Match (EM, \%) for the four GNN backbones and representative LLMs, evaluated on the same $70$-task GNN-eligible subset of the benchmark. Columns report overall EM, EM by graph size, and EM by combo size. Ret. denotes the retention ratio, computed as the largest-setting EM divided by the smallest-setting EM. Average rows give the per-column mean within each panel.}
\label{tab:gnn-vs-llms}
\footnotesize
\setlength{\tabcolsep}{2.5pt}
\begin{tabularx}{\linewidth}{l Y YYYY Y YYYY Y}
\toprule
                         & Overall & \multicolumn{5}{c}{Graph size} & \multicolumn{5}{c}{Combo size} \\
\cmidrule(lr){3-7}\cmidrule(lr){8-12}
Model                    & EM      & 10 & 100 & 1k & 10k & Ret. & 1 & 2 & 3 & 4 & Ret. \\
\midrule
\multicolumn{12}{l}{\textit{Panel A: GNN baselines (4 backbones, $70$ tasks each)}} \\
\midrule
GAT                      & $47.5$ & $62.9$ & $61.4$ & $42.9$ & $22.9$ & $36.4\%$ & $59.0$ & $56.2$ & $41.7$ & $25.0$ & $42.4\%$ \\
GCN                      & $48.2$ & $62.9$ & $61.4$ & $44.3$ & $24.3$ & $38.6\%$ & $60.0$ & $56.2$ & $41.7$ & $27.1$ & $45.2\%$ \\
GIN                      & $40.0$ & $58.6$ & $51.4$ & $31.4$ & $18.6$ & $31.7\%$ & $45.8$ & $45.0$ & $40.5$ & $22.9$ & $50.0\%$ \\
GraphSAGE                & $46.1$ & $64.3$ & $61.4$ & $41.4$ & $17.1$ & $26.6\%$ & $59.0$ & $54.2$ & $36.9$ & $27.1$ & $45.9\%$ \\
\cmidrule(l){2-12}
\textit{Average}         & $45.5$ & $62.2$ & $58.9$ & $40.0$ & $20.7$ & $33.3\%$ & $56.0$ & $52.9$ & $40.2$ & $25.5$ & $45.5\%$ \\
\midrule
\multicolumn{12}{l}{\textit{Panel B: LLMs --- coding mode}} \\
\midrule
DeepSeek-V3.2            & $63.1$ & $91.4$ & $78.3$ & $49.7$ & $32.9$ & $36.0\%$ & $67.6$ & $65.8$ & $64.0$ & $50.8$ & $75.1\%$ \\
DeepSeek-R1              & $64.7$ & $92.8$ & $74.9$ & $54.0$ & $37.1$ & $40.0\%$ & $71.6$ & $69.4$ & $64.2$ & $52.7$ & $73.6\%$ \\
Qwen2.5-72B              & $55.5$ & $80.0$ & $67.9$ & $46.9$ & $27.2$ & $34.0\%$ & $65.5$ & $61.0$ & $59.4$ & $32.1$ & $49.0\%$ \\
\cmidrule(l){2-12}
\textit{Average}         & $61.1$ & $88.1$ & $73.7$ & $50.2$ & $32.4$ & $36.8\%$ & $68.2$ & $65.4$ & $62.5$ & $45.2$ & $66.3\%$ \\
\midrule
\multicolumn{12}{l}{\textit{Panel C: LLMs --- textual mode}} \\
\midrule
DeepSeek-V3.2            & $55.5$ & $96.0$ & $71.1$ & $43.6$ & $11.4$ & $11.9\%$ & $61.5$ & $58.7$ & $56.4$ & $44.2$ & $71.9\%$ \\
DeepSeek-R1              & $44.4$ & $90.1$ & $49.0$ & $27.2$ & $11.3$ & $12.5\%$ & $51.1$ & $48.1$ & $49.0$ & $24.2$ & $47.4\%$ \\
Qwen2.5-72B              & $18.0$ & $47.0$ & $17.2$ & $6.6$ & $1.3$ & $2.8\%$ & $22.9$ & $18.9$ & $18.3$ & $12.5$ & $54.6\%$ \\
\cmidrule(l){2-12}
\textit{Average}         & $39.3$ & $77.7$ & $45.8$ & $25.8$ & $8.0$ & $10.3\%$ & $45.2$ & $41.9$ & $41.2$ & $27.0$ & $59.7\%$ \\
\bottomrule
\end{tabularx}
\end{table*}

\textbf{Finding 1: Overall performance follows LLM code-based reasoning $>$ GNNs $>$ LLM text-based reasoning.}
On the $70$ GNN-eligible tasks, the average Overall strict EM is $45.5$ for GNNs, $61.1$ for LLMs under code-based reasoning, and $39.3$ for LLMs under text-based reasoning. Thus, code-based LLM reasoning outperforms GNNs by $15.6$ percentage points and text-based LLM reasoning by $21.8$ percentage points, while GNNs lie between the two LLM reasoning modes and outperform text-based LLM reasoning by $6.2$ percentage points. 
More importantly, LLMs under code-based reasoning outperform GNNs not only in Overall EM, but also across every node-size bucket and every combo-size bucket. Even on the GNN-eligible subset, where the input/output format has been explicitly restricted to scalar prediction tasks suitable for GNN modeling, specialized graph-supervised training remains less accurate than general-purpose LLMs that generate Python programs and delegate execution to the interpreter. This suggests that executable program generation provides a clear empirical advantage for LLMs in graph algorithmic reasoning.

\textbf{Finding 2: As graph size increases, all reasoning modes degrade, but the retention ratio follows LLM code-based reasoning $>$ GNNs $>$ LLM text-based reasoning.}
As the graph size grows, all three reasoning modes exhibit performance degradation, but with different patterns. 
LLMs under code-based reasoning scale best: their average EM decreases from $88.1$ at $n{=}10$ to $32.4$ at $n{=}10{,}000$, corresponding to a retention ratio of $37\%$, the highest among the three reasoning modes. This trend suggests that code solutions provide stronger graph scale robustness: once the model generates an executable program, algorithm execution is delegated to the Python interpreter and is therefore less directly affected by graph size. Nevertheless, when graph data are embedded in the prompt, larger graphs still increase the likelihood of parsing failures, missing predictions, syntax errors, and runtime errors.
GNNs show moderate scalability, with average EM decreasing from $62.2$ to $20.7$, corresponding to a retention ratio of $33\%$. This degradation is consistent with the training setup: GNNs are trained only on graphs sampled from $[10,50]$ nodes, placing $n{=}1{,}000$ and $n{=}10{,}000$ far outside the training-size distribution. In addition, the magnitude of sum-pooling readouts can scale with graph size, which may make scalar prediction less stable on large graphs. LLMs under text-based reasoning degrade the fastest, with average EM dropping from $77.7$ to $8.0$ and a retention ratio of only $10\%$, indicating that performing step-by-step algorithmic reasoning over serialized graphs in natural language becomes highly unreliable under long graph contexts.

It is worth noting that although LLMs under text-based reasoning underperform GNNs overall, they still have an advantage on small graphs: at $n{=}10$, text-based LLM reasoning achieves a higher average EM than GNNs ($77.7$ vs.\ $62.2$). Therefore, when code-based reasoning is unavailable and the graph is small, text-based reasoning can be competitive, while for medium and large graphs task-specific GNN baselines are more reliable non-coding alternatives than text-based LLM reasoning.

\textbf{Finding 3: As combo size increases, all three reasoning modes face compositional reasoning challenges.}
Combo size denotes the number of seed tasks composed in each benchmark instance. As combo size increases from $1$ to $4$, all three reasoning modes experience performance drops. LLMs under code-based reasoning maintain the highest average EM across all combo sizes, with scores of $68.2$, $65.4$, $62.5$, and $45.2$, respectively. Their decline from combo $1$ to combo $3$ is relatively mild, but the drop becomes much sharper from combo $3$ to combo $4$, with a retention ratio of $66\%$. This indicates that code-based reasoning is relatively robust to moderate composition but still struggles with deeper task chains. GNNs also show a clear decline, with average EM decreasing from $56.0$ to $52.9$, $40.2$, and $25.5$, corresponding to a retention ratio of $46\%$. This suggests that cross-subtask dependencies introduced by higher compositionality are difficult for graph-level supervised models to capture. LLMs under text-based reasoning decrease from $45.2$ to $41.9$, $41.2$, and $27.0$, with a retention ratio of $60\%$. This higher retention should be interpreted cautiously, as the lower initial accuracy leaves less room for absolute degradation. Overall, compositional tasks challenge code execution, graph-supervised learning, and natural-language reasoning simultaneously, indicating that composite tasks are an important dimension for evaluating complex graph reasoning.

\begin{takeawaybox}
On the $70$-task GNN-eligible subset, the Overall strict EM follows the order: LLMs under code-based reasoning ($61.1$) $>$ GNNs ($45.5$) $>$ LLMs under text-based reasoning ($39.3$). Code-based LLM reasoning remains strongest across all graph sizes and combo sizes, suggesting that executable program generation provides a clear advantage for graph algorithmic reasoning. Although task-specific GNNs are trained from scratch on synthetic in-distribution samples for each task, they remain less accurate than code-based LLM reasoning. Their main value is as a stronger non-coding alternative to text-based LLM reasoning on medium and large graphs, where natural-language step-by-step reasoning over serialized graphs degrades sharply.
\end{takeawaybox}

\section{Prompt Design}
\label{sec:appendix_prompt_design}

\refstepcounter{subsection}
\label{prompt_stage1}
\begin{stageprompt}{Stage 1 Prompt: From Seed Tasks to a Composite Task and Graph Generator}

You are given a combo size, a list of sampled seed tasks, and candidate composition patterns for graph reasoning.

\textbf{Input.}
Each seed task includes its name, definition, graph type, and input/output specification. The candidate composition patterns are: sequential, constrained, hierarchical, counterfactual, map-reduce, aggregate, and logical-comparative.

\textbf{Task.}
Choose the most suitable composition pattern for the sampled seed tasks. If \texttt{combo size = 1}, treat it as a degenerate composite task and set \texttt{composition\_way} to \texttt{null}. Then generate:
\begin{itemize}
    \item a new composite task definition, and
    \item Python graph-generation code for that task.
\end{itemize}

The generated code must support graph sizes \texttt{10}, \texttt{100}, \texttt{1000}, and \texttt{10000}, and the generated graph should remain consistent with the requirements of the composed task.

\textbf{Output.}
Return only a JSON object with the fields:
\texttt{composition\_way}, \texttt{task\_name}, \texttt{task\_definition}, and \texttt{graph\_generation\_code}.

\end{stageprompt}

\refstepcounter{subsection}
\label{prompt_stage2}
\begin{stageprompt}{Stage 2 Prompt: From Task Definition to Executable Solver and Gold Labels}

You are given a graph reasoning task definition, the graph-data format, and an example test case.

\textbf{Input.}
The task specifies the required graph operations, expected input/output format, and example graph instance.

\textbf{Task.}
Generate executable Python solver code for the task. The solver should implement the task logic exactly as specified, read the graph and input parameters, carry intermediate results correctly for compositional tasks, and return the final answer. The generated code should remain valid across graph instances of different sizes.

\textbf{Output.}
Return only Python code. The code must include a task-specific \texttt{solve(...)} function together with any helper functions needed for execution on the generated graph instances.

\textbf{Revision.}
If the solver fails functional validation, revise it using the provided error message and human feedback until it passes.

\end{stageprompt}

\refstepcounter{subsection}
\label{prompt_stage3}
\begin{stageprompt}{Stage 3 Prompt: From Task Definition to Multiple Descriptions, Questions, and Loading Modes}

You are given a graph reasoning task definition and its reference solution program.

\textbf{Input.}
The task definition specifies the graph reasoning problem, and the reference solution provides the target task logic.

\textbf{Task.}
Generate multiple prompt forms for the same underlying task. Specifically:
\begin{itemize}
    \item an \textbf{explicit} description that directly states the graph problem,
    \item an \textbf{implicit} description that presents it as a realistic narrative,
    \item a \textbf{text-mode} question for natural-language reasoning,
    \item a \textbf{code-mode} question for solver completion,
    \item an \textbf{inline} loading script where graph data appear in the prompt,
    \item a \textbf{file} loading script where graph data are loaded from a file.
\end{itemize}

All outputs must remain consistent with the same task and reference solution.

\textbf{Output.}
Return only a JSON object with:
\texttt{description\_explicit}, \texttt{description\_implicit}, \texttt{question\_text}, \texttt{question\_code}, \texttt{loading\_script\_inline}, and \texttt{loading\_script\_file}.

\end{stageprompt}

\refstepcounter{subsection}
\label{prompt_stage4}
\begin{stageprompt}{Stage 4 Prompt: From Task Definition to Text and Code Evaluators}

You are given a graph reasoning task definition, a question format, and a reference solution program.

\textbf{Input.}
The task definition specifies the target output, the question format specifies how the answer is asked, and the reference solution defines the gold task logic.

\textbf{Task.}
Generate two task-specific evaluation scripts:
\begin{itemize}
    \item a \textbf{text evaluator} that extracts the final answer from a written response and compares it with the gold label,
    \item a \textbf{code evaluator} that executes the generated \texttt{solve} function and compares its output with the gold label.
\end{itemize}

The evaluators should support exact-match scoring and partial-credit scoring when appropriate, and they should handle the output type correctly.

\textbf{Output.}
Return only a JSON object with:
\texttt{eval\_script\_text} and \texttt{eval\_script\_code}.

\end{stageprompt}

\refstepcounter{subsection}
\label{prompt_stage5}
\begin{stageprompt}{Stage 5 Prompt: From Generated Task to Quality Scores and Filtering Decision}

You are given a graph reasoning task, its generated descriptions, question formats, and reference solution.

\textbf{Input.}
The input includes the task definition, explicit and implicit descriptions, text-mode and code-mode questions, and the reference solution.

\textbf{Task.}
Score the task on four quality checks:
\begin{itemize}
    \item \textbf{clarity}: is the task easy to understand?
    \item \textbf{graph suitability}: is the graph appropriate for the task?
    \item \textbf{naturalness}: does the real-world version sound realistic?
    \item \textbf{answer uniqueness}: is there one well-defined correct answer?
\end{itemize}

Use a 1--5 scale, provide a short justification for each score, and mark the task for removal if any score is below 3.

\textbf{Output.}
Return only a JSON object containing the four scores, their short justifications, and a final \texttt{keep\_task} decision.

\end{stageprompt}


\end{document}